\documentclass[twoside,11pt]{article}

\PassOptionsToPackage{round}{natbib}
\usepackage{arxiv}

\include{preamble}
\include{maths-preamble}

\jmlrheading{XX}{2021}{1-XX}{XX/21}{XX/XX}{16-603}{Matthew C.~Ashman, Thang D.~Bui, Cuong V.~Nguyen, Stratis Markou, Adrian Weller, Siddharth Swaroop, and Richard E.~Turner}
\ShortHeadings{Partitioned Variational Inference }{Ashman, Bui, Nguyen, Markou, Weller, Swaroop, and Turner}
\firstpageno{1}

\begin{document}

\title{Partitioned Variational Inference: A  Framework for Probabilistic Federated Learning}
\author{\name Matthew Ashman \email mca39@cam.ac.uk\\
\addr University of Cambridge, Cambridge, UK\AND
\name Thang D.~Bui \email thang.bui@sydney.edu.au\\
\addr University of Sydney, Sydney, Australia\AND
\name Cuong V.~Nguyen \email vcnguyen@cs.fiu.edu\\
\addr KFSCIS, Florida International University, Florida, USA\AND
\name Stratis Markou \email em626@cam.ac.uk \\
\addr University of Cambridge, Cambridge, UK\AND
\name Adrian Weller \email aw665@cam.ac.uk\ \\
\addr University of Cambridge, Cambridge, UK\AND
\name Siddharth Swaroop \email ss2163@cam.ac.uk\\
\addr University of Cambridge, Cambridge, UK\AND
\name Richard E.~Turner \email ret26@cam.ac.uk\\
\addr 
University of Cambridge, Cambridge, UK }

\maketitle

\begin{abstract}
The proliferation of computing devices has brought about an opportunity to deploy machine learning models on new problem domains using previously inaccessible data. Traditional algorithms for training such models often require data to be stored on a single machine with compute performed by a single node, making them unsuitable for decentralised training on multiple devices. This deficiency has motivated the development of federated learning algorithms, which allow multiple data owners to train collaboratively and use a shared model whilst keeping local data private. However, many of these algorithms focus on obtaining point estimates of model parameters, rather than probabilistic estimates capable of capturing model uncertainty, which is essential in many applications. Variational inference (VI) has become the method of choice for fitting many modern probabilistic models. In this paper we introduce partitioned variational inference (PVI), a general framework for performing VI in the federated setting. We develop new supporting theory for PVI, demonstrating a number of properties that make it an attractive choice for practitioners; use PVI to unify a wealth of fragmented, yet related literature; and provide empirical results that showcase the effectiveness of PVI in a variety of federated settings.

\end{abstract}



\section{Introduction}
\label{sec:introduction}
As networks of connected devices, data centres and centralised servers proliferate, there is a corresponding need for distributed machine learning algorithms---algorithms that operate across multiple platforms, utilising the computational capabilities and accounting for the privacy constraints of each. The family of machine learning algorithms required for this setting differs considerably from those required for traditional training pipelines, many of which require data to be stored in a single machine with compute performed by a single node. Federated learning attempts to bridge this gap, through algorithms that allow multiple data owners---or clients---to collaboratively train and use a shared model without sharing local data. Clients are then free to employ whichever privacy preserving strategies they feel necessary, such as differential privacy \citep{dwork2006calibrating,jalko+al:17}.
Federated learning algorithms can be helpful in a wide variety of settings. For example, a hospital may wish to develop a predictive model for a particular medical diagnosis. However, the data collected by an individual hospital may be sparse or biased towards a particular demographic. This has a detrimental effect on properties of the trained model, such as accuracy, robustness and fairness. Instead, it is desirable to collaborate with other hospitals to develop a shared model that utilises the data collected by each. Federated learning algorithms offer a route through which to achieve this whilst satisfying privacy constraints. Federated learning is challenging in practice as modern datasets are often inhomogeneously partitioned in terms of both the quantity of data at each client and the distribution of data at each client, the effects of which we describe in \Cref{sec:appen:inhomogeneity}. Consider the differences in data collected by a teaching metropolis hospital and a rural community hospital, for example. Further, communication channels between clients can be unreliable and costly making structured communication protocols difficult to implement, and those clients may also have different computational resources. Developing efficient algorithms that are robust to such difficulties is an active area of research that has received growing attention from the machine learning community.



Of great importance in many real world applications is the ability to reason in the presence of uncertainty. Probabilistic machine learning provides a principled and elegant framework through which this can be achieved \citep{ghahramani2015probabilistic}.  However, performing inference is often infeasible in practice and demands the use of approximations. A widely used approach is variational inference (VI), in which an approximation to the posterior distribution over model parameters is obtained as the solution to an optimisation problem. Whilst widely applied to a range of popular machine learning models---including Gaussian process models, latent topic models and deep generative models---vanilla VI does not immediately extend to the federated learning setting as it demands access to the entire dataset or unbiased samples thereof. 

The goal of this paper is to develop a framework, termed partitioned variational inference (PVI), which generalises VI to the federated learning setting. PVI unifies a wealth of existing schemes, providing practitioners with the flexibility to develop unique methods whilst respecting the constraints of a federated setting. We briefly summarise the contributions of this paper, focusing on the suitability of PVI for federated learning and the viewpoint of PVI as a unification of local and global VI methods.

\subsection{Partitioned Variational Inference: Theoretical Contribution}
\label{sec:intro:pvi_theory}
The main theoretical contribution of this paper is to introduce PVI, described in \Cref{sec:pvi}, a framework through which VI can be performed in the federated setting. We generalise and derive new supporting theory (including PVI fixed-point optimisation and mini-batch approximation), and demonstrate that PVI satisfies a number of properties that make it an attractive choice for practitioners. 

\subsection{Unification}
\label{sec:intro:unification}
Whilst the extension of VI to the federated learning setting motivates the development of PVI, the framework can be viewed as a generalisation of pre-existing local and global VI methods. From this viewpoint, in \Cref{sec:related-work} PVI is used to connect a large literature that has become fragmented with separated strands of related, but mutually uncited work. More specifically we unify work on: online VI \citep{ghahramani:2000,sato:2001,broderick+al:2013,bui+al:2017b,nguyen+al:2018}; global VI \citep{sato:2001,hensman+al:2012,hoffman+al:2013,salimans+knowles:2013,sheth+khardon:2016,sheth+al:2015,sheth+khardon:2016b}; local VI \citep{knowles+minka:2011,wand:2014,khan+li:2018}; power EP and related algorithms \citep{minka:2001, minka:2004, li+al:2015,hasenclever+al:2017,gelman+al:2014}; and stochastic mini-batch variants of these algorithms \citep{hoffman+al:2013,li+al:2015,khan+li:2018}. \Cref{fig:pvi_special_cases,fig:past_work} and \Cref{table:past-work} present a summary of these relationships in the context of PVI.

\subsection{Partitioned Variational Inference: Empirical Contribution}
\label{sec:intro:empirical}
In \Cref{sec:experiments}, we provide extensive experimentation showing that PVI is able to recover the global VI solution for logistic regression models and Bayesian neural networks in the presence of varying degrees and forms of inhomogeneous data partitions. We outline the inadequacies of existing approaches for performing approximate inference in the federated learning setting, highlighting how these shortcomings are addressed by the PVI framework. The experiments shed light on the settings in which different implementations of PVI perform well, providing guidance to practitioners.

\section{Partitioned Variational Inference}
\label{sec:pvi}
In this section, we introduce partitioned variational inference, a framework that extends VI to the federated learning setting as well as encompassing many other approaches to VI. We begin by framing PVI in terms of a series of variational free-energy optimisation problems performed locally by individual clients, proving several key properties of the algorithm that reveal the relationship to global VI. We assume the presence of a central server which coordinates communication between clients; however, it is important to note that a central server is not a requirement provided each client is able to communicate with all others. Skeleton code for the PVI algorithm is provided in \Cref{alg:vmp}, with different choices for step \blackcirc{1} yielding three variants of PVI with different properties.
In order to keep the development clear, we have separated most of the discussion of related work into \Cref{sec:related-work}. 

\subsection{Setup}
\label{sec:pvi:setup}
Consider the task of modelling a dataset $\bfy$ partitioned into $M$ local datasets $\{\bfy_m\}_{m=1}^M$, one for each of $M$ clients, with a parametric probabilistic model defined by the prior $p(\*\theta|\*\epsilon)$ over parameters $\*\theta$ and the likelihood function $p(\bfy|\*\theta,\*\epsilon) = \prod_{m=1}^{M}p(\bfy_m|\*\theta,\*\epsilon)$. For simplicity, we assume for the moment that the hyperparameters $\*\epsilon$ are fixed and suppress them to lighten the notation. We will discuss hyperparameter optimisation in \Cref{sec:hyper_learning}.

Exact Bayesian inference in this class of models is in general intractable, so we resort to VI, in which we find the member of a family of tractable distributions $\mcQ$ which minimising the KL-divergence between itself and the true posterior distribution.
\begin{equation}
    q^*(\*\theta) = \argmin_{q(\*\theta) \in \mcQ} \KL{q(\*\theta)}{p(\*\theta | \bfy)}. \nonumber
\end{equation}
This is equivalent to maximising the global (negative) free-energy $\mcF(q(\*\theta))$, which serves as a lower bound to the log marginal likelihood $\log p(\bfy)$:
\begin{equation}
    \begin{aligned}
    q^*(\*\theta) &= \argmax_{q(\*\theta) \in \mcQ} \underbrace{\Exp{q(\*\theta)}{\log p(\bfy | \*\theta)} - \KL{q(\*\theta)}{p(\*\theta)}}_{\mcF(q(\*\theta))} \\
    &= \argmax_{q(\*\theta) \in \mcQ} \log p(\bfy) - \KL{q(\*\theta)}{p(\*\theta | \bfy)}. \nonumber
    \end{aligned}
\end{equation}
Maximisation of the global free-energy therefore jointly obtains an approximation to the posterior, $q^*(\*\theta) \approx p(\*\theta | \bfy)$, and an approximation to the log marginal likelihood, $\mcF(q^*(\*\theta)) \approx \log p(\bfy)$, two quantities of interest in probabilistic machine learning. See \cite{wainwright2008graphical} and \cite{zhang2018advances} for two excellent reviews of VI. Throughout this paper, we refer to maximisation of the global free-energy as global VI. Unfortunately, evaluating the global free-energy directly requires access to the entire dataset, which violates the core constraint of federated learning. Rather than considering a single, global distribution, PVI decomposes the variational approximation in a manner that mimics the decomposition of the true posterior,
\begin{equation}
\begin{aligned}
    p(\*\theta | \bfy) &= \frac{1}{\mathcal{Z}} p(\*\theta) \prod_{m=1}^{M} p(\bfy_m|\*\theta) \\
    &\approx \frac{1}{\mcZ_q} p(\*\theta) \prod_{m=1}^{M} t_m(\*\theta) = q(\*\theta), \label{eq:approx}
\end{aligned}
\end{equation}
where $\mcZ_q$ is the normalising constant of the approximate posterior and $\mcZ = p(\bfy)$ is the normalising constant of the true posterior. $\mcZ$ is also known as the marginal likelihood. This decomposition is motivated by the fact that each client has access local data $\bfy_m$, which shall be used to refine the approximate likelihood $t_m(\*\theta)$ such that it closely approximates the effect the true likelihood term $p(\bfy_m | \*\theta)$ has on the posterior.
%
%
As we typically restrict the approximate likelihoods to lie within the same exponential family as the prior, computing the normalisation constant $\mcZ_q$ is straightforward and is generally not required unless an approximation to the log marginal likelihood is desired (see \Cref{prop:local-fe-global-fe}). In \Cref{sec:pvi:pvi_flavours}, we demonstrate that for one variant of PVI the product $p(\*\theta) \prod_{m=1}^{M} t_m(\*\theta)$ is automatically normalised ($\mcZ_q = 1$) regardless of the form of the approximate likelihoods. Finally, we will show in \Cref{sec:pvi:properties} that PVI will return an approximation to the log marginal likelihood $\log \mathcal{Z} = \log p(\bfy)$ in addition to the approximation of the posterior.

\subsection{PVI Skeleton}
\label{sec:pvi:skeleton}
\Cref{alg:vmp} details the PVI algorithm. We describe each of the three steps in detail below.
\begin{enumerate}[label=\protect\blackcirc{\arabic*}]
    \item (\emph{Server: client selection step.}) At each iteration $i$, a central server selects a set of approximate likelihoods $\{t^{(i-1)}_k(\*\theta)\}_{k \in b_i}$ to refine according to a schedule $b_i \subseteq \{1, \ldots M\}$.
    
    \item (\emph{Client: update step.}) Each approximate factor $t_k(\*\theta)$ is updated by communicating the most recent approximate posterior $q^{(i-1)}(\*\theta)$ to each client $k \in b_i$, each of which then solves the local optimisation problem
    \begin{equation}
        \label{eq:vfe}
        q^{(i)}_k(\*\theta) = \argmax_{q(\*\theta) \in \mcQ} \underbrace{\int q(\*\theta) \log \frac{q^{(i-1)}(\*\theta)p(\bfy_k | \*\theta)}{q(\*\theta)t^{(i-1)}_k(\*\theta)} \mathrm{d}\*\theta}_{\mcF^{(i)}_{k}(q(\*\theta))}
    \end{equation}
    where the optimisation is over a tractable family $\mcQ$. We refer to $\mcF^{(i)}_{k}(q(\*\theta))$ as the local (negative) variational free-energy for client $k$. Given $q^{(i)}_k(\*\theta)$, the updated approximate likelihood is found by division
    \begin{equation}
        t_k^{(i)}(\*\theta) \propto \frac{q^{(i)}_k(\*\theta)}{q^{(i-1)}(\*\theta)} t^{(i-1)}_k(\*\theta). \nonumber
    \end{equation}
    The change in approximate likelihoods, $\Delta^{(i)}_k(\*\theta) \propto \frac{t^{(i)}_k(\*\theta)}{t^{(i-1)}_k(\*\theta)}$, is communicated back to the server.
    
    \item (\emph{Server: update step.}) Finally, the approximate posterior is updated as
    \begin{equation}
        q^{(i)}(\*\theta) \propto q^{(i-1)}(\*\theta) \prod_{k \in b_i} \Delta^{(i)}_k(\*\theta). \nonumber
    \end{equation}
\end{enumerate}

Through modification of the update schedule executed by the server, three variants of PVI can be implemented: 
\begin{enumerate}
    \item \textit{sequential}, in which a single approximate likelihood is refined at each iteration;\footnote{The order in which the approximate likelihoods are updated can make a difference in the convergence of PVI. This is discussed in more detail in \Cref{sec:pvi:pvi_flavours}.}
    \item \textit{synchronous}, in which all approximate likelihoods are refined at each iteration;
    \item \textit{asynchronous}, in which approximate likelihoods are updated whenever clients become available for performing local computation. 
\end{enumerate}

\begin{algorithm}[tb]
   \caption{Partitioned Variational Inference}
   \label{alg:vmp}
    \begin{algorithmic}
    \STATE {\bfseries Input:} data partition $\{ \bfy_1, \ldots, \bfy_M \}$, prior $p(\*\theta)$. \\[5pt]
    \STATE Initialise:
    \begin{align*}
        t^{(0)}_m(\*\theta) &\defeq 1 \text{ for all } m = 1, 2, \ldots, M. \\
        q^{(0)}(\*\theta) &\defeq p(\*\theta).
    \end{align*}
    \FOR{$i=1,2,\ldots$ until convergence}
        \STATE \blackcirc{1} \emph{Server: client selection step.}
        \STATE $b_i \defeq$ set of indices of the next approximate likelihoods to refine. \\
        \STATE Communicate $q^{(i-1)}(\*\theta)$ to each client in $b_i$. \\
        \STATE ~\\
        \STATE \blackcirc{2} \emph{Client: update step.}
        \FOR{$k \in b_i$}
            \STATE Compute the new local approximate posterior:
            \begin{equation}
                q^{(i)}_k(\*\theta) \defeq \argmax_{q(\*\theta) \in \mathcal{Q}} \int q(\*\theta) \log \frac{q^{(i-1)}(\*\theta) p(\bfy_{k}|\*\theta)}{q(\*\theta) t^{(i-1)}_{k}(\*\theta)} \mathrm{d}\*\theta. \nonumber
            \end{equation}
            \STATE Update the approximate likelihood:
            \begin{align}
                t^{(i)}_{k}(\*\theta) &\propto \frac{q^{(i)}_k(\*\theta)}{q^{(i-1)}(\*\theta)} t^{(i-1)}_{k}(\*\theta) \nonumber
            \end{align}
            \STATE Communicate
            $\Delta^{(i)}_k(\*\theta) \propto \frac{t^{(i)}_k(\*\theta)}{t^{(i-1)}_k(\*\theta)}$
            to server.
        \ENDFOR
        \STATE ~\\
        \STATE \blackcirc{3} \emph{Server: update step.}
        \begin{equation}
            q^{(i)}(\*\theta) \propto q^{(i-1)}(\*\theta) \prod_{k \in b_i} \Delta^{(i)}_k(\*\theta).
            \nonumber
        \end{equation}
    \ENDFOR
\end{algorithmic}
\end{algorithm}

\subsection{PVI Properties}
\label{sec:pvi:properties}
The three implementations, alongside their strengths and weaknesses, are described in more detail in \Cref{sec:pvi:pvi_flavours}. Before doing so, we shall state three properties of PVI, derived in \Cref{sec:appen:proofs}: 1) the local free-energy optimisation is equivalent to a variational KL optimisation; 2) the sum of local free-energies is equal to the global free-energy; and 3) any fixed point of the algorithm is also a local optimum of global VI. These properties apply for general $\mcQ$, and are not limited to the exponential family.\footnote{However, we will only consider exponential family approximations in the experiments in \Cref{sec:experiments}.}


\begin{property}
    \label{prop:local-fe-opt}
    Maximising the local free-energy $\mathcal{F}^{(i)}_k(q(\*\theta))$ is equivalent to the  KL optimisation
    \begin{align}
        q^{(i)}_k(\*\theta) = \argmin_{q(\*\theta) \in \mathcal{Q}} \mathrm{KL} \left( q(\*\theta) ~\|~ \widehat{p}^{(i)}_k(\*\theta) \right), \nonumber
    \end{align}
    where $\widehat{p}^{(i)}_k(\*\theta) = \frac{1}{\widehat{\mathcal{Z}}^{(i)}_k} \frac{q^{(i-1)}(\*\theta)}{t^{(i-1)}_{k}(\*\theta)} p(\bfy_{k}|\*\theta) = \frac{1}{\widehat{\mathcal{Z}}^{(i)}_k} p(\*\theta) p(\bfy_{k}|\*\theta) \prod_{m \notin k } t^{(i-1)}_{m}(\*\theta) $ is known as the tilted distribution in the expectation propagation (EP, \citealp{minka:2001}) literature and is intractable. 
\end{property}

The proof is straightforward (see \Cref{sec:appen:local-FE-KL}). The tilted distribution can be justified as a sensible target as it removes the approximate likelihood $t^{(i-1)}_{k}(\*\theta)$ from the current approximate posterior and replaces it with the true likelihood $p(\bfy_{k}|\*\theta)$. In this way, the tilted distribution comprises one true likelihood, $M-1$ approximate likelihoods and the prior. The KL optimisation then incorporates the true likelihood's effect, in the context of the approximate likelihoods and the prior. 

\begin{property}
    \label{prop:local-fe-global-fe}
    Let $q(\*\theta) = \frac{1}{\mcZ_q} p(\*\theta) \prod_{m=1}^M t_m(\*\theta)$, $\mcF_m(q_m(\*\theta)) = \int q(\*\theta) \log \frac{q(\*\theta)p(\bfy_m | \*\theta)}{q_m(\*\theta) t_m(\*\theta)} \mathrm{d}\*\theta$ be the local free-energy for client $m$ and $\mcF(q(\*\theta)) = \int q(\*\theta) \log \frac{p(\*\theta)\prod_{m=1}^M t_m(\*\theta)}{q(\*\theta)} \mathrm{d}\*\theta$ be the global free-energy. We have
    \begin{align}
        \sum_{m=1}^M \mathcal{F}_m(q(\*\theta)) + \log \mcZ_{q} = \mathcal{F}(q(\*\theta)) \nonumber
    \end{align}
    i.e.\ the sum of the local free-energies and normalisation constant is equal to the global free-energy.
\end{property}

The proof is also straightforward (see \Cref{sec:appen:local-global-fe}). The importance of this result is that we can now perform efficient and principled hyperparameter optimisation, which we describe in \Cref{sec:hyper_learning}. This additional utility of PVI differentiates it from alternative schemes such as EP which, in general, do not provide a lower bound on the log marginal likelihood and so cannot be used for principled hyperparameter optimisation.

\begin{property}
\label{prop:fixed-point}
\sloppy Let $q^*(\*\theta) = \frac{1}{\mcZ_{q^*}} p(\*\theta) \prod_{m=1}^{M} t^*_m(\*\theta)$ be a fixed point of \Cref{alg:vmp}, $\mathcal{F}_m(q(\*\theta)) = \int q(\*\theta) \log \frac{q^*(\*\theta) p(\bfy_m|\*\theta)}{q(\*\theta) t^*_m(\*\theta)} \mathrm{d}\*\theta$ be the local free-energy for client $m$, and $\mathcal{F}(q(\*\theta)) = \int q(\*\theta) \log \frac{p(\*\theta) \prod_{m=1}^{M} p(\bfy_m|\*\theta)}{q(\*\theta)} \mathrm{d}\*\theta$ be the global free-energy. We have
\begin{equation}
    q^*(\*\theta) = \argmax_{q(\*\theta) \in \mathcal{Q}} \mathcal{F}_m(q(\*\theta)) \ \forall m \quad \implies \quad q^*(\*\theta) = \argmax_{q(\*\theta) \in \mathcal{Q}}  \mathcal{F}(q(\*\theta)) \nonumber
\end{equation}
i.e.~a PVI fixed point $q^*(\*\theta)$ is an optimum of global VI.
\end{property}

This result is more complex to show, but can be derived by computing the derivative and Hessian of the global free-energy and substituting into these expressions the derivatives and Hessians of the local free-energies (see \Cref{sec:appen:local-global}). The fact that the fixed point of PVI recovers a global VI solution (both the optimal $q(\*\theta)$ and the global free-energy) is the main theoretical justification for employing PVI. However, we do not believe that there is a Lyapunov function for PVI, indicating that it may potentially oscillate or diverge. Although in practice we found that damping prevented this behaviour, if convergence is required then, in theory, the double-loop algorithm of \cite{heskes2002expectation}, later developed by \cite{hasenclever+al:2017}, can be employed which guarantees convergence for power EP \citep{minka:2004} which, as we shall discuss in \Cref{sec:related-work}, can recover the solution of PVI. In practice, running the double-loop algorithm to convergence is slow and the inner-loop is often truncated, typically to just a single iteration. This removes any guarantee of convergence.\footnote{In fact, running the inner-loop algorithm of the double-loop algorithm for a single iteration recovers the original power EP algorithm. We do not believe this to be widely known by the research community.}







\begin{figure}[!ht]
    \centering
    \includegraphics[width=\textwidth]{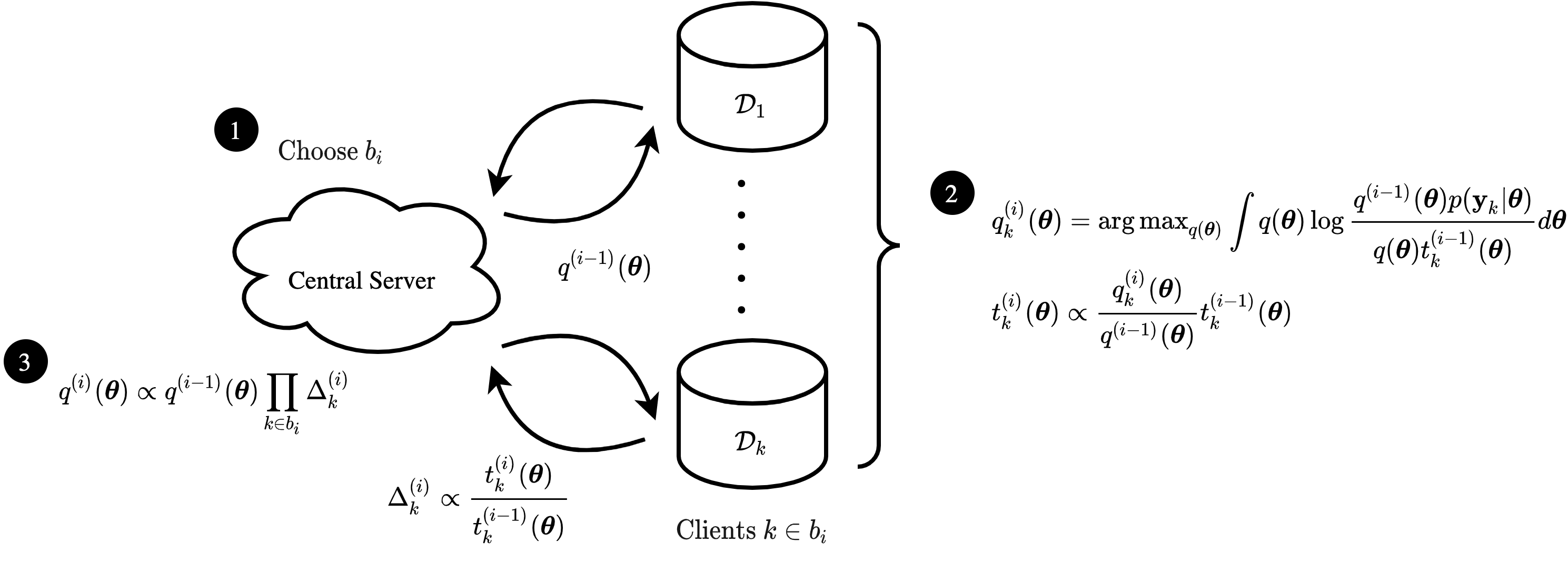}
    \caption{Steps of the PVI algorithm.}
    \label{fig:pvi_illustration}
\end{figure}

\subsection{The Three Implementations of PVI}
\label{sec:pvi:pvi_flavours}
We now describe the three implementations of PVI described in \Cref{sec:pvi:skeleton}, their strengths and weaknesses, and the settings in which they are most suitable. 

\subsubsection{Sequential PVI}
In sequential PVI, a single approximate likelihood is refined at each iteration. Sequential PVI enjoys the property that the update for the posterior $q^{(i)}(\*\theta)$ is automatically normalised when we set $t^{(i)}_k = \frac{q^{(i)}_k(\*\theta)}{q^{(i-1)}(\*\theta)} t^{(i-1)}_{k}(\*\theta)$ and $\Delta^{(i)}_k(\*\theta) = \frac{t^{(i)}_k(\*\theta)}{t^{(i-1)}_k(\*\theta)}$. This allows us to be more flexible with our choice of approximate likelihoods as we do not need to restrict them such that the normalisation constant can be computed. The proof is straightforward:
\begin{equation}
    \begin{aligned}
        q^{(i)}(\*\theta) &= p(\*\theta) \prod_{m=1}^M t^{(i)}_m(\*\theta) = t^{(i)}_{k}(\*\theta) p(\*\theta) \prod_{m\neq k}t_m^{(i-1)}(\*\theta) \\
        &= \frac{q^{(i)}_{k}(\*\theta)}{q^{(i-1)}(\*\theta)} \underbrace{p(\*\theta) \prod_{m=1}^M t_m^{(i-1)}(\*\theta)}_{q^{(i-1)}(\*\theta)} = q^{(i)}_{k}(\*\theta). \nonumber
    \end{aligned}
\end{equation}
Since $q^{(i)}_{k}(\*\theta)$ is normalised, so too is $q^{(i)}(\*\theta)$.

Damped updates are not required for sequential PVI, making it communication efficient as damping often leads to a greater number of smaller updates to attain convergence.\footnote{The total number of communications is taken to be the number of times clients are communicated with, rather than the number of times the server sends out the current approximate posterior.} This offers an important advantage when the cost of communication is large. However, as $K-1$ clients have to wait whilst a single approximate likelihood is updated at each iteration, sequential PVI exhibits poor time efficiency. Further, whilst we found sequential PVI to perform well in the homogeneous setting, we found that in the presence of a large number of inhomogeneous data partitions sequential PVI can struggle to converge without damping. We believe that inhomogeneous settings generally require smaller updates to be made per communication round so that all the approximate factors can come to an agreement, which damping achieves.

The order in which approximate likelihoods are updated will affect the rate at which sequential PVI converges. The simplest choice is to use a fixed, pre-defined order, which is common when implementing sequential variants of EP or belief propagation (BP) \citep{pearl1988probabilistic}. Although easy to implement, this is unlikely to achieve the fastest convergence possible. \cite{elidan2012residual} propose an alternative scheme for sequential BP, in which the order in which messages are updated depends on the magnitude of the previous update. They demonstrate empirically that their residual BP algorithm achieves faster convergence than the fixed-order baseline. A similar scheme for sequential PVI could result in improvements in the rate of convergence. However, it is not immediately clear which metric to use to measure the magnitude of the previous update. Simple vector norms may not be applicable when many of the parameters have little to no effect on the predictive distribution, such as in neural network models. 

Finally, we note that if each client is visited once then sequential PVI is equivalent to online/continual learning. We discuss this relationship in more detail in \Cref{sec:online}.

\subsubsection{Synchronous PVI}
In synchronous PVI, all approximate likelihoods are updated simultaneously at each iteration. Unlike the sequential case, the aggregation of approximate likelihoods is not automatically normalised. This typically means that we are restricted to using approximate likelihoods that lie in the same exponential family as the prior, so that the normalisation constant can be computed. A second limitation of synchronous PVI is that the product aggregation of approximate likelihoods may not be normalisable. To see this, consider the updated approximate posterior $q^{(i)}(\*\theta)$:
\begin{equation}
    \begin{aligned}
        q^{(i)}(\*\theta) &\propto p(\*\theta) \prod_{m=1}^M t_m^{(i)}(\*\theta) \\
        &\propto p(\*\theta) \prod_{m=1}^M \frac{q^{(i)}_m(\*\theta)}{q^{(i-1)}(\*\theta)} t^{(i-1)}_m(\*\theta) \\
        &= \left(\frac{1}{q^{(i-1)}(\*\theta)}\right)^M \underbrace{p(\*\theta) \left(\prod_{m=1}^M t^{(i-1)}_m(\*\theta)\right)}_{\propto q^{(i-1)}(\*\theta)}\left(\prod_{m=1}^M q^{(i)}_k(\*\theta)\right) \\
        &\propto \frac{\prod_{m=1}^M q^{(i)}_m(\*\theta)}{\left[q^{(i-1)}(\*\theta)\right]^{K-1}}. \nonumber
    \end{aligned}
\end{equation}
This bears close resemblance to the update scheme of the Bayesian committee machine (BCM, \citealp{tresp2000bayesian}). There is no guarantee that the update is normalisable, which results in an improper distribution for $q^{(i)}(\*\theta)$. To overcome this, we use damping of the approximate likelihoods: $t_k^{(i)}(\*\theta) \propto \left(\frac{q^{(i)}_k(\*\theta)}{q^{(i-1)}(\*\theta)}\right)^{\rho} t^{(i-1)}_k(\*\theta)$ for some $\rho \in (0, 1]$. We found choosing $\rho \propto \frac{1}{M}$ results in stable convergence. 

The use of damping makes synchronous PVI less communication efficient than sequential PVI, particularly in the case of a large number of clients when more damping is necessary. Yet, synchronous PVI can exhibit better time efficiency as all approximate likelihoods are updated in parallel.\footnote{Although not as time efficient as one might hope due to the need for damping. In fact, for homogeneous data partitions, choosing $\rho \propto \frac{1}{M}$ means both sequential and synchronous PVI effectively update a single approximate likelihood at each iteration, thus having similar time efficiencies.} We found synchronous PVI to be particularly effective in the presence of inhomogeneous data partitions. In general, the PVI approximate posterior will be biased towards the most recently visited dataset. However, as synchronous PVI updates the posterior by considering the entire dataset, it performs well even when partitions are inhomogeneous. In contrast, sequential PVI can perform poorly in the presence of inhomogeneous partitions, and can even fail to converge when $M$ is large unless damping is used.
 
\subsubsection{Asynchronous PVI}   
Rather than waiting for all clients to finish their local free-energy optimisation, in asynchronous PVI an updated approximate likelihood is integrated into the approximate posterior as soon as it becomes available. The corresponding client then receives the updated posterior from the server and resumes local free-energy optimisation. This is particularly useful when communicating over an unreliable channel, or when the update-frequencies are highly inhomogeneous due to inhomogeneous hardware or dataset sizes. As the approximate posterior is changed whilst clients perform local free-energy optimisation, the returned approximate likelihood may be stale given the update context. This also means that the updated approximate posterior may be improper, necessitating the use of damping.\footnote{We found a similar damping factor to that used in synchronous PVI was required for asynchronous PVI.} 

Out of all three PVI implementations, the communication efficiency of asynchronous PVI can be expected to be the worst.\footnote{To see why asynchronous PVI can be less communication efficient than synchronous PVI, consider the case in which a single client has significantly less data than the others. The small data client will finish its local optimisation faster than the others and, assuming clients communicate as soon as their local optimisation converges, communicate its update approximate likelihood back to the server. In synchronous PVI, the small client must wait until the other large data clients have finished their optimisation before proceeding with its next local optimisation. However, in asynchronous PVI the small client can perform its next local optimisation immediately---this optimisation is redundant as it will have received no information from any of the large data clients. Thus, unnecessary communications can be performed.} Yet, asynchronous PVI can be expected to be the most time-efficient.

\subsubsection{Summary}
Out of the three PVI implementations, sequential PVI offers the most theoretically appealing properties and is the most communication efficient and, as no damping is required, it is also often time efficient. However, due to the sequential visiting of data partitions and the accumulation of approximation errors, sequential PVI can struggle to converge when data partitions are inhomogeneous. Further, the ordering in which partitions are visited is likely to effect the rate of convergence, yet it is unclear which ordering to use. Synchronous PVI is in theory the most time efficient; however, it requires careful tuning of the damping factor to ensure this efficiency is realised without encountering improper distributions. Asynchronous PVI is the least communication efficient and requires damping, yet can be the most-time efficient and is the only implementation which can be deployed without modification in settings where communication between clients is unreliable.

\section{Client Update Step: Approaches for Optimising the Local Free-energies}
\label{sec:optim}
Having laid out the general framework for PVI, what remains to be decided is the method used for optimising the local free-energies at each client. We consider three choices: analytic updates, off-the-shelf optimisation methods and fixed-point iterations, as well as discussing how stochastic approximations can be combined with these approaches. 

\subsection{Analytic Local Free-energy Updates}
\label{sec:optim:analytic}
Each local free-energy $\mcF^{(i)}_k(q(\*\theta))$ is equivalent in form to a global free-energy with an effective prior $p_{\mathrm{eff}}(\*\theta) \propto q^{(i-1)}(\*\theta)/ t^{(i-1)}_{k}(\*\theta)$. Thus, in conjugate exponential family models the KL optimisations will be available in closed-form, for example in GP regression, and can be substituted back into the local variational free-energies to yield locally-collapsed bounds that are useful for hyperparameter optimisation \citep{bui+al:2017}.\footnote{This leads to a potential advantage of PVI over global VI in terms of speed of convergence: collapsed bounds can be leveraged on local datasets within each client where an application to the entire dataset would be computationally intractable.}

\subsection{Off-The-Shelf Optimisers for Local Free-Energy Optimisation}
\label{sec:optim:offtheshelf}
If analytic updates are not tractable, the local free-energy optimisations can be carried out using standard optimisers. 
If the partitions are small enough, non-stochastic optimisers such as BFGS be leveraged.\footnote{Again, this offers a potential computational advantage of PVI over global VI: non-stochastic optimisers can be deployed  in the large data setting by partitioning the data across multiple clients.} Of course, if a stochastic approximation like Monte Carlo VI is employed for each local optimisation, stochastic optimisers such as RMSProp \citep{tieleman+hinton:2012} or Adam \citep{kingma+ba:2014} might be more appropriate choices. In all cases, since the local free-energy is equivalent in form to a global free-energy with an effective prior $p_{\mathrm{eff}}(\*\theta) \propto q^{(i-1)}(\*\theta)/ t^{(i-1)}_{k}(\*\theta)$, PVI can be implemented via trivial modification to existing code for global VI. Whilst other local VI approaches, such as such as variational message passing \citep{winn+bishop:2005,winn+minka:2009,knowles+minka:2011}, can be deployed in the federated learning setting, they do not share the same flexibility and ease of implementation of PVI, instead requiring bespoke and closed-form updates for different likelihoods and cavity distributions.

For gradient-based updates, the training dynamics of global VI can be reproduced by synchronous PVI if a single local update is performed synchronously by all clients:
\begin{property}
    \label{prop:local-global-gradient}
    Terminating the local optimisation step in synchronous PVI after a single gradient-based update results in identical dynamics as global VI for $q(\*\theta)$, given by the following equation, regardless of the partition of the data:
    \begin{equation}
        \*\eta^{(i)}_q = \*\eta^{(i-1)}_q + \rho  \left.\frac{\mathrm{d}}{\mathrm{d} \*\eta_q} \mcF(\*\eta_q) \right|_{\*\eta_q = \*\eta^{(i-1)}_q} \nonumber
    \end{equation}
    where $\mcF(\*\eta_q) = \sum_{n=1}^N \Exp{q}{\log p(y_n | \*\theta)} - \KL{q(\*\theta)}{p(\*\theta)}$ is the global free-energy.
\end{property}
See \Cref{sec:local-global-gradient} for the proof. This demonstrates a further equivalence between PVI and global VI: employing a single iteration of gradient based updates for synchronous PVI is the same as employing gradient based updates for global VI.\footnote{It should be noted that global (batch) VI retains only a single set of global natural parameters, and thus is more memory efficient than PVI which retains $M$ sets of local natural parameters.} The equivalence provides us with an alternative method of reproducing the dynamics of global VI in the federated learning setting.\footnote{The alternative method which recovers global VI with biased mini-batches is to communicate $N = \sum_{k=1}^{K} N_k$ and $p(\*\theta)$ to each client, and have each client sequentially update the approximate posterior by performing a single/several optimisation step(s) the modified free-energy $\mcF_k(q(\*\theta)) = \frac{N_k}{N}\Exp{q(\*\theta)}{\log p(\bfy_k | \*\theta)} - \KL{q(\*\theta)}{p(\*\theta)}$. This method, however, requires clients to release their number of datapoints $N_k$.} The fact that PVI supports multiple optimisation steps per communication is key to achieving communication efficiency. In most settings performing more than a single optimisation step improves communication efficiency, although in highly inhomogeneous settings performing fewer optimisation steps might be advantageous as this has an effect comparable to damping.

It is important to stress that if the local optimisation step is run to convergence then the partition of the data does influence the dynamics for $q(\*\theta)$, and the dynamics of global VI are not recovered---if they were, PVI would offer no improvement in communication efficiency over this federated implementation of global VI.

\subsection{Local Free-Energy Fixed Point Updates, Natural Gradient Methods, and Mirror Descent}
\label{sec:optim:localfixedpoint}
An alternative to using off-the-shelf optimisers is to derive fixed-point update equations by zeroing the gradients of the local free-energy. These fixed-point updates have elegant properties for approximate posterior distributions that are in the exponential family.
\begin{property}
\label{prop:fixed-point-equations}
Let the prior and approximate likelihood factors be in the un-normalised exponential family $t_m(\*\theta) = t_m(\*\theta ; \*\eta_m) = \exp(\*\eta_m^\intercal T(\*\theta))$ so that the variational distribution is in the normalised exponential family $q(\*\theta) = \exp( \*\eta_q^\intercal T(\*\theta) - A(\*\eta_q) )$. Let $q^{(i)}_k(\*\theta)$ be the stationary point of the local free-energy such that $\left.\frac{\mathrm{d}\mathcal{F}_k^{(i)}(q(\*\theta))}{\mathrm{d}\*\eta_q}\right|_{q = q^{(i)}_k}  = \*0$, then, noting that the update step $t^{(i)}_k(\*\theta) = \frac{q^{(i)}_k(\*\theta)}{q^{(i-1)}(\*\theta)} t^{(i-1)}_k(\*\theta)$ implies $\*\eta^{(i)}_k = \*\eta^{(i)}_{q_k} - \*\eta^{(i-1)}_q + \*\eta^{(i-1)}_k$,
\begin{align}
\*\eta^{(i)}_{k}  &= \left.\mathbb{C}^{-1} \frac{\mathrm{d}}{\mathrm{d}\*\eta_q} \mathbb{E}_q \left[\log p(\bfy_{k}|\*\theta)\right]\right|_{q = q^{(i)}_k} \nonumber
\end{align}
where $\mathbb{C} \defeq \frac{\mathrm{d}^2A(\*\eta_q)}{\mathrm{d}\*\eta_q\mathrm{d}\*\eta_q} = \mathrm{cov}_{q(\*\theta)}[ T(\*\theta) T^\intercal(\*\theta)]$ is the Fisher Information. Moreover, the Fisher Information can be written as $\mathbb{C} = \frac{\mathrm{d}\*\mu_q}{\mathrm{d}\*\eta_q}$ where $\*\mu_q = \mathbb{E}_q \left[T(\*\theta)\right]$ is the mean parameter of $q(\*\theta)$. Hence,
\begin{align}
\label{eq:fixed-point-bi-simplified}
\*\eta^{(i)}_{k}  &
= \left.\frac{\mathrm{d}}{\mathrm{d}\*\mu_q} \mathbb{E}_q \left[\log p(\bfy_{k}|\*\theta)\right]\right|_{q=q^{(i)}_k}.
\end{align}
For some approximate posterior distributions $q(\*\theta)$, taking derivatives of the average log-likelihood with respect to the mean parameters is analytic (e.g.~Gaussian\footnote{However, in the stochastic case where Monte Carlo samples are employed to estimate these quantities, it can reduce variance to explicitly estimate the Fisher information, even if the derivative with respect to the mean parameters is analytic, see \citet{salimans+knowles:2013}.}) and for some it is not (e.g.~gamma).
\end{property}
These conditions, derived in \Cref{sec:appen:fp}, can be used as fixed point equations. That is, they can be iterated possibly with damping $\rho$, 
\begin{align}
\label{eq:fixed-point-bi-damp}
\*\eta^{(i)}_{k} \leftarrow (1-\rho)\*\eta^{(i)}_{k} + \left.\rho \frac{\mathrm{d}}{\mathrm{d}\*\mu_q} \mathbb{E}_q \left[\log p(\bfy_{k}|\*\theta)\right]\right|_{q=q^{(i)}_k}.
\end{align}
These iterations, which form an inner-loop in PVI, are themselves not guaranteed to converge (there is no Lyapunov function in general and so, for example, the local free-energy will not reduce at every step).

The fixed point updates are the natural gradients of the local free-energy and the damped versions are natural gradient ascent \citep{sato:2001,hoffman+al:2013,khannielsen2018fast,khan2021bayesian}. The natural gradients could also be used in other optimisation schemes \citep{hensman+al:2012,salimbeni+al:2018}. The damped updates are also equivalent to performing mirror-descent \citep{raskutti+mukherjee:2015, khan+li:2018}, a general form of proximal algorithm \citep{parikh+boyd:2014} that can be interpreted as trust-region methods. For more details about the relationship between these methods, see \Cref{sec:nat-grad-mirror-trust}. 


For these types of updates, there is an equivalent relationship between synchronous PVI and global VI as \Cref{prop:local-global-gradient}:
\begin{property} \label{prop:local-global-fp}
Terminating the local optimisation step in synchronous PVI  after a single iteration of the damped fixed-point update \eqref{eq:fixed-point-bi-damp} results in identical dynamics as global VI for $q(\*\theta)$ given by the following equation, regardless of the partitioning of the data
\begin{equation}
\begin{aligned}
    \*\eta^{(i)}_q &= (1 - \rho) \*\eta^{(i-1)}_q + \rho \left( \*\eta_0 + \left.\frac{\mathrm{d}}{\mathrm{d}\*\mu_{q}} \Exp{q}{\log p(\bfy | \*\theta)}\right|_{q=q^{(i-1)}}\right) \\ 
    &= (1 - \rho) \*\eta^{(i-1)}_q + \rho \left( \*\eta_0 + \sum_{n=1}^N \left.\frac{\mathrm{d}}{\mathrm{d}\*\mu_{q}} \Exp{q}{\log p(y_{n}|\*\theta)}\right|_{q=q^{(i-1)}}\right). \nonumber
\end{aligned}
\end{equation}
\end{property}
See \Cref{sec:local-global-same} for the proof. This provides a method through which damped fixed-point updates of global VI can be performed whilst conforming to the constraints of federated learning. As in the case of gradient-based updates, PVI achieves improved communication efficiency relative to the federated implementation of global VI by iterating the fixed-point update equation on each client until convergence, before communicating the result back to the server.

\subsection{Mini-Batch Approximations and Hyperparameter Learning}
\label{sec:optim:minibatch_hypers}
In the case where each partition includes a large number of data points, stochastic approximations can lead to faster convergence of the local optimisation procedure. There are two different ways of splitting up the local partitions: 
\begin{enumerate*}
    \item fixed sub-partitions of data;
    \item randomly sampled mini-batches of data.
\end{enumerate*} 
\Cref{sec:stochastic-approx} compares and contrasts these two difference approaches using damped fixed-point updates.

Finally, many probabilistic models rely on an appropriate choice of the hyperparameters $\*\epsilon$ to achieve good predictive performance on a task. The PVI framework offers a principled method for both performing hyperparameter optimisation through maximisation of the approximate marginal likelihood and performing approximate inference over hyperparameters if a distributional estimate is desired. See \Cref{sec:hyper_learning} for details. 

\section{Unification of Previous Work}

\label{sec:related-work}
\begin{figure}[!ht]
\centering
\includegraphics[scale=1.25]{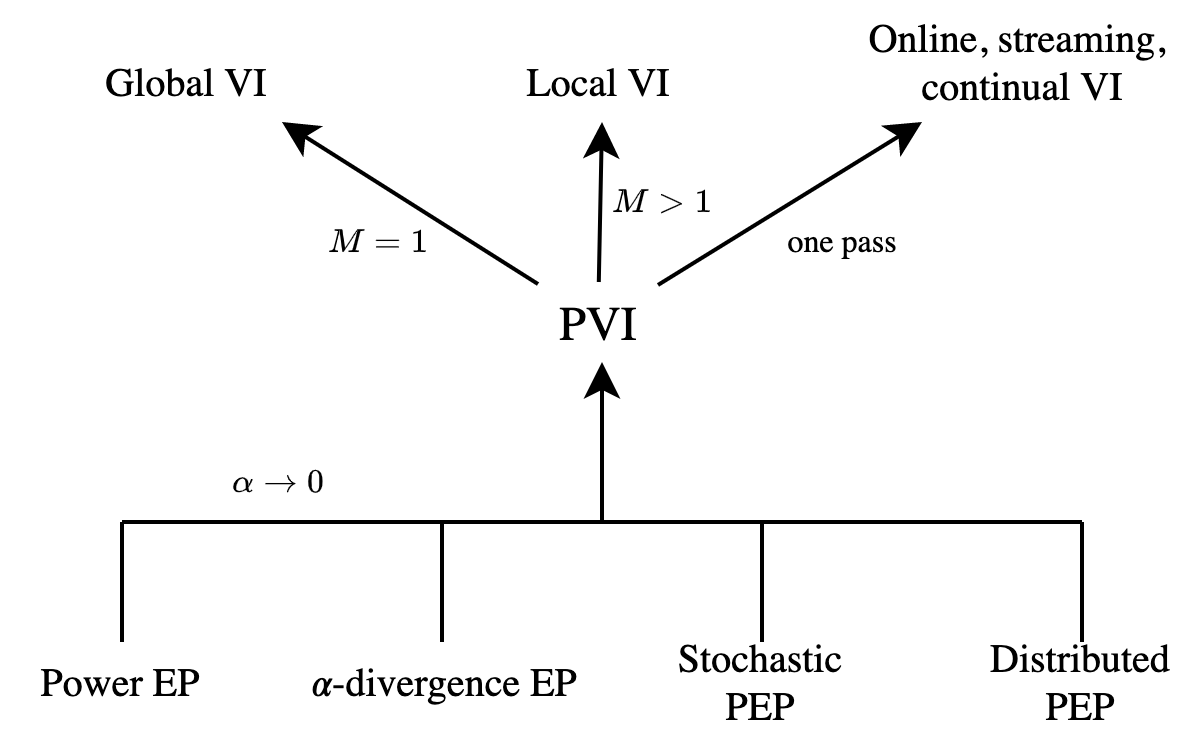}
\caption{Variational inference schemes encompassed by the PVI framework.}
\label{fig:pvi_special_cases}
\end{figure}

\begin{figure}[!ht]
\includegraphics[width=\linewidth]{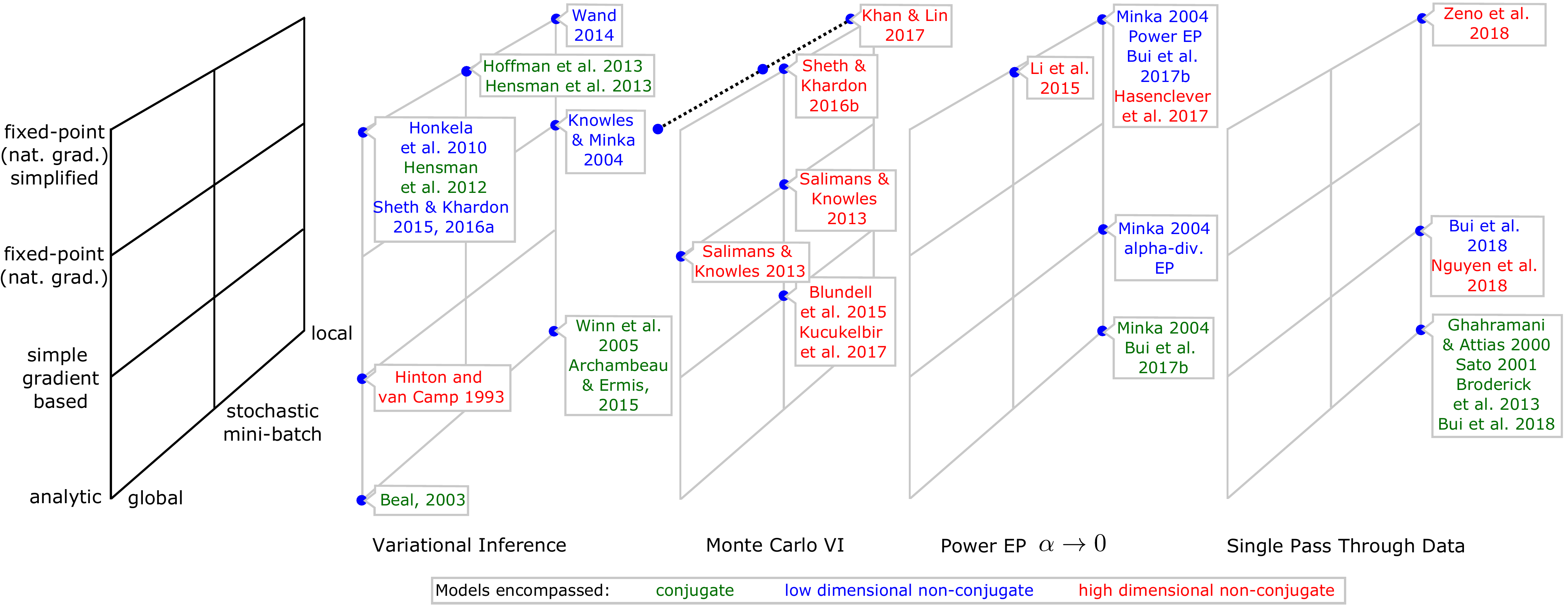}
\caption{The local VI framework unifies prior work. The granularity of the approximation and the optimisation method employed are two fundamental algorithmic dimensions that are shown as axes. Fixed-point updates are identical to natural gradient ascent with unit step size. The models encompassed by each paper are indicated by the color. See \Cref{table:past-work} for more information.}
\label{fig:past_work}
\end{figure}

The preceding sections have introduced PVI as a framework extending VI to the federated learning setting. However, if we depart from the federated learning setting and consider the data partitioning scheme to be a choice made by the practitioner, then PVI can more generally be interpreted as a unification of existing approaches to VI. These methods include global VI (\Cref{sec:global}), local VI (\Cref{sec:local}), online VI (\Cref{sec:online}) and a number of methods based on power EP (\Cref{sec:pep,sec:alpha_ep,sec:stochastic_pep,sec:distributed_pep}). A schematic showing the relationships between these methods at a high level is shown in \Cref{fig:pvi_special_cases}. The literature has been organised into \Cref{fig:past_work} and \Cref{table:past-work}. 

\subsection{Global VI Fixed-Point Methods}
\label{sec:global}

There has been a long history of applying the fixed-point updates for global VI (PVI where $M=1$). \citet{sato:2001} derived them for conjugate exponential family models, showing they recover the closed form updates for $q(\*\theta)$, and noting that damped fixed-point updates are equivalent to natural gradient ascent with unit step size ($\rho =1$). 
Sato's insight was subsequently built upon by several authors. \cite{honkela+al:2010} considered non-conjugate models, employed a Gaussian variational distribution and used natural gradients to update the variational distribution's mean.
\citet{hensman+al:2012} and \citet{hoffman+al:2013} applied the insight to conjugate models when optimising collapsed variational free-energies and deriving stochastic natural gradient descent, respectively. \citet{salimans+knowles:2013} apply the fixed-points to non-conjugate models where the expectations over $q(\*\theta)$ are intractable and use Monte Carlo to approximate them, but they explicitly calculate the Fisher information matrix.  
\citet{sheth+khardon:2016} and \citet{sheth+al:2015} treat non-conjugate models with Gaussian latent variables, employ the cancellation of the Fisher information, and analyse convergence properties. \citet{sheth+khardon:2016b} further extend this to two level-models through Monte Carlo essentially applying the Fisher information cancellation to \citet{salimans+knowles:2013}, but they were unaware of this prior work. 

\subsection{Local VI Fixed-Point Methods} 
\label{sec:local}
Whereas global VI seeks a single approximation to the full posterior distribution, local VI methods employ an approximate posterior distribution which decomposes as a product of factors, with each factor being iteratively refined. Typically, each factor involves only a subset of datapoints (PVI where $M > 1$) and may also include only a subset of variables over which the posterior is defined. Similar to global VI, there has been a long history of applying fixed-point updates in the local VI setting. \citet{knowles+minka:2011} derive them for non-conjugate variational message passing (VMP), but explicitly calculate the Fisher information matrix (except in a case where $q(\*\theta)$ was univariate Gaussian case where they do employ the cancellation). \citet{wand:2014} simplified VMP by applying the Fisher information cancellation to the case where $q(\*\theta)$ is multivariate Gaussian. 
The work is closely related to  \cite{salimans+knowles:2013} and \cite{sheth+khardon:2016b}, since although these papers use fixed-point updates for global VI, they show that these decompose over data points and thereby derive mini-batch updates that closely resemble fixed-point per-datapoint local VI. This is a result of \Cref{prop:local-global-fp}.

\cite{khan+li:2018} also extend VMP to employ Monte Carlo approximation and the Fisher information cancellation. They were unaware of \citet{wand:2014}, but extend this work by treating a wider range of approximate posteriors and models, stochastic updates, and a principled approach to damping.
Their approach starts from a global perspective, and uses the independence of likelihood terms to split up the posterior. Any partition of data can be used for their method (including $M = 1$), and natural-gradient updates must be used on any non-conjugate terms. If the mini-batches are selected randomly at each iteration, then a set of natural parameters for each datapoint must be stored. However, if the mini-batches are kept fixed across iterations then only a single set of natural parameters for each mini-batch needs to be maintained. In doing so, their method becomes exactly equivalent to PVI using fixed-point updates and terminating the local optimisation procedure after a single iteration. 


\subsection{Online, Streaming, Incremental and Continual VI as a Single Pass of PVI}
\label{sec:online}

If PVI makes a single sequential pass through the data, the approximate likelihoods do not need to be explicitly computed or stored as data partitions are not revisited. In this case PVI reduces to initialising the approximate posterior to be the prior, $q^{(0)}(\*\theta) = p(\*\theta)$  and then proceeds by optimising a sequence of local free-energies
\begin{align}
q^{(i)}(\*\theta) \defeq \argmax_{q(\*\theta) \in \mathcal{Q}} \int q(\*\theta) \log \frac{q^{(i-1)}(\*\theta) p(\bfy_{k}|\*\theta)}{q(\*\theta)} \mathrm{d}\*\theta.\nonumber
\end{align}
These have the form of standard variational inference with the prior replaced by the previous variational distribution $q^{(i-1)}(\*\theta)$. This idea---combining the likelihood from a new batch of data with the previous approximate posterior and projecting back to a new approximate posterior---underpins online variational inference \citep{ghahramani:2000,sato:2001}, streaming variational inference \citep{broderick+al:2013,bui+al:2017b}, and variational continual learning \citep{nguyen+al:2018,swaroop2019improving,loo2021generalized}. Early work on online VI used conjugate models and analytic updates \citep{ghahramani:2000,sato:2001,broderick+al:2013,bui+al:2017b}. This was followed by off-the-shelf optimisation approaches for non-conjugate models \citep{bui+al:2017b} and further extended to leverage Monte Carlo approximations of the local-free energy \citep{nguyen+al:2018}. Recently \citet{zeno+al:2018} use the variational continual learning framework of \citet{nguyen+al:2018}, but employ fixed-point updates instead. As these methods can be interpreted as finding the functional approximation $t^{(i)}(\*\theta) = \frac{q^{(i)}(\*\theta)}{q^{(i-1)}(\*\theta)}$ involving just a single datapoint or a group of datapoints, they too can be interpreted as local VI methods.

\subsection{Power EP as a Local VI Fixed-Point Method} 
\label{sec:pep}

There is also an important relationship between PVI methods employing fixed-point updates and power EP (PEP) \citep{minka:2004, minka:2005}. \Cref{prop:pep-fp} below states that the local VI fixed-point equations are recovered from the PEP algorithm as $\alpha \rightarrow 0$. 
\begin{property}
\label{prop:pep-fp}
The damped fixed-point equations are precisely those returned by the PEP algorithm, shown in \Cref{alg:pep}, in the limit as $\alpha \rightarrow 0$.
\end{property}
\begin{algorithm}[tb]
   \caption{One step of the PEP algorithm at the $i$-th iteration, for the $k$-th data partition}
   \label{alg:pep}
\begin{algorithmic}
    \STATE Compute the tilted distribution:
    $\hat{p}^{(i)}_{\alpha}(\*\theta) =  q^{(i-1)}(\*\theta) \left ( \frac{p(\mathbf{y}_{k} | \*\theta)}{t_{k}^{(i-1)}(\*\theta)}\right)^{\alpha}$.
    \STATE Moment match:
    $q_{\alpha}(\*\theta) = \mathrm{proj}( \hat{p}^{(i)}_{\alpha}(\*\theta) ) \; \text{such that} \; \mathbb{E}_{q(\*\theta)} \left[T(\*\theta)\right] = \mathbb{E}_{\hat{p}^{(i)}_{\alpha}(\*\theta)} \left[T(\*\theta)\right]$.
    \STATE Update the posterior distribution with damping $\rho$:
    $q^{(i)}(\*\theta) = \left ( q^{(i-1)}(\*\theta) \right )^{1-\rho/\alpha} \left( q_{\alpha}(\*\theta) \right)^{\rho/\alpha}$.
    \STATE Update the approximate likelihood:
    $t^{(i)}_{k}(\*\theta) = \frac{q^{(i)}(\*\theta)}{q^{(i-1)}(\*\theta)} t^{(i-1)}_{k}(\*\theta)$.
\end{algorithmic}
\end{algorithm}
Although we suspect \citet{knowles+minka:2011} knew of this relationship, and it is well known that  PEP has the same fixed-points as VI in this case, it does not appear to be widely known that variationally limited PEP yields exactly the same algorithm as fixed-point local VI. See \Cref{proof:pep-fp} for the proof.
 
\subsection{Alpha-Divergence EP as a Local VI Method With Off-The-Shelf Optimisation}
\label{sec:alpha_ep}

PVI is intimately related to alpha-divergence EP. If PVI's KL divergence is replaced by an alpha divergence $ \mathrm{D}_{\alpha} [p(\*\theta) || q(\*\theta) ] = \frac{1}{\alpha(1-\alpha)} \int \left [ \alpha p(\*\theta) + (1-\alpha) q(\*\theta) -  p(\*\theta)^{\alpha}  q(\*\theta)^{1-\alpha} \right] \mathrm{d}\*\theta$, we recover the alpha-divergence formulation of the PEP algorithm \citep{minka:2004, minka:2005}, which encompasses the current case as $\alpha \rightarrow 0$ and EP when $\alpha \rightarrow 1$ \citep{minka:2001}. The updates using this formulation are shown in \Cref{alg:pep_alpha}.
\begin{algorithm}[tb]
   \caption{One step of the PEP algorithm, as in \Cref{alg:pep}, but with alpha divergence minimisation}
   \label{alg:pep_alpha}
\begin{algorithmic}
    \STATE Compute the tilted distribution:
    $\hat{p}^{(i)}_k(\*\theta) =  q^{(i-1)}(\*\theta)  \frac{p(\mathbf{y}_{k} | \*\theta)}{t_{k}^{(i-1)}(\*\theta)}$.
    \STATE Find the posterior distribution:
    $q^{(i)}_k(\*\theta) \defeq \argmin_{q(\*\theta) \in \mathcal{Q}} \mathrm{D}_{\alpha} [\hat{p}^{(i)}_k(\*\theta) || q(\*\theta) ]$.
    \STATE Update the approximate likelihood:
    $t^{(i)}_{k}(\*\theta) = \frac{q^{(i)}_k(\*\theta)}{q^{(i-1)}(\*\theta)} t^{(i-1)}_{k}(\*\theta)$.
\end{algorithmic}
\end{algorithm}
The alpha divergence is typically very difficult to compute when more than one non-Gaussian likelihood is included in a data partition $\mathbf{y}_m$, meaning that for general alpha it would be appropriate to set $M=N$. The variational KL is the exception as it decomposes over data points. 

\subsection{Stochastic Power EP as a Stochastic Global VI Fixed-Point Method}
\label{sec:stochastic_pep}

The stochastic PEP (SPEP) algorithm \citep{li+al:2015} reduces the memory overhead of PEP by maintaining a single likelihood approximation that approximates the average effect a likelihood has on the posterior  $q(\*\theta) = p(\*\theta) t(\*\theta)^M $. Taking the variational limit of this algorithm, $\alpha \rightarrow 0$,  we recover global VI ($M=1$) with damped simplified fixed-point updates that employ a stochastic (mini-batch) approximation \citep{hoffman+al:2013}.
\begin{prop}
\label{prop:sep}
The mini-batch fixed-point equations are precisely those returned by the SPEP algorithm, shown in \Cref{alg:spep} in the limit that $\alpha \rightarrow 0$. 
\end{prop}
\begin{algorithm}[tb]
   \caption{One step of the SPEP algorithm at the $i$-th iteration, for the $k$-th data partition}
   \label{alg:spep}
\begin{algorithmic}
    \STATE Compute the tilted distribution:
    $\hat{p}^{(i)}_{k, \alpha}(\*\theta) =  q^{(i-1)}(\*\theta) \left ( \frac{p(\mathbf{y}_{k} | \*\theta)}{t^{(i-1)}(\*\theta)}\right)^{\alpha}$.
    \STATE Moment match:
    $q^{(i)}_{k, \alpha}(\*\theta) = \mathrm{proj}( \hat{p}^{(i)}_{k, \alpha}(\*\theta) )\; \text{such that} \; \mathbb{E}_{q^{(i)}_{k, \alpha}(\*\theta)} \left[T(\*\theta)\right] = \mathbb{E}_{\hat{p}^{(i)}_{\alpha}(\*\theta)} \left[T(\*\theta)\right]$.
    \STATE Update the posterior distribution with damping $\rho$:
    $$q^{(i)}_k(\*\theta) = \left ( q^{(i-1)}(\*\theta) \right )^{1-M \rho /\alpha} \left( q^{(i)}_{k, \alpha}(\*\theta) \right)^{M \rho/\alpha}.$$
    \STATE Update the approximate likelihood:
    $t^{(i)}_k(\*\theta) = \left (\frac{q^{(i)}_{k}(\*\theta)}{p(\*\theta)} \right)^{1/M}$.
\end{algorithmic}
\end{algorithm}
In this way the relationship between EP and SEP is the same as the relationship between fixed-point PVI and fixed-point mini-batch global VI (see \Cref{sec:stochastic-approx} where the two approaches differ by removing either an average natural parameter or a specific one). Similarly, if we altered PVI to maintain a single average likelihood approximation, as SEP does, we would recover mini-batch global VI.

\subsection{Distributed (Power) EP Methods} 
\label{sec:distributed_pep}

The convergent distributed PEP approach of \cite{hasenclever+al:2017} recovers a version of PVI as $\alpha \rightarrow 0$ with convergence guarantees. The PVI approach can be used in similar spirit to \citet{gelman+al:2014,xu2014distributed,hasenclever+al:2017} who use EP to split up datasets into small parts that are amenable to MCMC. Here PVI could be used to split up datasets so that they are amenable for optimisation.

\begin{table}[!ht]
        \setcounter{table}{0}
        \caption[]{Variational inference schemes encompassed by the PVI framework. (See next page.) 
        Selected past work has been organised into four categories: global VI (PVI with $M=1$), local VI (PVI with $M>1$), online VI and PEP variants. The citation to the work is provided along with the granularity of the method. 
        The optimisation used from the PVI perspective on this work is noted.  Abbreviations used here are: conjugate gradient (CG) and Monte Carlo (MC). The model class that the scheme encompasses is noted (conjugate versus non-conjugate) along with the specific models that the scheme was tested on. Model abbreviations are: non-linear state-space model (NSSM), non-linear factor analysis (NFA), latent Dirichlet allocation (LDA),  Poisson mixed model (PMM), heteroscedastic linear regression (HLR),  sparse Gaussian processes (SGPs), graphical model (GM),  logistic regression (LR),  beta-binomial (BB),  stochastic volatility model (SV),  probit regression (PR),  multinomial regression (MR), neural network (NN), Bayesian neural network (BNN),  gamma factor model (GFM),  Poisson gamma matrix factorisation (PGMF),  mixture of Gaussians (MoG),  Gaussian latent variable (GLV), switching linear dynamical system (SLDS).
        If the scheme proposed by the method has a name, this is noted in the final column. Abbreviations of the inference scheme are: variational inference (VI), stochastic variational inference (SVI), automatic differentiation VI (ADVI), incremental VI (IVI), variational message passing (VMP), non-conjugate variational message passing (NC-VMP),  simplified NC-VMP (SNC-VMP),  conjugate-computation VI (CVI),   power EP (PEP),  alpha-divergence PEP (ADPEP), convergent power EP (CPEP), stochastic natural gradient EP (SNEP),  stochastic power EP (SPEP), variational continual learning (VCL),  Bayesian gradient descent (BGD).}
        \label{table:past-work}
\end{table}

\afterpage{
\clearpage
\begin{landscape}
\begin{table*}[ht]
\small
\centering
\begin{tabular}{p{0.22\linewidth}p{0.10\linewidth}p{0.23\linewidth}p{0.24\linewidth}p{0.1\linewidth}}
\hline
\rule{0pt}{1.1em}\rule[-0.5em]{0pt}{1em}\textbf{Reference} & \textbf{Granularity} & \textbf{Optimisation}  & \textbf{Models} & \textbf{Name}\\
%
\hline
\multicolumn{3}{l}{\textbf{Global VI} [PVI $M=1$, see \Cref{sec:global}]} & & \\

\\
\citet{beal2003variational} & $M = 1$ & analytic & conjugate & VI \\
\citet{sato:2001} & $M = 1$ & analytic & conjugate (MoG) & 
\\
\citet{hinton+vancamp:1993} & $M = 1$ & gradient ascent & non-conjugate (NN) & 
\\
\citet{honkela+al:2010} & $M = 1$ & natural gradient (mean only)  & non-conj.~(MoG, NSSM, NFA)&  \\
\citet{hensman+al:2012} & $M = 1$ & CG with natural gradient& conjugate & 
\\
\citet{hensman+al:13} & $M = 1$ & stochastic natural gradient & conjugate & 
\\
\citet{hoffman+al:2013} & $M = 1$ & stochastic natural gradient & conjugate & SVI\\
\citet{kucukelbir+al:2017} & $M = 1$ & stochastic gradient descent & non-conjugate & ADVI \\
\citet{salimans+knowles:2013}   & $M = 1$ &  fixed-point + MC + stochastic& non-conjugate (PR, BB, SV) & 
\\
\citet{sheth+al:2015} & $M = 1$ & simplified fixed point & non-conjugate (GLV) &  \\
\citet{sheth+khardon:2016} & $M = 1$ & simplified fixed point & non-conjugate (GLV) &  \\
\cite{sheth+khardon:2016b} & $M = 1$ & simplified fixed point + MC & non-conjugate (two level) & \\
\hline
%
\multicolumn{3}{l}{\textbf{Local VI} [PVI $M > 1$, 
see \Cref{sec:local}]} & & \\ 
\citet{winn+bishop:2005} & $M > 1$ & analytic & conjugate (GM)  & VMP\\
\citet{archambeau+ermis:2015} & $M > 1$  & incremental & conjugate (LDA) & IVI\\
\citet{knowles+minka:2011} & $M > 1$  &  fixed-point & non-conjugate (LR, MR) & NC-VMP\\
\citet{wand:2014} & $M > 1$  & simplified fixed-point & non-conjugate (PMM, HLR)  & SNC-VMP \\
\citet{khan+li:2018}  & $M \geq 1$ & damped stochastic simplified fixed-point & conjugate / non-conj. (LR, GFM, PGMF) & CVI \\
\hline
%
\multicolumn{3}{l}{\textbf{Online VI} [one pass of PVI, see \Cref{sec:online}]} && \\
\citet{ghahramani:2000} & $M > 1$  & analytic & conjugate (MoG)  &\\
\citet{sato:2001} & $M > 1$ & analytic & conjugate (MoG) & online VB \\
\citet{broderick+al:2013} & $M > 1$ & analytic & conjugate (LDA) & streaming VI \\
\citet{bui+al:2017} & $M > 1$ & analytic/LBFGS & conjugate / non-conj. (SGPs) &  \\ 
\citet{nguyen+al:2018}  & $M > 1$ &Adam & non-conjugate (BNN) & VCL\\
\citet{zeno+al:2018} & $M > 1$ & fixed-point & non-conjugate (BNN) & BGD\\
\hline
%
\multicolumn{3}{l}{\textbf{Power EP} [PVI when $\alpha \rightarrow 0$, see \Cref{sec:pep,sec:alpha_ep,sec:stochastic_pep,sec:distributed_pep}]} & & \\
\citet{minka:2004, minka:2005}  &  $M > 1$ & series fixed point & non-conjugate (GM) & PEP  \\
\citet{minka:2004, minka:2005}  &  $M > 1$ & optimisation &  & ADPEP  \\
\citet{bui+al:2017b}  & $M > 1$ & analytic/fixed-point & conjugate / non-conj.~(GPs) & PEP \\
\citet{heskes2002expectation} & $M > 1$ & fixed-point updates in natural parameter space & non-conjugate (SLDS) & CPEP \\
\citet{hasenclever+al:2017} &  $M > 1$ & stochastic natural gradient updates in mean parameter space & non-conjugate (BNN) & SNEP \\
\citet{li+al:2015}  & $M > 1$ & stochastic fixed point & non-conjugate (LR, BNN) & SPEP \\
\hline
\end{tabular}
\setcounter{table}{0}
\caption{Variational inference schemes encompassed by the PVI framework. See previous page for full caption.}
\end{table*}
\end{landscape}
\clearpage
}

\section{Experiments}
\label{sec:experiments}
In this section we investigate the performance of sequential and synchronous PVI. Our aim is to demonstrate that PVI can recover the solution of global VI with greater communication efficiency in both simple and complex models, and in settings reflective of those in which PVI is likely to be deployed. Throughout, we make comparisons with global VI, VCL, streaming VB and BCM. We briefly describe these methods below. Global VI is implemented in the federated learning setting using synchronous PVI and performing a single local optimisation step using stochastic gradient descent. As discussed in \Cref{sec:optim:offtheshelf}, this recovers the exact dynamics of global VI. A more communication efficient but biased implementation of global VI is to communicate the current approximate posterior to a single client and perform a single stochastic update using a biased approximation of the global free-energy $\tilde{\mcF}_k(q(\*\theta)) = \frac{N_k}{N}\Exp{q(\*\theta)}{\log p(\bfy_k  |\*\theta)} - \KL{q(\*\theta)}{p(\*\theta)}$. We provide results for this implementation in \Cref{sec:appen:experiments}.

\begin{enumerate}[font=\bfseries, align=left, leftmargin=0pt, labelwidth=-5pt, labelsep=5pt]
    \item[Bayesian committee machine:] the Bayesian committee machine (BCM) \citep{tresp2000bayesian} performs VI for each dataset independently of the others, giving $M$ local approximate posteriors $\{q_m(\*\theta)\}_{m=1}^M$. The local approximate posteriors are then aggregated. We consider two BCM strategies:
    \begin{align}
        \textbf{BCM (same):}\quad &p(\*\theta | \bfy) \propto \frac{\prod_{m=1}^M p(\*\theta) p(\bfy_m | \*\theta)}{p(\*\theta)^{M - 1}} \approx \frac{\prod_{m=1}^M q_m(\*\theta)}{p(\*\theta)^{M-1}} \nonumber \\
        \textbf{BCM (split):}\quad &p(\*\theta | \bfy) \propto \prod_{m=1}^M p(\*\theta)^{N_m / N} p(\bfy_m | \*\theta) \approx \prod_{m=1}^M q_m(\*\theta). \nonumber
    \end{align}
    As alluded to in \Cref{sec:pvi:pvi_flavours}, BCM (same) is equivalent to a single round of synchronous PVI with no damping. Whilst the true global posterior decomposes as a normalised product of local posteriors, there is no guarantee that the normalised product of approximations to the local posteriors will form a good approximation to the global posterior due to the interaction between approximation errors. This is particularly so in the case of non-identifiable models such as BNNs, for which each local mean-field Gaussian approximation can be drawn to a different mode, and when they are combined the disagreement between the local approximations leads to a poor posterior approximation.
    
    \item[Variational continual learning:] a continual learning method, variational continual learning (VCL) \citep{nguyen+al:2018} visits each client in sequential order, performing VI with the prior equal to the previous approximate posterior, 
    \begin{equation}
    \label{eq:vcl_objective}
        q^{(i)}(\*\theta) = \argmax_{q(\*\theta)} \Exp{q(\*\theta)}{\log p(\bfy_m | \*\theta)} - \KL{q(\*\theta)}{q^{(i-1)}(\*\theta)}.
    \end{equation}
    This is equivalent to performing sequential PVI for a single round of client updates. Whilst exact implementation of the online Bayesian update rule is insensitive to the order in which clients are visited, this is not true for VCL. In general, we expect the final approximate posterior learnt by VCL to be biased towards performing well on the most recently visited dataset due to the accumulation of approximation errors.
    
    \item[Streaming variational Bayes:] analogous to the relationship between assumed density filtering (ADF) and EP, streaming variational Bayes (VB) performs sequential PVI without replacement, minimising the local free-energy \eqref{eq:vcl_objective} rather than \eqref{eq:vfe}. 
\end{enumerate}

In each experiment we either evaluate the performance of a logistic regression model or Bayesian neural network, which we define below.
\begin{enumerate}[font=\bfseries, align=left, leftmargin=0pt, labelwidth=-5pt, labelsep=5pt]
    \item[Logistic regression model:] we use a logistic regression model defined by the likelihood
    \begin{equation}
        p(y | \*\theta, \bfx) = \sigma\left(\tilde{\bfx}^{\top} \*\theta\right)^{y}\left( 1 - \sigma\left(\tilde{\bfx}^{\top} \*\theta\right)\right)^{1 - y} \nonumber
    \end{equation}
    where $\sigma(z) = \frac{1}{1 + e^{-z}}$ is the sigmoid function and $\tilde{\bfx} = \left[1,\ \bfx^{\top}\right]^{\top}$. In all experiments we use a standard normal prior on the parameters and a mean-field Gaussian variational approximation. The posterior predictive distribution is approximated using the probit approximation \citep{spiegelhalter1990sequential}:
    \begin{equation}
        \begin{aligned}
            p(y_* = 1 | \bfx_*, \bfy, \bfX) \approx \sigma\left(\frac{\*\mu^{\top}_q \tilde{\bfx}_*}{\sqrt{1 + \pi \tilde{\bfx}_*^{\top} \text{diag}(\*\sigma^2_q) \tilde{\bfx}_*}}\right) \nonumber
        \end{aligned}
    \end{equation}
    where $\*\mu_q$ and $\*\sigma_q$ and the mean and standard deviation of $q(\*\theta)$.
    
    \item[Bayesian neural network:] we use a fully connected neural network with ReLU activation functions. In all experiments we use a standard normal prior on the parameters and a mean-field Gaussian variational approximation. The posterior predictive distribution is approximated using the Monte Carlo estimate
    \begin{equation}
        p(y_* = 1 | \bfx_*, \bfy, \bfX) \approx \frac{1}{S} \sum_{s=1}^S p(y_* = 1 | \*\theta^{(s)}, \bfx_*) \quad \text{where}\ \*\theta^{(s)} \sim q(\*\theta). \nonumber
    \end{equation}
\end{enumerate}

\subsection{UCI Classification}
\label{sec:experiments:uci}
The purpose of this experiment is to evaluate the performance of PVI on relatively simple datasets under both homogeneous and inhomogeneous data distributions. We consider three binary classification tasks available on the UCI machine learning repository \citep{dua2019uci}: \textit{adult}, with the task being to predict whether income exceeds \$50,000/year based on census data; \textit{bank}, with the task being to predict if a client will subscribe to a term deposit; and \textit{credit}, with the task being the predict whether individuals default on payments. All three tasks contain sensitive financial information, thus are datasets for which the use of federated learning is suitable. For each of the federated learning schemes, we perform approximate inference of the parameters of a logistic regression model. We follow the dataset distribution scheme of \cite{sharma2019differentially} to split each dataset into $M=10$ partitions---five small and five large partitions. For each task, we consider a homogeneous split (A) and highly inhomogeneous split (B). A detailed description of the datasets and corresponding partitions is provided in \Cref{sec:appen:experiments:uci}.

\begin{figure}
    \begin{subfigure}[h]{\textwidth}
    \centering
    \includegraphics[width=\textwidth]{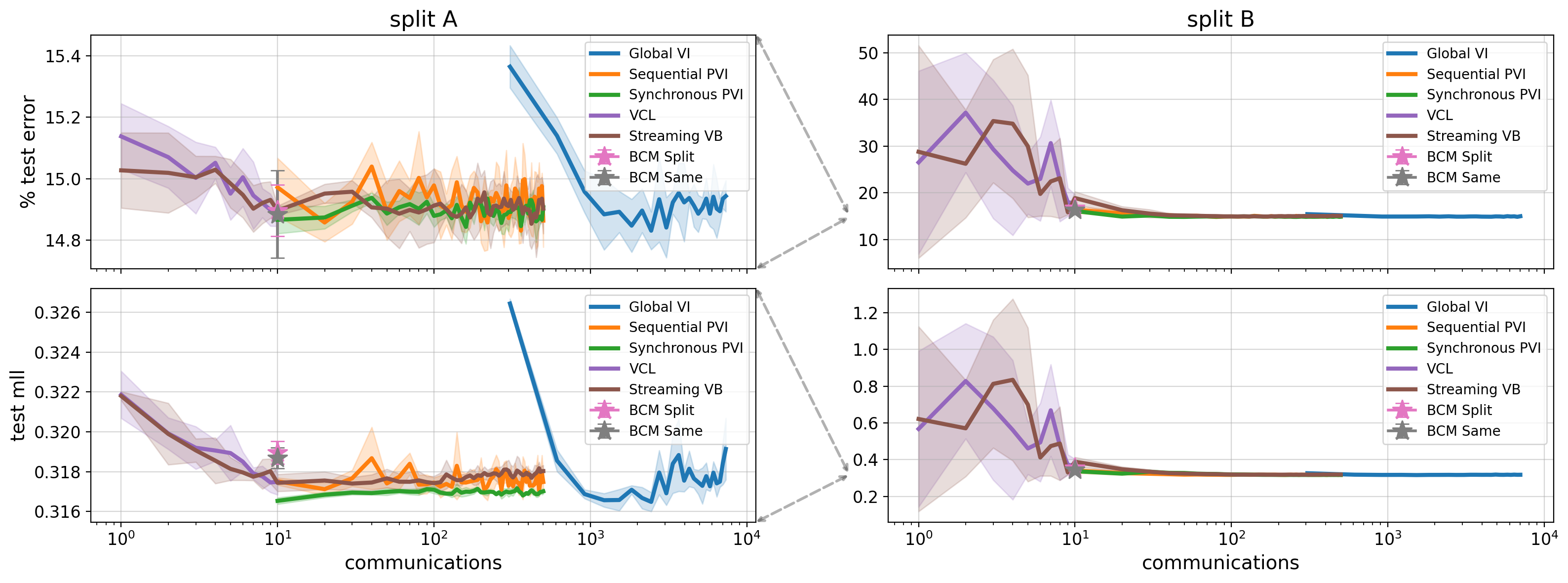}
    \caption{Adult}
    \label{fig:adult_pvi}
    \end{subfigure}
    \begin{subfigure}[h]{\textwidth}
    \centering
    \includegraphics[width=\textwidth]{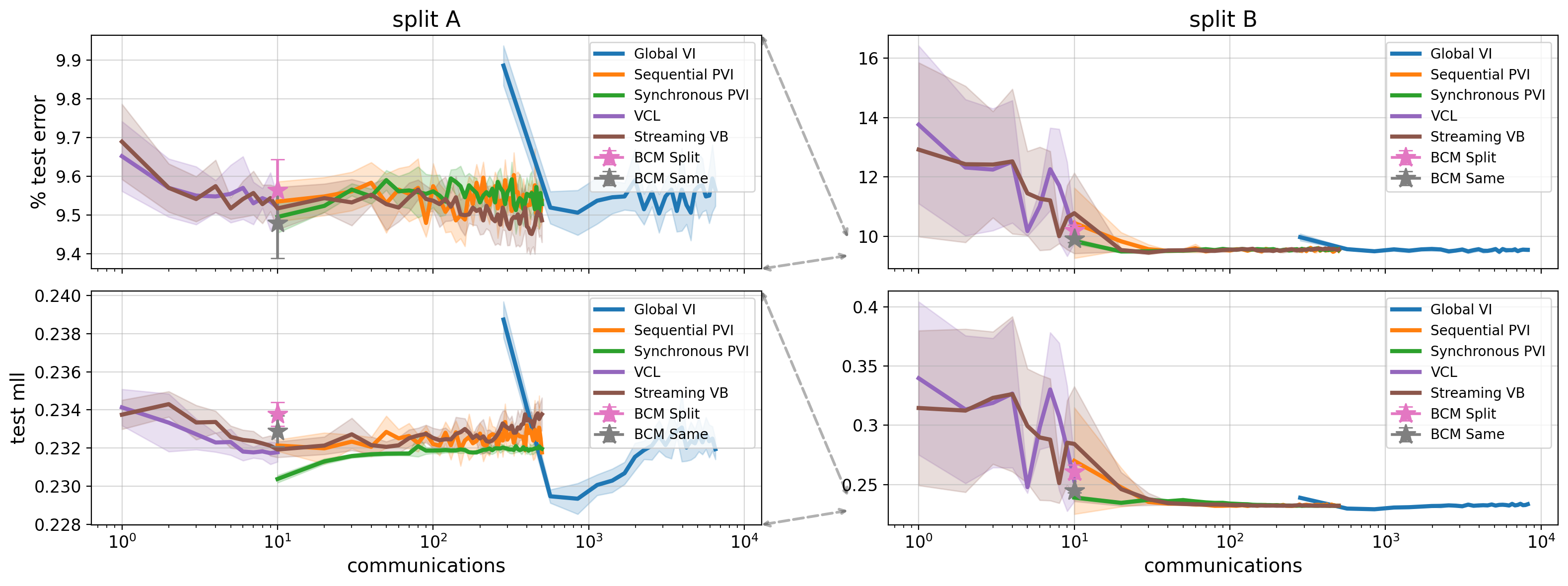}
    \caption{Bank}
    \label{fig:bank_pvi}
    \end{subfigure}
    \begin{subfigure}[h]{\textwidth}
    \centering
    \includegraphics[width=\textwidth]{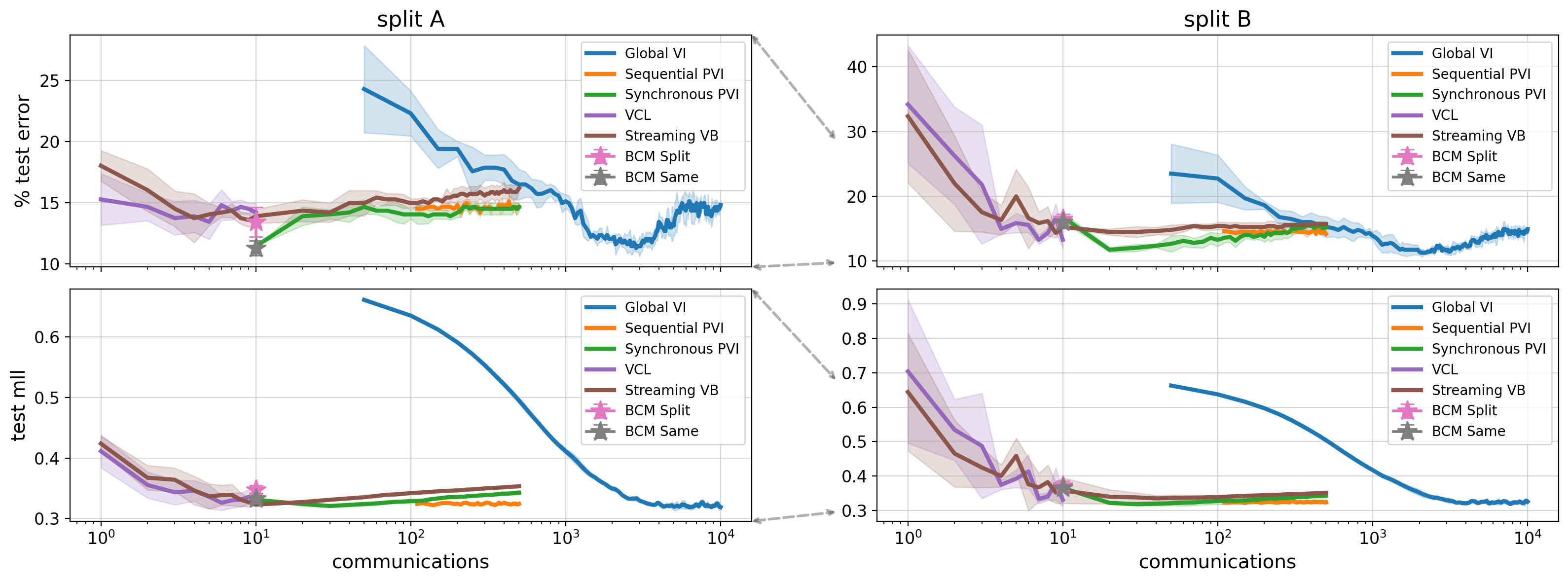}
    \caption{Credit}
    \label{fig:credit_pvi}
    \end{subfigure}
    \label{fig:uci_pvi}
    \caption[Predictive performance on the UCI classification tasks.]{A comparison between the predictive performance of the federated learning schemes on the test datasets of each of the UCI classification tasks. The mean $\pm$ standard deviation averaged over five random initialisations is shown. See text for details.}
\end{figure}

\Cref{fig:adult_pvi,,fig:bank_pvi,,fig:credit_pvi} compare the predictive performance of the approximate posteriors learnt by each of the federated learning schemes on each of the three UCI binary classification tasks. A damping factor of $\rho = 0.2$ is used for synchronous PVI. For both splits, we observe that both sequential and synchronous PVI converge towards the predictive performance obtained by the `optimal' approximate posterior learnt by global VI---this is in accordance with the theory presented in \Cref{sec:pvi}. For the homogeneous split A, convergence is achieved after a single round of communications.\footnote{This corresponds to a single communication for synchronous PVI and $K$ communications for sequential PVI.} Convergence takes noticeably longer for the inhomogeneous split B, but is nonetheless achieved after only a few rounds of communications. Importantly, the results demonstrate that PVI is able to recover the predictive performance obtained using global VI with far greater communication efficiency. 
Despite being relatively simple tasks, the plots reveal the comparatively poor performance of both BCM methods and VCL relative to PVI in the presence of inhomogeneity. Although the posterior distribution is identifiable, the client-level approximate factors are interdependent and so a single round of communications---as with BCM methods---is unable to identify the globally optimal approximate posterior. Similarly, the poor performance of VCL arises due to the interdependence of approximate factors and accumulation of approximation errors, the latter resulting in an approximate posterior biased towards performing well on the distribution of the dataset visited last. This explains the relatively high variance in final test set performance of VCL. The inadequacies of BCM and VCL demonstrate the importance of revisiting clients to model interdependencies between approximate factors. Yet, this is not sufficient---we also need the deletion step. Without the deletion step of PVI, streaming VB `over-counts' the observations resulting in over-sharpening of the likelihood approximation and overfitting, as illustrated by the predictive performance plot on the credit dataset.\footnote{The same degradation in predictive performance does not occur for the larger adult and bank datasets. Because the datasets are so large and the prior distribution relatively weak, the likelihood is much sharper than the prior and the posterior distribution resembles a delta on the maximum likelihood estimate. Since the maximum likelihood estimate is independent of the number of data points, both streaming VB and PVI converge towards the same solution.}

\subsection{Mortality Prediction}
\label{sec:experiments:mp}
In this experiment, our aim is to investigate the performance of PVI as the complexity of both the dataset and model is increased. We consider the mortality prediction task of the MIMIC-III dataset \citep{johnson2016mimic, harutyunyan2019multitask}. The presence of sensitive medical data motivates the use of federated learning. The dataset consists of recordings of 17 clinical variables recorded over the first 48 hours of each ICU stay, of which there are 42,276, corresponding to 33,798 unique patients. Each of the 17 recordings is divided into seven subsequences---full time series; first 10\% of time; first 25\% of time, first 50\% of time, last 50\% of time, last 25\% of time; and last 10\% of time---from which six statistical features are extracted. In total, each ICU stay is represented by a $17\times 7 \times 6 = 714$ dimensional feature vector.

\begin{figure}
    \begin{subfigure}[h]{\textwidth}
    \centering
    \includegraphics[width=\textwidth]{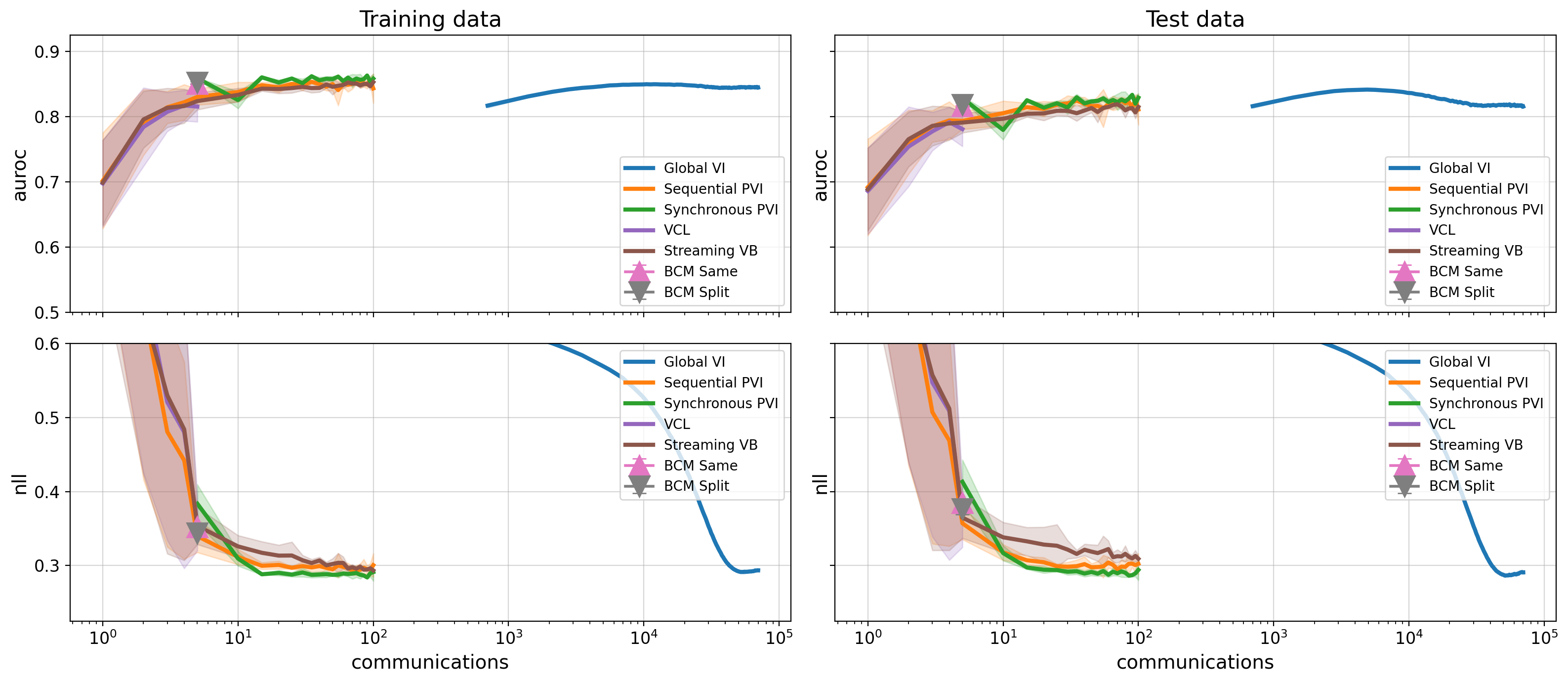}
    \caption{Logistic regression model.}
    \label{fig:mp_pvi_lr}
    \end{subfigure}
    \begin{subfigure}[h]{\textwidth}
    \centering
    \includegraphics[width=\textwidth]{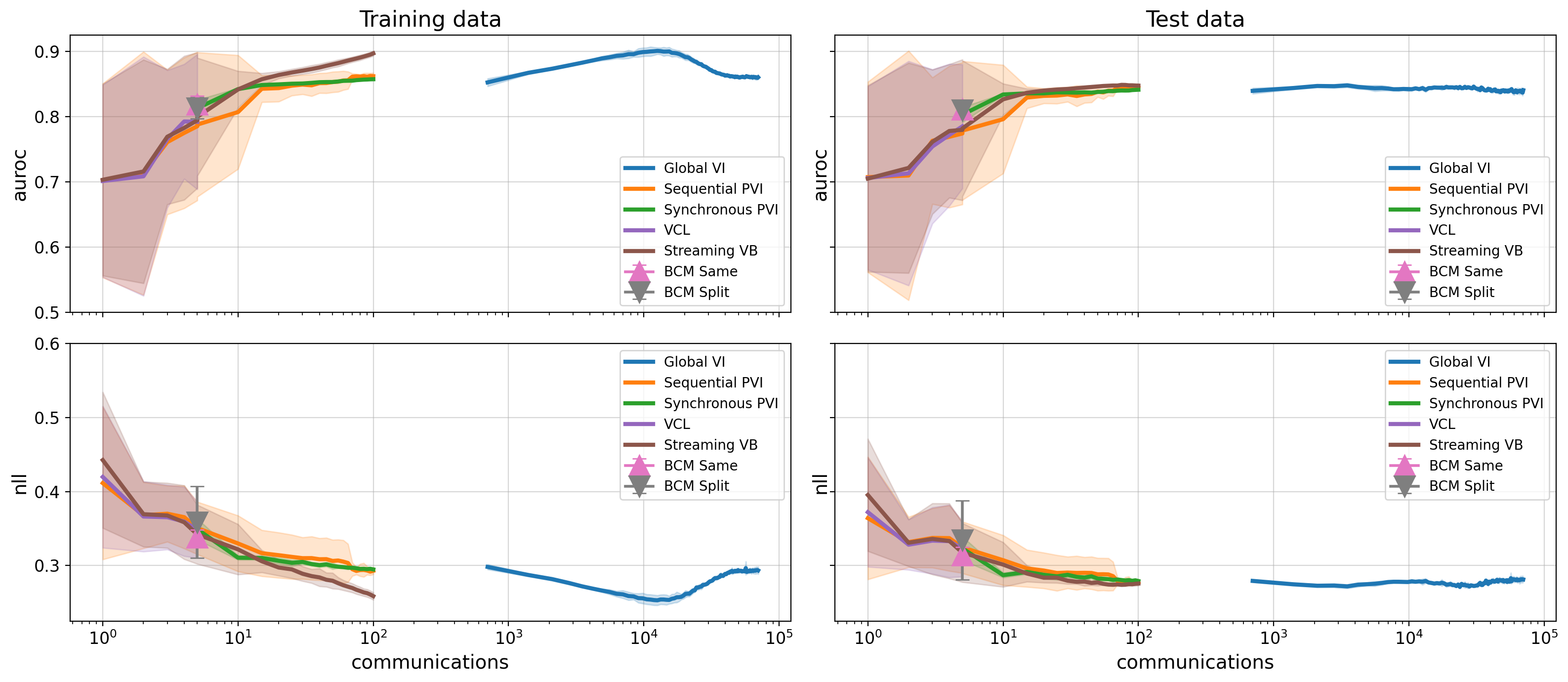}
    \caption{Two-layered BNN with 100 hidden units.}
    \label{fig:mp_pvi_bnn}
    \end{subfigure}
    \label{fig:mp_pvi}
    \caption[Predictive performance on the mortality prediction classification task using a logistic regression model.]{A comparison between the predictive performance of the federated learning schemes on the mortality prediction task. The mean $\pm$ standard deviation averaged over five random initialisations is shown. See text for details.}
\end{figure}

We construct $M=5$ inhomogeneous partitions of the data by performing k-means clustering on the inputs, the purpose being to simulate the inhomogeneity of patients that one might observe at different hospital---see \Cref{sec:appen:experiments:mortality} for details. We compare the performance of the federated learning methods in performing approximate inference in a simple logistic regression model and more complex two-layered BNN with 50 hidden units per layer. A damping factor of $\rho = 0.4$ is used for synchronous PVI. \Cref{fig:mp_pvi_lr,,fig:mp_pvi_bnn} plot the predictive performance of the methods.
Both PVI methods converge towards the same predictive performance achieved by global VI---doing so in far fewer communications---consolidating the observations made in \Cref{sec:experiments:uci} and demonstrating the effectiveness of PVI in more complex models. Convergence is noticeably smoother for the simpler logistic regression model than for the BNN, a consequence of the approximate posterior of the BNN being more complex. 

\subsection{Federated MNIST}
\label{sec:experiments:fmnist}
In this experiment, we evaluate the performance of PVI on a complex multi-class classification problem and investigate how performance is affected in the presence of severe inhomogeneity. The MNIST multi-class classification task \citep{lecun1998gradient} consists of 60,000 $28\times 28$ greyscale images of handwritten digits between 0 and 9. We consider performing federated learning on two distinct partitions of the dataset: a homogeneous split, in which the dataset is partitioned equally into $M=10$ groups of 6,000 datapoints each; and the inhomogeneous split of \cite{mcmahan2017communication}, in which the dataset is partitioned into $M=100$ groups of 600 datapoints, with each group consisting of examples of just two digits. The inhomogeneous split is highly non-IID and was introduced with the intention of testing federated learning algorithms to their limits.

We investigate the performance of the federated learning methods in performing approximate inference in a single-layered BNN with 200 hidden units. For synchronous PVI a damping factor of $\rho = 0.2$ is used for the homogeneous split and $\rho = 0.025$ is used for the inhomogeneous split. In both cases, we found this to be the smallest damping factor necessary for stable convergence. For sequential PVI, we used a damping factor of $\rho = 0.3$ for the inhomogeneous split which we found to result in faster convergence than the use of no damping. 

\subsubsection{Homogeneous Partition}
\begin{figure}[!ht]
    \centering
    \includegraphics[width=\textwidth]{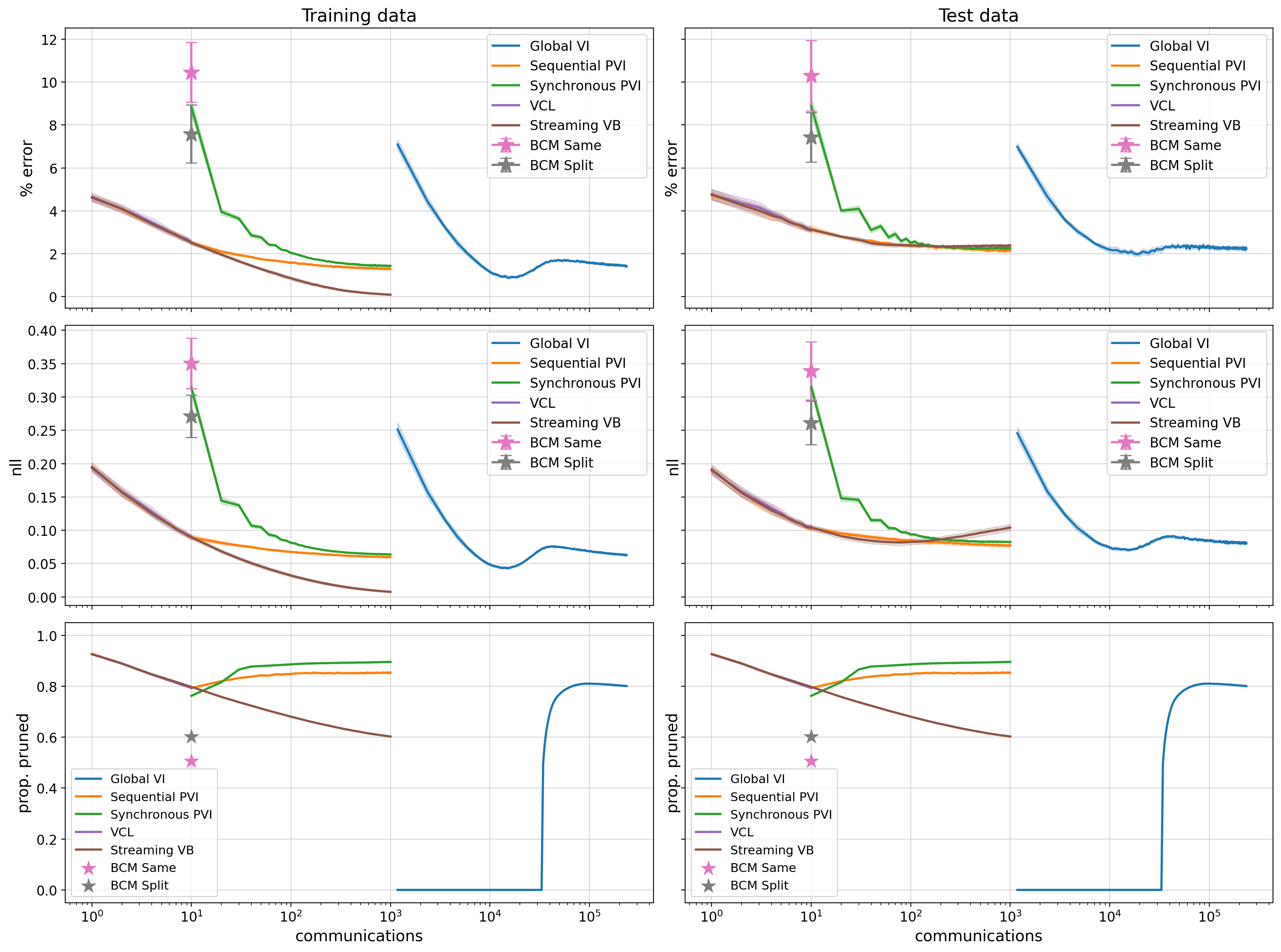}
    \caption[Predictive performance on the homogeneous split of the MNIST classification task.]{Plots of the predictive performance of the federated learning schemes on the homogeneous split of the MNIST classification task. The split is constructed by randomly partitioning the 60,000 training images into $M=10$ partitions. The mean $\pm$ standard deviation computed using five random initialisations is shown. See text for details.}
    \label{fig:fmnist_homo_pvi}
\end{figure}

\Cref{fig:fmnist_homo_pvi} compares the performance of federated learning schemes on the homogeneous split of the MNIST dataset.
As for the simpler binary classification tasks, the predictive performance of both PVI methods converges towards the same asymptote as global VI with far greater communication efficiency. Both BCM methods perform poorly, which again is due to the non-identifiability of the approximate posterior. VCL achieves reasonable predictive performance; yet, it is evident that by revisiting clients and modifying their contribution to the approximate posterior---as in sequential PVI---predictive performance can be improved. Without the deletion step, streaming VB overfits the training dataset which results in a degradation in test set performance.

\begin{figure}[!ht]
    \centering
    \includegraphics[width=\textwidth]{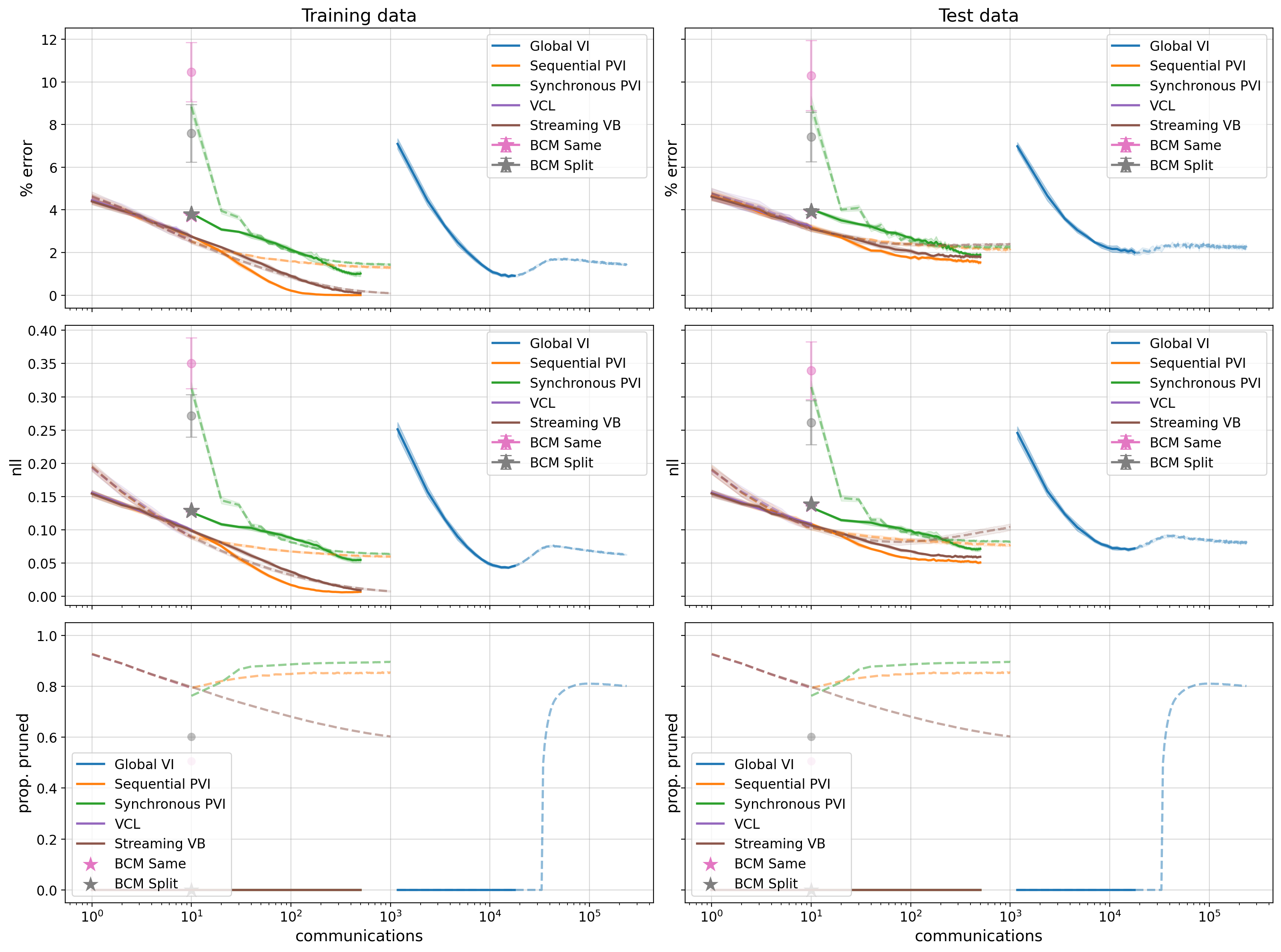}
    \caption[Predictive performance on the homogeneous split of the MNIST classification task when early stopping is applied.]{Plots of the predictive performance of the federated learning schemes on the homogeneous split of the MNIST classification task. The solid lines indicate the use of early stopping, whereas the dashed lines indicate the free-energies are optimised until convergence. The mean $\pm$ standard deviation computed using five random initialisations is shown. See text for details.}
    \label{fig:fmnist_homo_es_pvi}
\end{figure}

A pathology of mean-field variational approximations to BNNs is the over-pruning effect \citep{trippe2018overpruning}. \Cref{fig:fmnist_homo_pvi} demonstrates this in full effect, showing that as the number of pruned parameters increases the predictive performance worsens.\footnote{A parameter, $\theta_i$, is marked as pruned if $\textrm{KL}(q(\theta_i)\|p(\theta_i)) < 0.1$.} To mitigate this effect, we consider applying early stopping to the local free-energy optimisations by checking for convergence of the expected log-likelihood term $\Exp{q(\*\theta)}{\log p(\bfy_k | \*\theta)}$. Rather than initialise at the context distribution, it is important to initialise the approximate posteriors at a `tight' distribution for all updates to ensure early stopping can be applied.\footnote{If initialised at the context distribution as before, the local free-energies do not get the opportunity to stop early.} \Cref{fig:fmnist_homo_es_pvi} compares the performance of the federated learning schemes with early stopping based on the expected log-likelihood term on the homogeneous split. 
Observe that the use of early stopping is able to identify the best predictive performance for global VI with good precision, indicating its effectiveness. For the PVI methods and streaming VB, the approximate posterior overfits to the training dataset. Unlike streaming VB without early stopping, however, this does not correspond to a worsening in test set predictive performance: all three methods achieve a final test set negative log-likelihood better than that achieved without the use of early stopping, with streaming PVI and VB outperforming global VI. 

\subsubsection{Inhomogeneous Partition}
\begin{figure}[!ht]
    \centering
    \includegraphics[width=\textwidth]{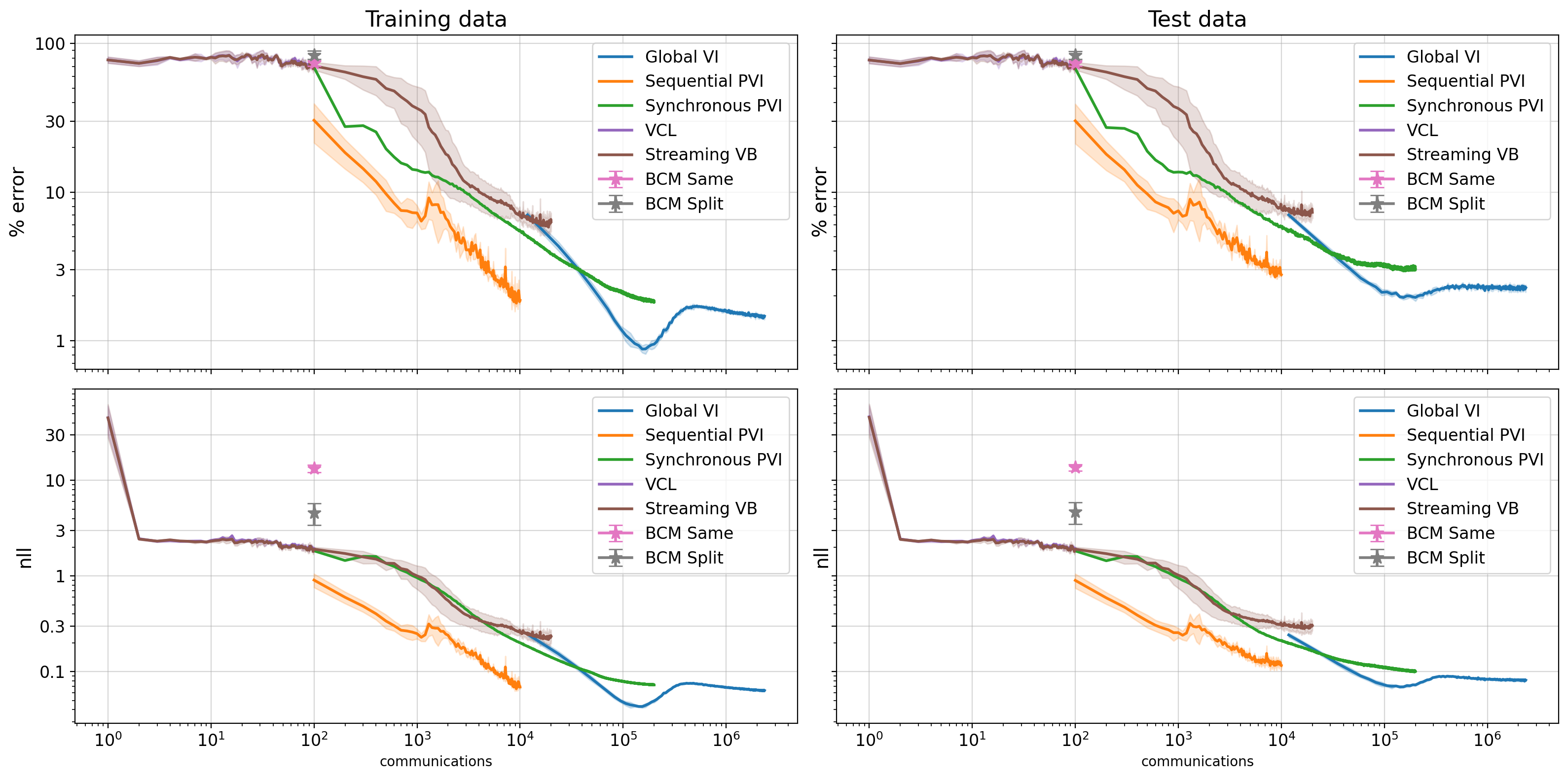}
    \caption[Predictive performance on the inhomogeneous split of the MNIST classification task.]{Plots of the predictive performance of the federated learning schemes on the inhomogeneous split of the MNIST classification task. The split is constructed by partitioning the 60,000 training images into $M=100$ partitions, each with at most two digits. The mean $\pm$ standard deviation computed using five random initialisations is shown. See text for details.}
    \label{fig:fmnist_inhomo_pvi}
\end{figure}

\Cref{fig:fmnist_inhomo_pvi} compares the performance of the federated learning schemes on the inhomogeneous split of the MNIST dataset. Note that the performance metrics are plotted on the log-scale, as the variation between the performance of each method is significantly greater than in any of the previous experiments.

The communication efficiency of all methods is considerably worse than in the homogeneous case. For global VI, this is because 100 communications are required to simulate a single iteration of the exact VI dynamics (one for each client), rather than 10 communications in the homogeneous case. The initial communication efficiency of both PVI methods is far better than that of global VI; however, they achieve a worse final predictive performance with a test set error of 3\%. This is a consequence of 1) the inhomogeneity of client data distributions, making it difficult for the approximate likelihoods to agree upon aggregation, and 2) the large number of clients, which means that approximate likelihoods are updated infrequently in the case of sequential PVI and a crippling large degree of damping is used in the synchronous PVI case. The interaction of these two factors hinders the effectiveness of both methods. Nonetheless, we emphasise that the degree of inhomogeneity present in this experiment is significant, and likely to be greater than that encountered in many practical situations. We expect extreme inhomogeneity to demand relatively small updates are made at each client before communication, which makes it impossible to develop communication efficient algorithms.

\section{Conclusion}
\label{sec:conclusion}
This paper provided a general framework for performing variational inference in the federated learning setting, partitioned variational inference (PVI). 

We provided theoretical and empirical results showing that PVI is able to recover the global VI solution whilst respecting the data privacy constraints of federated learning, demonstrating its efficacy in the presence of varying degrees and forms of inhomogeneity across data partitions. Departing from the federated learning setting, we showed that the PVI framework flexibly subsumes many existing VI methods as special cases and sheds light on the connections between a wealth of techniques. 


One of the key contributions of this work is the connection of deterministic local message passing methods with global optimisation-based schemes. Each of these different branches of approximate inference has been arguably developed independently of each other, for example, existing probabilistic programming toolkits tend to work primarily with one approach but not both \citep[see e.g.][]{tran+al:2017,minka+al:2018}. This paper suggests these methods are inter-related and practitioners could benefit from a unified framework, i.e.~there are ways to expand the existing probabilistic programming packages to gracefully handle both approaches. Additionally, the PVI framework could be used to automatically choose a granularity level and an optimisation scheme that potentially offer a better inference method for the task or the model at hand. It is, however, unclear how flexible variational approximations such as mixtures of exponential family distributions \cite[see e.g.][]{sudderth+al:2010} or normalising flows \citep{rezende+mohamed:2015} can be efficiently and tractably accommodated in the PVI framework. We leave these directions as future work.

The experiments in \Cref{sec:experiments} demonstrated that PVI is well-suited to learning with decentralised data. Deployment of PVI in this setting is practical as its implementation only requires a straightforward modification of existing global VI implementations. We have also explored how this algorithm allows data parallelism---each local client stores a complete copy of the model---and communication efficient, uncertainty-aware updates between clients. A potential future extension of the proposed approach is to use model parallelism. That is, in addition to decentralising the data and computation across the clients, the model itself is partitioned. As commonly done in many deep learning training algorithms, model parallelism could be achieved by assigning the parameters (and computation) of different layers of the network to different devices. Alternatively, for spatio-temporal datasets modelled using spatio-temporally distributed parameters, each client could support part of the full model relevant for its local region of space. Another potential research direction is PVI with coordinator-free, peer-to-peer only communication between clients. This could be achieved by a client passing each update to several randomly selected other clients, who then apply the changes, rather than to a central parameter server.

Whilst our implementation of PVI performed either sequential or synchronous local optimisation until convergence, the relationship between PVI and global VI---namely, that they share the same fixed-points---holds regardless of how optimisation is performed. In \Cref{sec:optim} we demonstrated how the exact dynamics of global VI can be recovered by terminating the local optimisation procedure of synchronous PVI after a single update step is performed. The space in-between performing a single update and running to convergence is largely unexplored, and could result in algorithms that improve on the communication efficiency of our implementation of PVI. Other directions we are yet to explore include stochastic implementations of synchronous PVI, the use of alpha-divergences to define the local free-energy, and a dynamic scheduling of clients.

\section*{Acknowledgements}
\label{sec:acknowledgements}
We would like to thank Jonathan So for his insights on the properties of the double-loop algorithm for EP. Matthew Ashman is supported by the George and Lilian Schiff Foundation. Adrian Weller acknowledges support from a Turing AI Fellowship under grant EP/V025379/1, The Alan Turing Institute, and the Leverhulme Trust via CFI. Siddharth Swaroop is supported by an EPSRC DTP studentship and Microsoft Research EMEA PhD Award. Richard E.\ Turner is supported by Google, Amazon, ARM, Improbable and EPSRC grant EP/T005386/1.

\bibliography{pvi}

\newpage

\appendix

\section{Hyperparameter Learning}
\label{sec:hyper_learning}

Many probabilistic models depend on a set of hyperparameters $\*\epsilon$ and it is often necessary to learn suitable settings from data to achieve good performance on a task. One method is to optimise the variational free-energy thereby approximating maximum likelihood learning. The gradient of the global variational free-energy decomposes into a set of local computations, as shown in \Cref{sec:hyper-gradients},
\begin{align}
\frac{\mathrm{d}}{ \mathrm{d} \*\epsilon} \mathcal{F}(\*\epsilon,q(\*\theta)) =& \sum_{m=1}^M \mathbb{E}_{q(\*\theta)} \left [\frac{\mathrm{d}}{ \mathrm{d} \*\epsilon} \log p(\mathbf{y}_m|\*\theta,\*\epsilon) \right] + \mathbb{E}_{q(\*\theta)} \left[\frac{\mathrm{d}}{ \mathrm{d} \*\epsilon} \log p(\*\theta|\*\epsilon) \right]. \nonumber
\end{align}
This expression holds for general $q(\*\theta)$ and is valid both for coordinate ascent (iterating between updating $\*\epsilon$ with $q(\*\theta)$ fixed and updating $q(\*\theta)$ with $\*\epsilon$ fixed) and for optimising the collapsed bound (where the approximate posterior optimises the global free-energy $q(\*\theta) = q^*(\*\theta)$ and therefore depends implicitly on $\*\epsilon$). The latter does not require us to compute $q(\*\theta)$ using PVI, but rather query the gradients of the log-likelihood for each data partition with respect to the hyperparameters. Notice that this expression is amenable to stochastic approximation which leads to optimization schemes that use only local information at each step. When combined with different choices for the optimization of the local free-energies w.r.t.~$q(\*\theta)$, this leads to a wealth of possible hyperparameter optimization schemes.

In cases where a distributional estimate for the hyperparameters is necessary, e.g.~in continual learning, the PVI framework above can be extended to handle the hyperparameters. In particular, the approximate posterior in \eqref{eq:approx} can be modified as follows,
\begin{align}
    q(\*\theta, \*\epsilon) &= \frac{1}{\mcZ_q} p(\*\epsilon) p(\*\theta|\*\epsilon)  \prod_{m=1}^{M} t_m(\*\theta, \*\epsilon) 
    \approx \frac{1}{\mathcal{Z}} p(\*\epsilon) p(\*\theta|\*\epsilon) \prod_{m=1}^{M} p(\bfy_m|\*\theta, \*\epsilon) = p(\*\theta, \*\epsilon | \bfy), \nonumber
\end{align}
where the approximate likelihood factor $t_m(\*\theta, \*\epsilon)$ now involves both the model parameters and the hyperparameters. Similar to \eqref{eq:vfe}, the approximate posterior above leads to the following local variational free-energy,
\begin{align}
    \mathcal{F}^{(i)}_k(q(\*\theta,\*\epsilon)) = \int \mathrm{d}\*\theta\mathrm{d}\*\epsilon q(\*\theta,\*\epsilon) \log \frac{q^{(i-1)}(\*\theta, \*\epsilon) p(\bfy_{k}|\*\theta, \*\epsilon)}{q(\*\theta, \*\epsilon) t^{(i-1)}_{k}(\*\theta, \*\epsilon)}. \nonumber
\end{align}
Note that this approach retains all favourable properties of PVI such as local computation and flexibility in choosing optimization strategies and stochastic approximations.

\section{Proofs}
\label{sec:appen:proofs}

\subsection{Relating Local KL Minimisation to Local Free-Energy Minimisation: Proof of \Cref{prop:local-fe-opt}}
\label{sec:appen:local-FE-KL}
Substituting the tilted distribution $\widehat{p}^{(i)}_k(\*\theta)$ into the KL divergence yields,
\begin{align*}
\mathrm{KL} \left( q(\*\theta) \| \widehat{p}^{(i)}_k(\*\theta) \right)  &= \int \mathrm{d}\*\theta q(\*\theta) \log  \frac{  q(\*\theta) \widehat{\mcZ}^{(i)}_k t^{(i-1)}_{k}(\*\theta) }{ q^{(i-1)}(\*\theta) p(\bfy_{k}|\*\theta)} \\
&= \log \widehat{\mcZ}^{(i)}_k - \int \mathrm{d}\*\theta q(\*\theta) \log  \frac{ q^{(i-1)}(\*\theta) p(\bfy_{k}|\*\theta) }{q(\*\theta)t^{(i-1)}_{k}(\*\theta)}.
\end{align*}
Hence $\mcF^{(i)}_k(q(\*\theta)) = \log \widehat{\mcZ}^{(i)}_k - \mathrm{KL} \left( q(\*\theta) \| \widehat{p}^{(i)}_k(\*\theta) \right)$.

\subsection{The Sum of Local Free-Energies is Equal to the Global Free-Energy}
\label{sec:appen:local-global-fe}
Let $q(\*\theta) = \frac{1}{\mcZ_q} p(\*\theta) \prod_{m=1}^M t_m(\*\theta)$. We have
\begin{equation}
\begin{aligned}
   \sum_{m=1}^M \mcF_m(q(\*\theta)) + \log \mcZ_q &= \sum_{m=1}^M \int q(\*\theta) \log \frac{q(\*\theta) p(\bfy_m | \*\theta)}{q(\*\theta) t_m(\*\theta)} \mathrm{d}\*\theta + \log \mcZ_q \\
   &= \int q(\*\theta) \log \frac{\prod_{m=1}^M p(\bfy_m | \*\theta)}{\prod_{m=1}^M t_m(\*\theta)} \mathrm{d}\*\theta + \log \mcZ_q \\
   &= \int q(\*\theta) \log \frac{p(\*\theta) \prod_{m=1}^M p(\bfy_m | \*\theta)}{\frac{1}{\mcZ_q}p(\*\theta)\prod_{m=1}^M t_m(\*\theta)} \mathrm{d}\*\theta \\
   &= \int q(\*\theta) \log \frac{p(\*\theta)\prod_{m=1}^M p(\bfy_m | \*\theta)}{q(\*\theta)} \mathrm{d}\*\theta \\
   &= \mcF(q(\*\theta)) \nonumber.
\end{aligned}
\end{equation}

\subsection{Maximisation of Local Free-Energies Implies Maximisation of Global Free-Energy: Proof of \Cref{prop:fixed-point}}
\label{sec:appen:local-global}
Let $\*\eta_q$ and $\*\eta^*_q$ be the variational parameters of $q(\*\theta)$ and $q^*(\*\theta)$ respectively. We can write $q(\*\theta) = q(\*\theta; \*\eta_q)$ and $q^*(\*\theta) = q(\*\theta; \*\eta^*_q) = \frac{1}{\mcZ_{q^*}} p(\*\theta) \prod_m t_m(\*\theta; \*\eta^*_q)$. First, we show that at convergence, the derivative of the global free-energies equals the sum of the derivatives of the local free-energies. The derivative of the local free-energy $\mathcal{F}_m(q(\*\theta))$ w.r.t. $\*\eta_q$ is:
\begin{align*}
\frac{\mathrm{d} \mathcal{F}_m(q(\*\theta))}{\mathrm{d} \*\eta_q} &= \frac{\mathrm{d}}{\mathrm{d} \*\eta_q} \int \mathrm{d}\*\theta q(\*\theta;\*\eta_q) \log \frac{q(\*\theta; \*\eta^*_q) p(\bfy_m|\*\theta)}{q(\*\theta;\*\eta_q) t_m(\*\theta; \*\eta^*_q)}\\
&= \frac{\mathrm{d}}{\mathrm{d} \*\eta_q}  \int \mathrm{d}\*\theta q(\*\theta;\*\eta_q) \log \frac{p(\bfy_m|\*\theta)}{t_m(\*\theta; \*\eta^*_q)} + \frac{\mathrm{d}}{\mathrm{d} \*\eta_q} \int \mathrm{d}\*\theta q(\*\theta;\*\eta_q) \log \frac{q(\*\theta; \*\eta^*_q)}{q(\*\theta;\*\eta_q)}\\
&= \frac{\mathrm{d}}{\mathrm{d} \*\eta_q} \int \mathrm{d}\*\theta q(\*\theta;\*\eta_q) \log \frac{p(\bfy_m|\*\theta)}{t_m(\*\theta; \*\eta^*_q)} + \int \mathrm{d}\*\theta \frac{\mathrm{d} q(\*\theta;\*\eta_q)}{\mathrm{d} \*\eta_q} \log \frac{q(\*\theta; \*\eta^*_q)}{q(\*\theta;\*\eta_q)} \\
&\quad - \cancelto{0}{\int \mathrm{d}\*\theta \frac{\mathrm{d} q(\*\theta;\*\eta_q)}{\mathrm{d} \*\eta_q}}.
\end{align*}
Thus, at convergence when $\*\eta_q = \*\eta^*_q$,
\begin{align*}
\frac{\mathrm{d} \mathcal{F}_m(q(\*\theta))}{\mathrm{d} \*\eta_q} \biggr|_{\*\eta_q = \*\eta^*_q} = \frac{\mathrm{d}}{\mathrm{d} \*\eta_q} \int \mathrm{d}\*\theta q(\*\theta;\*\eta_q) \log \frac{p(\bfy_m|\*\theta)}{t_m(\*\theta; \*\eta^*_q)} \biggr|_{\*\eta_q = \*\eta^*_q}.
\end{align*}
Summing both sides over all $m$,
\begin{align*}
    \sum_m \frac{\mathrm{d} \mathcal{F}_m(q(\*\theta))}{\mathrm{d} \*\eta_q} \biggr|_{\*\eta_q = \*\eta^*_q} &= \frac{\mathrm{d}}{\mathrm{d} \*\eta_q} \int \mathrm{d}\*\theta q(\*\theta;\*\eta_q) \log \frac{\prod_m p(\bfy_m|\*\theta)}{\prod_m t_m(\*\theta; \*\eta^*_q)} \biggr|_{\*\eta_q = \*\eta^*_q} \\
    &= \frac{\mathrm{d}}{\mathrm{d} \*\eta_q} \int \mathrm{d}\*\theta q(\*\theta;\*\eta_q) \log \frac{p(\*\theta) \prod_m p(\bfy_m|\*\theta)}{q(\*\theta;\*\eta^*_q)} \biggr|_{\*\eta_q = \*\eta^*_q} \\
    &\ - \cancelto{0}{\frac{\mathrm{d}}{\mathrm{d} \*\eta_q} \int \mathrm{d}\*\theta q(\*\theta;\*\eta_q) \log \mcZ_{q^*} \biggr|_{\*\eta_q = \*\eta_q^*}}.
\end{align*}
Now consider the derivative of the global free-energy $\mathcal{F}(q(\*\theta))$:
\begin{align*}
\frac{\mathrm{d} \mathcal{F}(q(\*\theta))}{\mathrm{d} \*\eta_q} &= \frac{\mathrm{d}}{\mathrm{d} \*\eta_q} \int \mathrm{d}\*\theta q(\*\theta; \*\eta_q) \log \frac{p(\*\theta) \prod_m p(\bfy_m|\*\theta)}{q(\*\theta; \*\eta_q)} \\
&= \int \mathrm{d}\*\theta \frac{\mathrm{d} q(\*\theta; \*\eta_q)}{\mathrm{d} \*\eta_q} \log \frac{p(\*\theta) \prod_m p(\bfy_m|\*\theta)}{q(\*\theta; \*\eta_q)} - \cancelto{0}{\int \mathrm{d}\*\theta \frac{\mathrm{d} q(\*\theta;\*\eta_q)}{\mathrm{d} \*\eta_q}}.
\end{align*}
Hence,
\begin{align*}
\frac{\mathrm{d} \mathcal{F}(q(\*\theta))}{\mathrm{d} \*\eta_q} \biggr|_{\*\eta_q = \*\eta^*_q} = \int \mathrm{d}\*\theta \frac{\mathrm{d} q(\*\theta; \*\eta_q)}{\mathrm{d} \*\eta_q} \log \frac{p(\*\theta) \prod_m p(\bfy_m|\*\theta)}{q(\*\theta; \*\eta^*_q)} \biggr|_{\*\eta_q = \*\eta^*_q} = \sum_m \frac{\mathrm{d} \mathcal{F}_m(q(\*\theta))}{\mathrm{d} \*\eta_q} \biggr|_{\*\eta_q = \*\eta^*_q}.
\end{align*}
For all $m$, since $q^*(\*\theta) = \argmax_{q(\*\theta) \in \mathcal{Q}} \mathcal{F}_m(q(\*\theta))$, we have $\frac{\mathrm{d} \mathcal{F}_m(q(\*\theta))}{\mathrm{d} \*\eta_q} \big|_{\*\eta_q = \*\eta^*_q} = 0$, which implies:
\begin{align*}
\frac{\mathrm{d} \mathcal{F}(q(\*\theta))}{\mathrm{d} \*\eta_q} \biggr|_{\*\eta_q = \*\eta^*_q} = \sum_m \frac{\mathrm{d} \mathcal{F}_m(q(\*\theta))}{\mathrm{d} \*\eta_q} \biggr|_{\*\eta_q = \*\eta^*_q} = 0.
\end{align*}
Thus, $q^*(\*\theta)$ is an extremum of $\mathcal{F}(q(\*\theta))$.

Now we show that $q^*(\*\theta)$ is a maximum of $\mathcal{F}(q(\*\theta))$ by considering the Hessian $\frac{\mathrm{d}^2 \mathcal{F}(q(\*\theta))}{\mathrm{d} \*\eta_q \mathrm{d} \*\eta^\intercal_q}$ of the global free-energy at convergence. Similar to the derivative case, we now show that the Hessian of the global free-energies equals the sum of the Hessians of the local free-energies. The Hessian of the local free-energy $\mathcal{F}_m(q(\*\theta))$ w.r.t. $\*\eta_q$ is:
\begin{align*}
\frac{\mathrm{d}^2 \mathcal{F}_m(q(\*\theta))}{\mathrm{d} \*\eta_q \mathrm{d} \*\eta^\intercal_q} &= \frac{\mathrm{d}^2}{\mathrm{d} \*\eta_q \mathrm{d} \*\eta^\intercal_q} \int \mathrm{d}\*\theta q(\*\theta;\*\eta_q) \log \frac{q(\*\theta; \*\eta^*_q) p(\bfy_m|\*\theta)}{q(\*\theta;\*\eta_q) t_m(\*\theta; \*\eta^*_q)} \\
&= \frac{\mathrm{d}^2}{\mathrm{d} \*\eta_q \mathrm{d} \*\eta^\intercal_q}  \int \mathrm{d}\*\theta q(\*\theta;\*\eta_q) \log \frac{p(\bfy_m|\*\theta)}{t_m(\*\theta; \*\eta^*_q)} + \frac{\mathrm{d}^2}{\mathrm{d} \*\eta_q \mathrm{d} \*\eta^\intercal_q} \int \mathrm{d}\*\theta q(\*\theta;\*\eta_q) \log \frac{q(\*\theta; \*\eta^*_q)}{q(\*\theta;\*\eta_q)} \\
&= \frac{\mathrm{d}^2}{\mathrm{d} \*\eta_q \mathrm{d} \*\eta^\intercal_q}  \int \mathrm{d}\*\theta q(\*\theta;\*\eta_q) \log \frac{p(\bfy_m|\*\theta)}{t_m(\*\theta; \*\eta^*_q)} + \frac{\mathrm{d}}{\mathrm{d} \*\eta_q} \biggl( \int \mathrm{d}\*\theta \frac{\mathrm{d} q(\*\theta; \*\eta_q)}{\mathrm{d} \*\eta^\intercal_q} \log \frac{q(\*\theta; \*\eta^*_q)}{q(\*\theta;\*\eta_q)} \\
& \hspace{225pt} - \cancelto{0}{\int \mathrm{d}\*\theta \frac{\mathrm{d}q(\*\theta; \*\eta_q)}{\mathrm{d} \*\eta^\intercal_q}} \biggr) \\
&=  \frac{\mathrm{d}^2}{\mathrm{d} \*\eta_q \mathrm{d} \*\eta^\intercal_q}  \int \mathrm{d}\*\theta q(\*\theta;\*\eta_q) \log \frac{p(\bfy_m|\*\theta)}{t_m(\*\theta; \*\eta^*_q)} + \int \mathrm{d}\*\theta \frac{\mathrm{d^2} q(\*\theta; \*\eta_q)}{\mathrm{d} \*\eta_q \mathrm{d} \*\eta^\intercal_q} \log \frac{q(\*\theta; \*\eta^*_q)}{q(\*\theta;\*\eta_q)} \\
&\ - \int \mathrm{d}\*\theta \frac{\mathrm{d}\log q(\*\theta; \*\eta_q)}{\mathrm{d} \*\eta_q} \frac{\mathrm{d}q(\*\theta; \*\eta_q)}{\mathrm{d} \*\eta^\intercal_q}.
\end{align*}
The final term can be written as
\begin{align*}
    \left(\int \mathrm{d}\*\theta \frac{\mathrm{d}\log q(\*\theta; \*\eta_q)}{\mathrm{d} \*\eta_q} \frac{\mathrm{d}q(\*\theta; \*\eta_q)}{\mathrm{d} \*\eta^\intercal_q}\right)^\intercal &= \frac{\mathrm{d}}{\mathrm{d}\*\eta_q} \int \mathrm{d}\*\theta q(\*\theta; \*\eta_q) \frac{\mathrm{d}\log q(\*\theta; \*\eta_q)}{\mathrm{d} \*\eta^\intercal_q} - \cancelto{0}{\int \mathrm{d}\*\theta \frac{\mathrm{d}^2 q(\*\theta; \*\eta_q)}{\mathrm{d}\*\eta_q\mathrm{d}\*\eta^\intercal_q}} \\
    &= \frac{\mathrm{d}}{\mathrm{d}\*\eta_q} \int \mathrm{d}\*\theta \frac{\mathrm{d} q(\*\theta; \*\eta_q)}{\mathrm{d} \*\eta^\intercal_q} = \*0.
\end{align*}
Thus
\begin{equation*}
    \frac{\mathrm{d}^2 \mathcal{F}_m(q(\*\theta))}{\mathrm{d} \*\eta_q \mathrm{d} \*\eta^\intercal_q} =  \frac{\mathrm{d}^2}{\mathrm{d} \*\eta_q \mathrm{d} \*\eta^\intercal_q}  \int \mathrm{d}\*\theta q(\*\theta;\*\eta_q) \log \frac{p(\bfy_m|\*\theta)}{t_m(\*\theta; \*\eta^*_q)} + \int \mathrm{d}\*\theta \frac{\mathrm{d^2} q(\*\theta; \*\eta_q)}{\mathrm{d} \*\eta_q \mathrm{d} \*\eta^\intercal_q} \log \frac{q(\*\theta; \*\eta^*_q)}{q(\*\theta;\*\eta_q)}.
\end{equation*}
At convergence when $\*\eta_q = \*\eta^*_q$,
\begin{align*}
\frac{\mathrm{d}^2 \mathcal{F}_m(q(\*\theta))}{\mathrm{d} \*\eta_q \mathrm{d} \*\eta^\intercal_q} \biggr|_{\*\eta_q = \*\eta^*_q} 
&= \frac{\mathrm{d}^2}{\mathrm{d} \*\eta_q \mathrm{d} \*\eta^\intercal_q} \int \mathrm{d}\*\theta q(\*\theta;\*\eta_q) \log \frac{p(\bfy_m|\*\theta)}{t_m(\*\theta; \*\eta^*_q)} \biggr|_{\*\eta_q = \*\eta^*_q}
\end{align*}
Summing both sides over all $m$,
\begin{align*}
\sum_m \frac{\mathrm{d}^2 \mathcal{F}_m(q(\*\theta))}{\mathrm{d} \*\eta_q \mathrm{d} \*\eta^\intercal_q} \biggr|_{\*\eta_q = \*\eta^*_q} 
&= \frac{\mathrm{d}^2}{\mathrm{d} \*\eta_q \mathrm{d} \*\eta^\intercal_q} \int \mathrm{d}\*\theta q(\*\theta;\*\eta_q) \log \frac{\prod_m p(\bfy_m|\*\theta)}{\prod_m t_m(\*\theta; \*\eta^*_q)} \biggr|_{\*\eta_q = \*\eta^*_q} \\
&= \frac{\mathrm{d}^2}{\mathrm{d} \*\eta_q \mathrm{d} \*\eta^\intercal_q} \int \mathrm{d}\*\theta q(\*\theta;\*\eta_q) \log \frac{p(\*\theta) \prod_m p(\bfy_m|\*\theta)}{q(\*\theta;\*\eta^*_q)} \biggr|_{\*\eta_q = \*\eta^*_q} \\
&\ - \cancelto{0}{\frac{\mathrm{d}^2}{\mathrm{d} \*\eta_q \mathrm{d} \*\eta^\intercal_q} \int \mathrm{d}\*\theta q(\*\theta;\*\eta_q) \log \mcZ_{q^*} \biggr|_{\*\eta_q = \*\eta_q^*}}.
\end{align*}
\begin{align*}
\frac{\mathrm{d}^2 \mathcal{F}_m(q(\*\theta))}{\mathrm{d} \*\eta_q \mathrm{d} \*\eta^\intercal_q} &= \frac{\mathrm{d}^2}{\mathrm{d} \*\eta_q \mathrm{d} \*\eta^\intercal_q} \int \mathrm{d}\*\theta q(\*\theta;\*\eta_q) \log \frac{q(\*\theta; \*\eta^*_q) p(\bfy_m|\*\theta)}{q(\*\theta;\*\eta_q) t_m(\*\theta; \*\eta^*_q)} \\
&= \frac{\mathrm{d}^2}{\mathrm{d} \*\eta_q \mathrm{d} \*\eta^\intercal_q}  \int \mathrm{d}\*\theta q(\*\theta;\*\eta_q) \log \frac{p(\bfy_m|\*\theta)}{t_m(\*\theta; \*\eta^*_q)} + \frac{\mathrm{d}^2}{\mathrm{d} \*\eta_q \mathrm{d} \*\eta^\intercal_q} \int \mathrm{d}\*\theta q(\*\theta;\*\eta_q) \log \frac{q(\*\theta; \*\eta^*_q)}{q(\*\theta;\*\eta_q)} \\
&= \frac{\mathrm{d}^2}{\mathrm{d} \*\eta_q \mathrm{d} \*\eta^\intercal_q} \int \mathrm{d}\*\theta q(\*\theta;\*\eta_q) \log \frac{p(\bfy_m|\*\theta)}{t_m(\*\theta; \*\eta^*_q)} + \int \mathrm{d}\*\theta \frac{\mathrm{d}^2 q(\*\theta;\*\eta_q)}{\mathrm{d} \*\eta_q \mathrm{d} \*\eta^\intercal_q} \log \frac{q(\*\theta; \*\eta^*_q)}{q(\*\theta;\*\eta_q)} \\
& \qquad \qquad \qquad \qquad \qquad \qquad \qquad \qquad - \cancelto{0}{\frac{\mathrm{d}}{\mathrm{d} \*\eta_q} \int \mathrm{d}\*\theta \frac{\mathrm{d} q(\*\theta;\*\eta_q)}{\mathrm{d} \*\eta^\intercal_q}}.
\end{align*}
At convergence when $\*\eta_q = \*\eta^*_q$,
\begin{align*}
\frac{\mathrm{d}^2 \mathcal{F}_m(q(\*\theta))}{\mathrm{d} \*\eta_q \mathrm{d} \*\eta^\intercal_q} \biggr|_{\*\eta_q = \*\eta^*_q} 
= \frac{\mathrm{d}^2}{\mathrm{d} \*\eta_q \mathrm{d} \*\eta^\intercal_q} \int \mathrm{d}\*\theta q(\*\theta;\*\eta_q) \log \frac{p(\bfy_m|\*\theta)}{t_m(\*\theta; \*\eta^*_q)} \biggr|_{\*\eta_q = \*\eta^*_q}.
\end{align*}
Summing both sides over all $m$,
\begin{align*}
\sum_m \frac{\mathrm{d}^2 \mathcal{F}_m(q(\*\theta))}{\mathrm{d} \*\eta_q \mathrm{d} \*\eta^\intercal_q} \biggr|_{\*\eta_q = \*\eta^*_q} 
&= \frac{\mathrm{d}^2}{\mathrm{d} \*\eta_q \mathrm{d} \*\eta^\intercal_q} \int \mathrm{d}\*\theta q(\*\theta;\*\eta_q) \log \frac{\prod_m p(\bfy_m|\*\theta)}{\prod_m t_m(\*\theta; \*\eta^*_q)} \biggr|_{\*\eta_q = \*\eta^*_q} \\
&= \frac{\mathrm{d}^2}{\mathrm{d} \*\eta_q \mathrm{d} \*\eta^\intercal_q} \int \mathrm{d}\*\theta q(\*\theta;\*\eta_q) \log \frac{p(\*\theta) \prod_m p(\bfy_m|\*\theta)}{q(\*\theta;\*\eta^*_q)} \biggr|_{\*\eta_q = \*\eta^*_q} \\
&\ - \cancelto{0}{\frac{\mathrm{d}^2}{\mathrm{d} \*\eta_q \mathrm{d} \*\eta^\intercal_q} \int \mathrm{d}\*\theta q(\*\theta;\*\eta_q) \log \mcZ_{q^*} \biggr|_{\*\eta_q = \*\eta_q^*}}.
\end{align*}
Now consider the Hessian of the global free-energy $\mathcal{F}(q(\*\theta))$:
\begin{align*}
\frac{\mathrm{d}^2 \mathcal{F}(q(\*\theta))}{\mathrm{d} \*\eta_q \mathrm{d} \*\eta^\intercal_q} &= \frac{\mathrm{d}^2}{\mathrm{d} \*\eta_q \mathrm{d} \*\eta^\intercal_q} \int \mathrm{d}\*\theta q(\*\theta; \*\eta_q) \log \frac{p(\*\theta) \prod_m p(\bfy_m|\*\theta)}{q(\*\theta; \*\eta_q)} \\
&= \frac{\mathrm{d}}{\mathrm{d}\*\eta_q} \left(\int \mathrm{d}\*\theta \frac{\mathrm{d}q(\*\theta)}{\mathrm{d}\*\eta^\intercal_q} \log \frac{p(\*\theta)\prod_m p(\bfy_m | \*\theta)}{q(\*\theta; \*\eta_q)} - \cancelto{0}{\int \mathrm{d}\*\theta q(\*\theta; \*\eta_q) \frac{\mathrm{d} \log q(\*\theta; \*\eta_q)}{\mathrm{d}\*\eta^\intercal_q}} \right) \\
&= \int \mathrm{d}\*\theta \frac{\mathrm{d}^2 q(\*\theta)}{\mathrm{d}\*\eta_q \mathrm{d}\*\eta^\intercal_q} \log \frac{p(\*\theta)\prod_m p(\bfy_m | \*\theta)}{q(\*\theta; \*\eta_q)} - \cancelto{0}{\int \mathrm{d}\*\theta \frac{\mathrm{d}\log q(\*\theta; \*\eta_q)}{\mathrm{d}\*\eta_q} \frac{\mathrm{d}q(\*\theta; \*\eta_q)}{\mathrm{d}\*\eta^\intercal_q}} \\
\end{align*}
Hence,
\begin{align*}
\frac{\mathrm{d}^2 \mathcal{F}(q(\*\theta))}{\mathrm{d} \*\eta_q \mathrm{d} \*\eta^\intercal_q} \biggr|_{\*\eta_q = \*\eta^*_q} 
= \int \mathrm{d}\*\theta \frac{\mathrm{d}^2 q(\*\theta; \*\eta_q)}{\mathrm{d} \*\eta_q \mathrm{d} \*\eta^\intercal_q} \log \frac{p(\*\theta) \prod_m p(\bfy_m|\*\theta)}{q(\*\theta; \*\eta_q)} \biggr|_{\*\eta_q = \*\eta^*_q}
= \sum_m \frac{\mathrm{d}^2 \mathcal{F}_m(q(\*\theta))}{\mathrm{d} \*\eta_q \mathrm{d} \*\eta^\intercal_q} \biggr|_{\*\eta_q = \*\eta^*_q}.
\end{align*}
For all $m$, since $q^*(\*\theta) = \argmax_{q(\*\theta) \in \mathcal{Q}} \mathcal{F}_m(q(\*\theta))$, the Hessian $\frac{\mathrm{d}^2 \mathcal{F}_m(q(\*\theta))}{\mathrm{d} \*\eta_q \mathrm{d} \*\eta^\intercal_q} \big|_{\*\eta_q = \*\eta^*_q}$ is negative definite, which implies that the Hessian $\frac{\mathrm{d}^2 \mathcal{F}(q(\*\theta))}{\mathrm{d} \*\eta_q \mathrm{d} \*\eta^\intercal_q} \big|_{\*\eta_q = \*\eta^*_q}$ of the global free-energy is also negative definite. Therefore, $q^*(\*\theta)$ is a maximum of $\mathcal{F}(q(\*\theta))$.

\subsection{Derivation of Fixed Point Equations: Proof of \Cref{prop:fixed-point-equations}}
\label{sec:appen:fp}
Assume that the approximate likelihoods $\{t_m(\*\theta)\}_{m=1}^M$ are in the un-normalised exponential family. That is, $t_m(\*\theta) = \exp(\*\eta_m^\intercal T(\*\theta) + c_m)$ for some constant $c_m$. In this section we  absorb the constant into the natural parameters $\*\eta^\intercal_m \leftarrow [c_m, \*\eta^\intercal_m]$ and add a corresponding unit element into the sufficient statistics $T(\*\theta)^\intercal \leftarrow [1,T(\*\theta)^\intercal]$. To lighten notation we still denote the sufficient statistic vector as $T(\*\theta)$ so , 
\begin{align}
t_m(\*\theta) = t_m(\*\theta ; \*\eta_m) = \exp(\*\eta_m^\intercal T(\*\theta)), \nonumber
\end{align}
where $\*\eta_m$ is the natural parameters and $T(\*\theta)$ is the augmented sufficient statistics.
We also assume that the prior $p(\*\theta)$ and the variational distributions $q^{(i)}(\*\theta)$ are normalised exponential family distributions.

To derive the fixed point updates for the local variational inference algorithm, we consider maximizing the local variational free energy $\mcF^{(i)}_k(q(\*\theta))$ in \eqref{eq:vfe}. Assuming that at iteration $i$, the variational distribution $q^{(i-1)}(\*\theta)$ obtained from the previous iteration has the form:
\begin{align}
q^{(i-1)}(\*\theta) = \exp( {\*\eta^{(i-1)}_q}^\intercal T(\*\theta) - A(\*\eta^{(i-1)}_q)), \nonumber
\end{align}
and the target distribution $q(\*\theta)$ that we optimise has the form:
\begin{align}
q(\*\theta) = \exp( \*\eta_q^\intercal T(\*\theta) - A(\*\eta_q) ), \nonumber
\end{align}
where $A(\cdot)$ is the log-partition function. Let ${ \*\eta = \*\eta^{(i-1)}_q - \*\eta_q - \*\eta^{(i-1)}_{k} }$, we can write the local free-energy as:
\begin{align*}
\mcF^{(i)}_k(q(\*\theta)) &= A(\*\eta_q) - A(\*\eta^{(i-1)}_q) + \Exp{q}{\log p(\bfy_{k}|\*\theta)} + \*\eta^\intercal \Exp{q}{T(\*\theta)}.
\end{align*}
Take the derivative of $\mcF^{(i)}_k(q(\*\theta))$ w.r.t.~$\*\eta_q$ and note that $\frac{\mathrm{d}A(\*\eta_q)}{\mathrm{d}\*\eta_q} = \Exp{q}{T(\*\theta)}$, we have:
\begin{align*}
\frac{\mathrm{d}\mcF^{(i)}_k(q(\*\theta))}{\mathrm{d}\*\eta_q}  &= \frac{\mathrm{d}}{\mathrm{d}\*\eta_q} \Exp{q}{\log p(\bfy_{k}|\*\theta)} + \frac{\mathrm{d}^2A(\*\eta_q)}{\mathrm{d}\*\eta_q\mathrm{d}\*\eta_q} \*\eta.
\end{align*}
Let $\*\eta_q = \*\eta^{(i)}_{q_k}$ when this derivative is equal to zero, such that
\begin{align}
\label{eq:fixed-point-q}
\*\eta^{(i)}_{q_k}  &= \left.\mathbb{C}^{-1}\frac{\mathrm{d}}{\mathrm{d}\*\eta_q} \Exp{q}{\log p(\bfy_{k}|\*\theta)}\right|_{\*\eta_q = \*\eta^{(i)}_{q_k}} + \*\eta^{(i-1)}_q - \*\eta^{(i-1)}_{k},
\end{align}
where $\mathbb{C} \defeq \frac{\mathrm{d}^2A(\*\eta_q)}{\mathrm{d}\*\eta_q\mathrm{d}\*\eta_q} = \mathrm{cov}_{q(\*\theta)}[ T(\*\theta) T^\intercal(\*\theta)]$ is the Fisher Information.
Note that $t^{(i)}_k(\*\theta) = \frac{q^{(i)}_k(\*\theta)}{q^{(i-1)}(\*\theta)} t^{(i-1)}_k(\*\theta)$ implies $\*\eta_k^{(i)} = \*\eta^{(i)}_{q_k} - \*\eta^{(i-1)}_q + \*\eta_k^{(i-1)}$. Substituting this into \eqref{eq:fixed-point-q} results in the following local fixed-point update for \Cref{alg:vmp}:
\begin{align}
\*\eta^{(i)}_{k}  &= \left. \mathbb{C}^{-1} \frac{\mathrm{d}}{\mathrm{d}\*\eta_q} \Exp{q}{\log p(\bfy_{k}|\*\theta)}\right|_{q = q^{(i)}_k}. \nonumber
\end{align}
It is worth emphasising that $\*\eta^{(i)}_{q_k}$ is not the natural parameters of the approximate posterior retained by the server, but rather the natural parameters defined by $\*\eta^{(i)}_{q_k} = \*\eta^{(i-1)}_q - \*\eta^{(i-1)}_k + \*\eta^{(i)}_k$.\footnote{The difference is important in synchronous and asynchronous PVI, where in general $q^{(i)}(\*\theta) \neq q^{(i)}_k(\*\theta)$.}
The Fisher Information can be written as $\mathbb{C} = \frac{\mathrm{d}\*\mu_q}{\mathrm{d}\*\eta_q}$ where $\*\mu_q = \Exp{q}{T(\*\theta)}$ is the mean parameter of $q(\*\theta)$. This leads to a cancellation of the Fisher information,
\begin{align}
\*\eta^{(i)}_{k}  &
= \left. \frac{\mathrm{d}}{\mathrm{d}\*\mu_q} \Exp{q}{\log p(\bfy_{k}|\*\theta)}\right|_{q = q^{(i)}_k}. \nonumber
\end{align}

\subsection{Equivalence of Local and Global Gradient-Based Updates: Proof of \Cref{prop:local-global-gradient}}
\label{sec:local-global-gradient}
Here we show that using SGD to perform a single update of the local optimisation step in synchronous PVI has equivalent dynamics as using SGD to perform a single update of global VI. Applying SGD to the PVI local optimisation step involves the updates
\begin{equation}
    \left[\*\eta^{(i)}_{q_k}\right]_t = \left[\*\eta^{(i)}_{q_k}\right]_{t - 1} + \rho \left.\frac{\mathrm{d}}{\mathrm{d} \*\eta_q} \mcF_k(q(\*\theta)) \right|_{\*\eta_q = \left[\*\eta^{(i)}_{q_k}\right]_{t-1}}. \nonumber
\end{equation}
where $\left[\*\eta^{(i)}_{q_k}\right]_t = \*\eta^{(i-1)}_q - \*\eta^{(i-1)}_k + \left[\*\eta^{(i)}_k\right]_t$. Note that $t$ indexes the inner loop of damped fixed-point iterations. Initialising $\left[\*\eta^{(i)}_k\right]_0 = \*\eta^{(i-1)}_k$ results in the following initial update step 
\begin{equation}
    \left[\*\eta^{(i)}_{q_k}\right]_1 = \*\eta^{(i-1)}_q + \rho \left.\frac{\mathrm{d}}{\mathrm{d} \*\eta_q} \mcF_k(q(\*\theta)) \right|_{\*\eta_q = \*\eta^{(i-1)}_q}, \nonumber
\end{equation}
or equivalently,
\begin{equation}
    \left[\*\eta^{(i)}_k\right]_1 = \*\eta^{(i-1)}_k + \rho \left.\frac{\mathrm{d}}{\mathrm{d} \*\eta_q} \mcF_k(q(\*\theta)) \right|_{\*\eta_q = \*\eta^{(i-1)}_q}. \nonumber
\end{equation}
When a single iteration of synchronous gradient-based updates of all the natural parameters is performed, the natural parameters of $q^{(i)}(\*\theta)$ are updated as follows,
\begin{equation}
\begin{aligned}
    \*\eta^{(i)}_q &= \*\eta_0 + \sum_{m=1}^M \*\eta^{(i-1)}_m + \rho \left.\frac{\mathrm{d}}{\mathrm{d} \*\eta_q} \mcF_m(\*\eta_q) \right|_{\*\eta_q = \*\eta^{(i-1)}_q} \\
    &= \*\eta^{(i-1)}_q + \rho \frac{\mathrm{d}}{\mathrm{d} \*\eta_q} \left. \sum_{m=1}^M \mcF_m(\*\eta_q) \right|_{\*\eta_q = \*\eta^{(i-1)}_q}.
\end{aligned} \nonumber
\end{equation}
From \Cref{sec:appen:local-global}, we have
\begin{align*}
    \left.\frac{\mathrm{d} \mathcal{F}_m(\*\eta)}{\mathrm{d} \*\eta_q}\right|_{\*\eta_q = \*\eta^{(i-1)}_q} = \left. \frac{\mathrm{d}}{\mathrm{d} \*\eta_q} \int \mathrm{d}\*\theta q(\*\theta) \log \frac{p(\bfy_m|\*\theta)}{t^{(i-1)}_m(\*\theta)} \right|_{\*\eta_q = \*\eta^{(i-1)}_q}.
\end{align*}
Summing over all $m$ gives
\begin{align*}
    \left. \sum_{m=1}^M \frac{\mathrm{d} \mathcal{F}_m(\*\eta)}{\mathrm{d} \*\eta_q} \right|_{\*\eta_q = \*\eta^{(i-1)}_q} &= \left. \frac{\mathrm{d}}{\mathrm{d}\*\eta_q} \int q(\*\eta) \log \frac{\prod_m p(\bfy_m | \*\theta)}{t^{(i-1)}_m(\*\theta)} \mathrm{d}\*\theta \right|_{\*\eta_q = \*\eta^{(i-1)}_q} \\
    &= \left. \frac{\mathrm{d}}{\mathrm{d}\*\eta_q} \int q(\*\eta) \log \frac{p(\*\theta)\prod_n p(y_n | \*\theta)}{q^{(i-1)}(\*\theta)} \mathrm{d}\*\theta \right|_{\*\eta_q = \*\eta^{(i-1)}_q}.
\end{align*}
The derivative of the global free-energy evaluated at $\*\eta_q = \*\eta^{(i-1)}_q$ is also given by
\begin{align*}
    \frac{\mathrm{d} \mathcal{F}(q(\*\theta))}{\mathrm{d} \*\eta_q} \biggr|_{\*\eta_q = \*\eta^{(i-1)}_q} = \int \mathrm{d}\*\theta \frac{\mathrm{d} q(\*\theta)}{\mathrm{d} \*\eta_q} \log \frac{p(\*\theta) \prod_n p(y_n|\*\theta)}{q^{(i-1)}(\*\theta)} \biggr|_{\*\eta_q = \*\eta^{(i-1)}_q} = \sum_m \frac{\mathrm{d} \mathcal{F}_m(q(\*\theta))}{\mathrm{d} \*\eta_q} \biggr|_{\*\eta_q = \*\eta^{(i-1)}_q}.
\end{align*}
Thus, the update for natural parameters of $q^{(i)}(\*\theta)$ can equivalently be expressed as
\begin{equation}
    \*\eta^{(i)}_q = \*\eta^{(i-1)}_q + \rho \left.\frac{\mathrm{d}}{\mathrm{d} \*\eta_q} \mcF(\*\eta_q) \right|_{\*\eta_q = \*\eta^{(i-1)}_q} \nonumber
\end{equation}
which is exactly the same as performing a single SGD update of global VI.

\subsection{Equivalence of Local and Global Iterative Fixed-Point Updates: Proof of \Cref{prop:local-global-fp}}
\label{sec:local-global-same}
Here we show that running a single iteration of damped fixed-point updates for synchronous PVI has equivalent dynamics for $q^{(i)}(\*\theta)$---the approximate posterior retained by the server---as running damped fixed-point updates for batch VI ($M = 1$). The PVI damped iterative fixed-point updates (\Cref{prop:fixed-point-equations}) are given by,
\begin{equation}
    \left[\*\eta^{(i)}_k\right]_t = (1 - \rho)\left[\*\eta^{(i)}_k\right]_{t-1} + \rho \left. \frac{\mathrm{d}}{\mathrm{d}\*\mu_q} \Exp{q}{\log p(\bfy_{k}|\*\theta)}\right|_{q = \left[q^{(i)}_k\right]_{t-1}}. \nonumber
\end{equation}
where $\left[\*\eta^{(i)}_{q_k}\right]_t = \*\eta^{(i-1)}_q - \*\eta^{(i-1)}_k + \left[\*\eta^{(i)}_k\right]_t$. Initialising $\left[\*\eta^{(i)}_k\right]_{0} = \*\eta^{(i-1)}_k$ results in the following initial updates
\begin{equation}
    \left[\*\eta^{(i)}_k\right]_1 = (1 - \rho)\*\eta^{(i-1)}_k + \rho \left. \frac{\mathrm{d}}{\mathrm{d}\*\mu_q} \Exp{q}{\log p(\bfy_{k}|\*\theta)}\right|_{q = q^{(i-1)}}. \nonumber
\end{equation}
where we have used the fact that $\left[q^{(i)}_k\right]_0 = q^{(i-1)}$.
We have explicitly denoted the dependence of the approximate posterior on the iteration number $i$ as the dynamics of this is the focus. When a single iteration of synchronous fixed-point updates of all the natural parameters is performed, such that $\{\*\eta^{(i)}_m = \left[\*\eta^{(i)}_m\right]_1\}_{m=1}^M$, the natural parameters of $q(\*\theta)$ are updated as follows,
\begin{equation}
\begin{aligned}
    \*\eta^{(i)}_q &= \*\eta_0 + \sum_{m=1}^M (1 - \rho)\*\eta^{(i-1)}_m + \rho \left.\frac{\mathrm{d}}{\mathrm{d}\*\mu_q} \Exp{q}{\log p(\bfy_{m}|\*\theta)}\right|_{q = q^{(i-1)}} \\
    &= (1 - \rho) \*\eta^{(i-1)}_q + \rho \left(\*\eta_0 + \sum_{n=1}^N \left.\frac{\mathrm{d}}{\mathrm{d}\*\mu_{q}} \Exp{q}{\log p(y_{n}|\*\theta)}\right|_{q = q^{(i-1)}}\right).
\end{aligned} \nonumber
\end{equation}
Here, in the last line, we have used the fact that the data are independent conditioned on $\*\theta$. Now consider application of a single iteration of the damped fixed-point updates to the batch case ($M=1$) 
\begin{equation}
\begin{aligned}
    \*\eta_q^{(i)} &= (1 - \rho)\*\eta^{(i-1)}_q + \rho \left( \*\eta_0 + \left. \frac{\mathrm{d}}{\mathrm{d}\*\mu_{q}} \Exp{q}{\log p(\bfy | \*\theta)}\right|_{q = q^{(i-1)}} \right) \\
    &= (1 - \rho) \*\eta^{(i-1)}_q + \rho \left(\*\eta_0 + \sum_{n=1}^N \left.\frac{\mathrm{d}}{\mathrm{d}\*\mu_{q}} \Exp{q}{\log p(y_{n}|\*\theta)}\right|_{q = q^{(i-1)}}\right).
\end{aligned} \nonumber
\end{equation}
Therefore the updates for $q(\*\theta)$ are identical in the two cases.

\subsection{Relations Between Methods Employing Stochastic Approximation of Fixed-Points / Natural Gradient Updates} \label{sec:stochastic-approx}

There are two distinct ways to apply stochastic approximations to PVI fixed-point updates. The first considers randomly sampled mini-batches of data $\bfy_{m, b} \stackrel{\text{iid}}{\sim}p_m(y)$ where $p_m(y) = \frac{1}{N_m} \sum_{n=1}^{N_m} \delta(y - y_{m, n})$, whereas the second considers fixed sub-groups of data $\{\bfy_{m, l}\}_{l=1}^L$. \footnote{We have used a distinct notation for a mini-batch ($\bfy_l$) and a data group ($\bfy_m$) since the former will be selected iid from the data set and will vary at each epoch, whilst the latter need not be determined in this way and is fixed across epochs.} To simplify the notation, we consider the global VI case $M=1$ and denote the mini-batches and sub-groups of data $\bfy_b$ and $\bfy_l$, respectively.

\subsubsection{Stochastic Approximation Within the Local Free-Energy}
\label{sec:stochastic-approx:stoc_approx_1}
The first form of stochastic approximation leverages the fact that each local free-energy decomposes into a sum over data points and can, therefore, be approximated by sampling mini-batches within each data partition. In the case where each partition includes a large number of data points, this leads to algorithms that converge more quickly than the batch variants---since a reasonable update for the approximate posterior can often be determined from just a few data points---and this faster convergence opens the door to processing larger data sets.

In the global VI case, the damped simplified fixed-point updates are
\begin{align}
\*\eta^{(i)}_{q}  &
= (1-\rho)\*\eta^{(i-1)}_{q} + \rho \left ( \*\eta_0 + \left. \frac{\mathrm{d}}{\mathrm{d}\*\mu_q} \Exp{q}{\log p(\bfy|\*\theta)}\right|_{q=q^{(i-1)}} \right). \nonumber
\end{align}
\citet{hoffman+al:2013,sheth+khardon:2016b} employ stochastic mini-batch approximation of $\Exp{q}{\log p(\bfy|\*\theta)} = \sum_{n} \Exp{q(\*\theta)}{\log p(y_n|\*\theta)} = N \Exp{p_{\text{data}}(y),q(\*\theta)}{\log p(y_n|\*\theta)} $ by sub-sampling the data distribution $\mathbf{y}_b \stackrel{\text{iid}}{\sim} p_{\text{data}}(y)$ where $p_{\text{data}}(y) = \frac{1}{N} \sum_{n=1}^N \delta(y - y_n)$. This approximation yields
\begin{align}
\*\eta^{(i)}_{q}  &
= (1-\rho)\*\eta^{(i-1)}_{q} + \rho \left ( \*\eta_0 + \left.B \frac{\mathrm{d}}{\mathrm{d}\*\mu_q} \Exp{q}{\log p(\bfy_b|\*\theta)}\right|_{q=q^{(i-1)}} \right). \label{eq:stoch-global}
\end{align}
where $\bfy_b$ are a mini-batch containing $N/B$ data points.  \citet{li+al:2015} show that their stochastic power EP algorithm recovers precisely these same updates when $\alpha \rightarrow 0$ (this is related to \Cref{prop:pep-fp}). The relation to EP-like algorithms can be made more explicit by writing \eqref{eq:stoch-global} as
\begin{align}
\*\eta^{(i)}_{q}  &
= \*\eta^{(i-1)}_{q} + \rho' \left (   \left. \frac{\mathrm{d}}{\mathrm{d}\*\mu_q} \Exp{q}{\log p(\bfy_b|\*\theta)}\right|_{q=q^{(i-1)}} - \*\eta^{(i-1)}_{\text{like}}/B \right). \label{eq:stoch-global-EP}
\end{align}
Where $\rho' = \rho B$ is a rescaled learning rate and $\*\eta^{(i-1)}_{\text{like}} = \*\eta^{(i-1)}_{q} - \*\eta_{0}$ is the portion of the approximate posterior natural parameters that approximates the likelihoods. As such, $\*\eta^{(i-1)}_{\text{like}}/B$ is the contribution a mini-batch likelihood makes on average to the posterior. 
So, interpreting the update in these terms, the new approximate posterior is equal to the old approximate posterior with $\rho'$ of the average mini-batch likelihood approximation removed and $\rho'$ of the approximation from the current mini-batch, $\left.\frac{\mathrm{d}}{\mathrm{d}\*\mu_q} \Exp{q}{\log p(\bfy_b|\*\theta)}\right|_{q=q^{(i-1)}}$, added in place of it.

\subsubsection{Stochastic Scheduling of Updates Between Local Free-Energies}
\label{sec:stochastic-approx:stoc_approx_2}
\citet{khan+li:2018} take a different approach, employing damped simplified fixed-point updates for local VI and then using an update schedule that selects a mini-batch at random and then updates the local natural parameters for these data-points in parallel. That is for all data points in the mini-batch $\bfy_b$ they apply a single iteration of the damped simplified fixed-point updates
\begin{align}
\*\eta^{(i)}_{n}  & = (1-\rho) \*\eta^{(i-1)}_{n} + \rho \left.\frac{\mathrm{d}}{\mathrm{d}\*\mu_{q}} \Exp{q} {\log p(y_{n}|\*\theta)}\right|_{q=q^{(i-1)}} \label{eq:stoch-fully-local}
\end{align}
and update the global parameters as $\*\eta^{(i)}_q = \*\eta_0 + \sum_{n: y_n \in \bfy_b} \*\eta_n^{(i)} + \sum_{n': y_{n'} \notin \bfy_b} \*\eta^{(i-1)}_{n'}$. This local update incurs a  memory overhead that is $N$ times larger due to the need maintain $N$ sets of local parameters rather than just one. A more memory efficient approach is to fix the mini-batches across epochs and to visit the sub-groups $\left\{\bfy_l\right\}_{l=1}^L$ in random order. Doing so requires storing $L$ natural parameters instead, corresponding to one for each mini-batch. For simplified fixed point updates, this yields
\begin{align}
\*\eta^{(i)}_{l}  & = (1-\rho) \*\eta^{(i-1)}_{l} + \rho \left.\frac{\mathrm{d}}{\mathrm{d}\*\mu_{q}} \Exp{q}{\log p(\bfy_{l}|\*\theta)}\right|_{q=q^{(i-1)}} \label{eq:stoch-local}
\end{align}
with the global parameters updated as $\*\eta^{(i)}_q = \*\eta_0 + \*\eta^{(i)}_l + \sum_{p\neq l} \*\eta^{(i-1)}_p$. Interestingly, both of these local updates (\eqref{eq:stoch-fully-local} and \eqref{eq:stoch-local}) result in a subtly different update to $q(\*\theta)$ as the stochastic global update above (\eqref{eq:stoch-global-EP}),
\begin{align}
    \*\eta^{(i)}_{q}  &
    = \*\eta^{(i-1)}_{q} + \rho \left ( \left.\frac{\mathrm{d}}{\mathrm{d}\*\mu_q} \Exp{q}{\log p(\bfy_l|\*\theta)}\right|_{q=q^{(i-1)}} - \*\eta^{(i-1)}_{l} \right)\label{eq:stoch-local-EP}
\end{align}
where we have used $\bfy_l$ to denote both the sub-group and mini-batch of data. Here the deletion step is explicitly revealed again, but now it involves removing the natural parameters of the $l$th approximate likelihood $\*\eta^{(i-1)}_{l}$ rather than the average mini-batch likelihood approximation $\*\eta^{(i-1)}_{\text{like}}/B$. If the first approach in \eqref{eq:stoch-local} employs learning rates that obey the Robins Munro conditions, the fixed points will be identical to the second approach in \eqref{eq:stoch-local-EP} and they will correspond to optima of the global free-energy. Note that \eqref{eq:stoch-local-EP} is equivalent to employing PVI and terminating the local optimisation step for sub-group $\bfy_l$ after a single iteration of the simplified damped fixed-point update. As discussed in \Cref{sec:optim}, in the federated learning setting running multiple iterations of the damped simplified fixed-point updates can reduce the number of communications required. However, since we are considering stochastic approximations within local free-energy optimisations all sub-groups $\bfy_l$ are assumed to be stored on the same device and so the communication efficiency is irrelevant. 



\subsubsection{Comparing and Contrasting Stochastic Approximations}
\label{sec:stochastic-approx:comparing_stoc}
Each variety of update has its own pros and cons (compare \eqref{eq:stoch-global-EP} to \eqref{eq:stoch-local-EP}). The stochastic global update \eqref{eq:stoch-global-EP} does not support general online learning and distributed asynchronous updates, but it is more memory efficient and faster to converge in the batch setting. 

Consider applying both methods to online learning where $\bfy_b$ or $\bfy_l$ correspond to the new data seen at each stage and $\rho'$ and $\rho$ are user-determined learning rates for the two algorithms. General online learning poses two challenges for the stochastic global update \eqref{eq:stoch-global-EP}. First, the data are typically not iid (due to covariate or data set shift over time). Second, general online learning does not allow all old data points to be revisited, demanding incremental updates instead. This means that when a new batch of data is received we must iterate on just these data to refine the approximate posterior, before moving on and potentially never returning. Iterating \eqref{eq:stoch-global-EP} on this new data is possible, but it would have disastrous consequences as it again breaks the iid mini-batch assumption and would just result in $q(\*\theta)$ fitting the new data and forgetting the old previously seen data. A single update could be made, but this will normally mean that the approach is data-inefficient and slow to learn. On the other hand, iterating local updates \eqref{eq:stoch-fully-local} or \eqref{eq:stoch-local} works nicely as these naturally incorporate a deletion step that removes just the contribution from the current mini-batch and can, therefore, be iterated to our heart's content.


The stochastic global update \eqref{eq:stoch-global} does have two important advantages over the local updates \eqref{eq:stoch-local-EP}. First, the memory footprint is $L$ times smaller only requiring a single set of natural parameters to be maintained rather than $L$ of them. Second, it can be faster to converge when the mini-batches are iid. Contrast what happens when new data are seen for the first time in the two approaches. For simplicity assume $\rho' = \rho =1$. In the second approach, $\*\eta^{(i-1)}_{l} = \*0$ as the data have not been seen before, but the first approach effectively uses $\*\eta^{(i-1)}_{l} = \*\eta^{(i-1)}_{\text{like}}/L$. That is, the first approach effectively estimates the approximate likelihood for new data, based on those for previously seen data. This is a sensible approximation for homogeneous mini-batches. A consequence of this is that the learning rate $\rho'$ can be much larger than $\rho$ (potentially greater than unity) resulting in faster convergence of the approximate posterior. 
It would be interesting to consider modifications of the local updates \eqref{eq:stoch-local-EP} that estimate the $l$th approximate likelihood based on information from all other data partitions. For example, in the first pass through the data, the approximate likelihoods for unprocessed mini-batches could be updated to be equal to the last approximate likelihood or to the geometric average of previous approximate likelihoods. Alternatively, ideas from inference networks could be employed for this purpose.

\subsection{The Relationship Between Natural Gradient, Mirror-Descent,  Trust-Region and Proximal Methods}
\label{sec:nat-grad-mirror-trust}

Each step of gradient ascent of parameters $\*\eta$ on a cost $\mathcal{C}$ can be interpreted as the result of an optimisation problem derived from linearizing the cost function around the old parameter estimate, $\mathcal{C}(\*\eta) \approx \mathcal{C}(\*\eta^{(i-1)}) + \nabla_{\*\eta} \mathcal{C}(\*\eta^{(i-1)}) (\*\eta - \*\eta^{(i-1)}) $ where $\nabla_{\*\eta} \mathcal{C}(\*\eta^{(i-1)}) = \frac{\mathrm{d}\mathcal{C}(\*\eta)}{\mathrm{d}\*\eta} |_{\*\eta = \*\eta^{(i-1)}}$ and adding a soft constraint on the norm of the parameters: 
\begin{align}
\*\eta^{(i)} = \*\eta^{(i-1)} + \rho \frac{\mathrm{d}\mathcal{C}(\*\eta)}{\mathrm{d}\*\eta} \;\; \Leftrightarrow \;\; \*\eta^{(i)} = \argmax_{ \*\eta }  \nabla_{\*\eta} \mathcal{C}(\*\eta^{(i-1)}) \*\eta - \frac{1}{2 \rho} || \*\eta - \*\eta^{(i-1)} ||_2^2. \nonumber
\end{align}
Here the terms in the linearised cost that do not depend on $\*\eta$ have been dropped as they do not effect the solution of the optimisation problem.  The linearisation of the cost ensures that there is an analytic solution to the optimisation and the soft constraint ensures that we do not move too far from the previous setting of the parameters into a region where the linearisation is inaccurate.  

This reframing of gradient ascent reveals that it is making an Euclidean assumption about the geometry parameter space and suggests generalisations of the procedure that are suitable for different geometries. In our case, the parameters are natural parameters of a distribution and measures of proximity that employ the KL divergence are natural. 

The main result of this section is that the following optimisation problems:
\begin{align}
&\text{\it KL proximal method:} \; \; \*\eta_q^{(i)} = \argmax_{ \*\eta_q }  \;  \nabla_{\*\eta} \mcF^{(i)}_k(q^{(i-1)}(\*\theta)) \*\eta_q - \frac{1}{\rho} \mathrm{KL} \left( q(\*\theta) ~\|~ q^{(i-1)}(\*\theta) \right). \label{eq-opt1}\\
&\text{\it KL trust region:} \; \; \*\eta_q^{(i)} = \argmax_{ \*\eta_q }  \;  \nabla_{\*\eta} \mcF^{(i)}_k(q^{(i-1)}(\*\theta)) \*\eta_q  \; \text{s.t.} \; \mathrm{KL} \left( q(\*\theta) ~\|~ q^{(i-1)}(\*\theta) \right) \le \gamma. \label{eq-opt2}\\
&\text{\it KL$^s$ trust region:} \; \; \*\eta_q^{(i)} = \argmax_{ \*\eta_q }  \;  \nabla_{\*\eta} \mcF^{(i)}_k(q^{(i-1)}(\*\theta)) \*\eta_q  \; \text{s.t.} \; \mathrm{KL}^{\text{s}} \left( q(\*\theta) ~\|~ q^{(i-1)}(\*\theta) \right) \le \gamma. \label{eq-opt3}\\
&\text{\it Mirror descent:} \; \; \*\mu_q^{(i)} = \argmax_{ \*\mu_q }  \; \nabla_{\*\mu} \mcF^{(i)}_k(q^{(i-1)}(\*\theta)) \*\mu_q  - \frac{1}{\rho} \mathrm{KL} \left( q^{(i-1)}(\*\theta)  ~\|~  q(\*\theta)\right). \nonumber
\end{align}
All yield the same updates as the damped fixed point equations / natural gradient ascent:
\begin{align*}
\*\eta^{(i)}_{k}  &
= (1-\rho)\*\eta^{(i-1)}_{k} + \rho \frac{\mathrm{d}}{\mathrm{d}\*\mu^{(i-1)}_q} \Exp{q^{(i-1)}}{\log p(\bfy_{k}|\*\theta)}\\
& = (1-\rho)\*\eta^{(i-1)}_{k} + \rho \left [ \frac{\mathrm{d} \*\mu^{(i-1)}_q}{\mathrm{d}\*\eta^{(i-1)}_q} \right ]^{-1} \frac{\mathrm{d}}{\mathrm{d}\*\eta^{(i-1)}_q} \Exp{q^{(i-1)}}{\log p(\bfy_{k}|\*\theta)}.
\end{align*}
In the first three cases (\eqref{eq-opt1} - \eqref{eq-opt3}), this equivalence only holds exactly in the general case if the parameter changes $\Delta \*\eta_q^{(i)} =  \*\eta_q^{(i)}-\*\eta_q^{(i-1)}$ are small.

Here the {\it KL proximal method} is the straightforward generalisation of the gradient ascent example that replaces the Euclidean norm by the exclusive KL divergence. The {\it KL trust region method}  uses a hard constraint on the same KL instead, but rewriting this as a Lagrangian recovers the KL proximal method with $1/\rho$ being the Lagrange multiplier. The {\it KL$^s$ trust region method},  often used to justify natural gradient ascent,  employs the symmetrised KL divergence instead of the exclusive KL. The symmetrised KL is the average of the exclusive and inclusive KLs, $\mathrm{KL}^{\text{s}} \left( q(\*\theta) ~\|~ q^{(i-1)}(\*\theta) \right) = \frac{1}{2} \mathrm{KL} \left( q(\*\theta) ~\|~ q^{(i-1)}(\*\theta) \right) +  \frac{1}{2} \mathrm{KL}\left( q^{(i-1)}(\*\theta)  ~\|~  q(\*\theta)\right)$.  {\it Mirror descent}, in its most general form, uses a Bregman divergence to control the extent to which the parameters change rather than a KL divergence. However, when applied to exponential families, the Bregman divergence becomes the inclusive KL divergence yielding the form above \citep{raskutti+mukherjee:2015,khan+al:2016,khan+li:2018}. Note that this last method operates in the mean parameter space and the equivalence is attained by mapping the mean parameter updates back to the natural parameters. Mirror descent has the advantage of not relying on the small parameter change assumption to recover natural gradient ascent. Having explained the rationale behind these approaches we will now sketch how they yield the fixed-point updates.

The equivalence of the {\it KL proximal method} can be shown by differentiating the cost w.r.t. $ \*\eta_q $  and  substituting in the following expressions:
\begin{align*}
\nabla_{\*\eta} \mcF^{(i)}_k(q^{(i-1)}(\*\theta)) &= \frac{\mathrm{d}}{\mathrm{d}\*\eta_{q^{(i-1)}}} \Exp{q^{(i-1)}}{\log p(\bfy_{k}|\*\theta)} - \frac{\mathrm{d} \*\mu_{q^{(i-1)}}}{\mathrm{d}\*\eta_{q^{(i-1)}}} \*\eta^{(i-1)}_{k},\\
\frac{\mathrm{d}\mathrm{KL} \left( q(\*\theta) ~\|~ q^{(i-1)}(\*\theta) \right) }{\mathrm{d}\*\eta_q}  &= \frac{\mathrm{d} \*\mu_{q}}{\mathrm{d}\*\eta_{q}} \left ( \*\eta_{q} - \*\eta_{q}^{(i-1)} \right) \approx  \frac{\mathrm{d} \*\mu_{q^{(i-1)}}}{\mathrm{d}\*\eta_{q^{(i-1)}}} \left ( \*\eta_{q} - \*\eta_{q}^{(i-1)} \right) .
\end{align*}
In the second line above the approximation results from the assumption of small parameter change $\Delta \*\eta_q^{(i)}$ (or alternatively local constancy of the Fisher information). Equating the derivatives to zero and rearranging recovers the fixed point equations.

The equivalence of the {\it KL trust region method} is now simple to show as the associated Lagrangian, $\mathcal{L}(\*\eta_q) = \nabla_{\*\eta} \mcF^{(i)}_k(q^{(i-1)}(\*\theta)) \*\eta_q  - \frac{1}{\rho} \left( \mathrm{KL} \left( q(\*\theta) ~\|~ q^{(i-1)}(\*\theta) \right) - \gamma \right)$, is the {\it proximal method} up to an additive constant.

Similarly, the {\it KL$^s$ trust region method} can also be rewritten as a Lagrangian $\mathcal{L}(\*\eta_q) = \nabla_{\*\eta} \mcF^{(i)}_k(q^{(i-1)}(\*\theta)) \*\eta_q  - \frac{1}{\rho} \left( \mathrm{KL}^{\text{s}} \left( q(\*\theta) ~\|~ q^{(i-1)}(\*\theta) \right) - \gamma \right)$. For small changes in the approximate posterior natural parameters $\Delta \*\eta_q^{(i)} =  \*\eta_q^{(i)}-\*\eta_q^{(i-1)}$, the symmetrised KL can be approximated using a second order Taylor expansion, 
\begin{align}
\mathrm{KL}^{\text{s}} \left( q(\*\theta) ~\|~ q^{(i-1)}(\*\theta) \right) \approx \frac{1}{2} \left ( \*\eta_{q} - \*\eta_{q}^{(i-1)} \right)^{\top} \frac{\mathrm{d} \*\mu_{q^{(i-1)}}}{\mathrm{d}\*\eta_{q^{(i-1)}}} \left ( \*\eta_{q} - \*\eta_{q}^{(i-1)} \right). \nonumber
\end{align}
This is the same form as the exclusive KL takes, the inclusive and exclusive KL divergences being locally identical around their optima. Taking derivatives of the Lagrangian and setting them to zero recovers the fixed point equations again. 

The {\it mirror descent} method can be shown to yield the fixed points by noting that 
\begin{align*}
\nabla_{\*\mu} \mcF^{(i)}_k(q^{(i-1)}(\*\theta))  &= \frac{\mathrm{d}}{\mathrm{d}\*\mu_{q^{(i-1)}}} \Exp{q^{(i-1)}}{\log p(\bfy_{k}|\*\theta)} -  \*\eta_{k},\\
\frac{\mathrm{d}\mathrm{KL} \left( q^{(i-1)}(\*\theta) ~\|~ q(\*\theta) \right) }{\mathrm{d}\*\mu_q}  &=  \*\eta_{q} - \*\eta_{q}^{(i-1)} .
\end{align*}
Differentiating the mirror descent objective and substituting these results in recovers usual update. The last result above can be found using convex duality. For a full derivation and more information on the relationship between mirror descent and natural gradients see \cite{raskutti+mukherjee:2015}.

It is also possible to define optimisation approaches analogous to the above that do not linearise the free-energy term and instead perform potentially multiple updates of the nested non-linear optimisation problems \citep{theis+hoffman:2015, khan+al:2015, khan+al:2016, khan+li:2018}.

\subsection{The Equivalence of the Simplified Fixed-Point Updates and Power EP $\alpha \rightarrow 0$: Proof of \Cref{prop:pep-fp}}
\label{proof:pep-fp}
Assume PEP is using unnormalised distributions throughout:
\begin{enumerate}
\item $p^{(i)}_{\alpha}(\*\theta) = q^{(i-1)}(\*\theta) \left( \frac{p(\mathbf{y}_{k} | \*\theta)}{t_{k}^{(i-1)}(\*\theta)}\right)^{\alpha} $ \textrm{\% form tilted}
\item $q_{\alpha}(\*\theta) = \mathrm{proj}( p^{(i)}_{\alpha}(\*\theta) )$  \% moment match 
\item $q^{(i)}(\*\theta) = \left ( q^{(i-1)}(\*\theta) \right )^{1-1/\alpha} \left( q_{\alpha}(\*\theta) \right)^{1/\alpha}$ \% update posterior
\item $t^{(i)}_{k}(\*\theta) = \frac{q^{(i)}(\*\theta)}{q^{(i-1)}(\*\theta)} t^{(i-1)}_{k}(\*\theta) $ \% update approximate likelihood.
\end{enumerate}
Note that we have used the shorthand notation $p^{(i)}_{k, \alpha} = p^{(i)}_{\alpha}$, $q^{(i)}_{k, \alpha} = q_{\alpha}$ and $q^{(i)}_k = q^{(i)}$.
From (2,3,4), we have:
\begin{align}
\log t^{(i)}_{k}(\*\theta) = \frac{1}{\alpha} \left( \log \mathrm{proj}( p^{(i)}_{\alpha}(\*\theta) ) - \log q^{(i-1)}(\*\theta) \right) + \log t^{(i-1)}_{k} (\*\theta). \label{eq:log-ti-1}
\end{align}
Using the augmented sufficient statistics $T(\*\theta)$ as in \Cref{sec:appen:fp} and the fact that $q_{\alpha}(\*\theta)$ is unnormalised, we can write:
\begin{align}
\log q_{\alpha}(\*\theta) = \log \mathrm{proj}( p^{(i)}_{\alpha}(\*\theta) ) = \*\eta^\intercal_{q_{\alpha}} T(\*\theta). \nonumber
\end{align}
Let $\*\mu_{q_{\alpha}} = \int q_{\alpha}(\*\theta) T(\*\theta) \mathrm{d}\*\theta$ be the mean parameters of $q_{\alpha}(\*\theta)$. From the moment matching projection, $\*\mu_{q_{\alpha}} = \int p^{(i)}_{\alpha}(\*\theta) T(\*\theta) \mathrm{d}\*\theta$.
Using Taylor series to expand $\log \mathrm{proj}( p^{(i)}_{\alpha}(\*\theta) )$ as a function of $\alpha$ about $\alpha = 0$:
\begin{align}
\log \mathrm{proj}( p^{(i)}_{\alpha}(\*\theta) ) = \log \mathrm{proj}( p^{(i)}_{\alpha}(\*\theta) ) \big|_{\alpha=0} + \alpha \left( \frac{\mathrm{d}}{\mathrm{d} \alpha} \log \mathrm{proj}( p^{(i)}_{\alpha}(\*\theta) ) \big|_{\alpha=0} \right) + F(\alpha), \nonumber
\end{align}
where $F(\alpha) = \sum_{t=2}^{\infty} \frac{\alpha^{\top}}{t!} ( \frac{\mathrm{d}^{\top}}{\mathrm{d} \alpha^{\top}} \log \mathrm{proj}( p^{(i)}_{\alpha}(\*\theta) ) \big|_{\alpha=0} )$ collects all the high order terms in the expansion.
Since $\log \mathrm{proj}( p^{(i)}_{\alpha}(\*\theta) ) \big|_{\alpha=0} = \log q^{(i-1)}(\*\theta)$, the above equation becomes:
\begin{align}
\log \mathrm{proj}( p^{(i)}_{\alpha}(\*\theta) ) = \log q^{(i-1)}(\*\theta) + \alpha \left( \frac{\mathrm{d}}{\mathrm{d} \alpha} \log \mathrm{proj}( p^{(i)}_{\alpha}(\*\theta) ) \big|_{\alpha=0} \right) + F(\alpha). \label{eq:logproj}
\end{align}
Now consider $\frac{\mathrm{d}}{\mathrm{d} \alpha} \log \mathrm{proj}( p^{(i)}_{\alpha}(\*\theta) )$, we have:
\begin{align*}
\frac{\mathrm{d}}{\mathrm{d} \alpha} \log \mathrm{proj}( p^{(i)}_{\alpha}(\*\theta) ) 
&= T(\*\theta)^\intercal \frac{\mathrm{d} \*\eta_{q_{\alpha}}}{\mathrm{d} \alpha} = T(\*\theta)^\intercal \frac{\mathrm{d} \*\eta_{q_{\alpha}}}{\mathrm{d} \*\mu_{q_{\alpha}}} \frac{\mathrm{d} \*\mu_{q_{\alpha}}}{\mathrm{d} \alpha} \\
&= T(\*\theta)^\intercal \frac{\mathrm{d} \*\eta_{q_{\alpha}}}{\mathrm{d} \*\mu_{q_{\alpha}}} \frac{\mathrm{d}}{\mathrm{d} \alpha} \int p^{(i)}_{\alpha}(\*\theta) T(\*\theta) \mathrm{d}\*\theta \\
&= T(\*\theta)^\intercal \frac{\mathrm{d} \*\eta_{q_{\alpha}}}{\mathrm{d} \*\mu_{q_{\alpha}}} \frac{\mathrm{d}}{\mathrm{d} \alpha} \int q^{(i-1)}(\*\theta) \left( \frac{p(\mathbf{y}_{k} | \*\theta)}{t_{k}^{(i-1)}(\*\theta)}\right)^{\alpha} T(\*\theta) \mathrm{d}\*\theta \\
&= T(\*\theta)^\intercal \frac{\mathrm{d} \*\eta_{q_{\alpha}}}{\mathrm{d} \*\mu_{q_{\alpha}}} \int q^{(i-1)}(\*\theta) \left( \frac{p(\mathbf{y}_{k} | \*\theta)}{t_{k}^{(i-1)}(\*\theta)} \right)^{\alpha} \log \frac{p(\mathbf{y}_{k} | \*\theta)}{t_{k}^{(i-1)}(\*\theta)} T(\*\theta) \mathrm{d}\*\theta \\
&= T(\*\theta)^\intercal \frac{\mathrm{d} \*\eta_{q_{\alpha}}}{\mathrm{d} \*\mu_{q_{\alpha}}} \frac{\mathrm{d}}{\mathrm{d} \*\eta_{q^{(i-1)}}} \int q^{(i-1)}(\*\theta) \left( \frac{p(\mathbf{y}_{k} | \*\theta)}{t_{k}^{(i-1)}(\*\theta)} \right)^{\alpha} \log \frac{p(\mathbf{y}_{k} | \*\theta)}{t_{k}^{(i-1)}(\*\theta)} \mathrm{d}\*\theta,
\end{align*}
where $\*\eta_{q^{(i-1)}}$ is the natural parameters of $q^{(i-1)}$.

Thus,
\begin{align*}
\frac{\mathrm{d}}{\mathrm{d} \alpha} \log \mathrm{proj}( p^{(i)}_{\alpha}(\*\theta) ) \big|_{\alpha=0} 
&= T(\*\theta)^\intercal \frac{\cancel{\mathrm{d} \*\eta_{q^{(i-1)}}}}{\mathrm{d} \*\mu_{q^{(i-1)}}} \frac{\mathrm{d}}{\cancel{\mathrm{d} \*\eta_{q^{(i-1)}}}} \int q^{(i-1)}(\*\theta) \log \frac{p(\mathbf{y}_{k} | \*\theta)}{t_{k}^{(i-1)}(\*\theta)} \mathrm{d}\*\theta \\
&= T(\*\theta)^\intercal \frac{\mathrm{d}}{\mathrm{d} \*\mu_{q^{(i-1)}}} \int q^{(i-1)}(\*\theta) \log \frac{p(\mathbf{y}_{k} | \*\theta)}{t_{k}^{(i-1)}(\*\theta)} \mathrm{d}\*\theta.
\end{align*}
Plug this into \eqref{eq:logproj}, we obtain:
\begin{align}
\log \mathrm{proj}( p^{(i)}_{\alpha}(\*\theta) ) = \log q^{(i-1)}(\*\theta) + \alpha T(\*\theta)^\intercal \frac{\mathrm{d}}{\mathrm{d} \*\mu_{q^{(i-1)}}} \int q^{(i-1)}(\*\theta) \log \frac{p(\mathbf{y}_{k} | \*\theta)}{t_{k}^{(i-1)}(\*\theta)} \mathrm{d}\*\theta + F(\alpha). \nonumber
\end{align}
From this equation and \eqref{eq:log-ti-1},
\begin{equation}
\begin{aligned}
\log t^{(i)}_{k}(\*\theta) 
&= \frac{1}{\alpha} \left( \alpha T(\*\theta)^\intercal \frac{\mathrm{d}}{\mathrm{d} \*\mu_{q^{(i-1)}}} \int q^{(i-1)}(\*\theta) \log \frac{p(\mathbf{y}_{k} | \*\theta)}{t_{k}^{(i-1)}(\*\theta)} \mathrm{d}\*\theta + F(\alpha) \right) + \log t^{(i-1)}_{k} (\*\theta) \\
&= T(\*\theta)^\intercal \frac{\mathrm{d}}{\mathrm{d} \*\mu_{q^{(i-1)}}} \int q^{(i-1)}(\*\theta) \log \frac{p(\mathbf{y}_{k} | \*\theta)}{t_{k}^{(i-1)}(\*\theta)} \mathrm{d}\*\theta + \frac{F(\alpha)}{\alpha} + \log t^{(i-1)}_{k} (\*\theta) \\
&= T(\*\theta)^\intercal \frac{\mathrm{d}}{\mathrm{d} \*\mu_{q^{(i-1)}}} \int q^{(i-1)}(\*\theta) \log p(\mathbf{y}_{k} | \*\theta) \mathrm{d}\*\theta \\
&{\hskip 5mm} - T(\*\theta)^\intercal \frac{\mathrm{d}}{\mathrm{d} \*\mu_{q^{(i-1)}}} \int q^{(i-1)}(\*\theta) \log t_{k}^{(i-1)}(\*\theta) \mathrm{d}\*\theta + \frac{F(\alpha)}{\alpha} + \log t^{(i-1)}_{k} (\*\theta). \label{eq:log-ti-2}
\end{aligned}
\end{equation}
Note that:
\begin{align*}
T(\*\theta)^\intercal \frac{\mathrm{d}}{\mathrm{d} \*\mu_{q^{(i-1)}}} \int q^{(i-1)}(\*\theta) \log t_{k}^{(i-1)}(\*\theta) \mathrm{d}\*\theta 
&= \log t_{k}^{(i-1)}(\*\theta) \frac{\mathrm{d}}{\mathrm{d} \*\mu_{q^{(i-1)}}} \int T(\*\theta)^\intercal q^{(i-1)}(\*\theta) \mathrm{d}\*\theta \\
&= \log t_{k}^{(i-1)}(\*\theta) \frac{\mathrm{d} \*\mu_{q^{(i-1)}}}{\mathrm{d} \*\mu_{q^{(i-1)}}} \\
&= \log t_{k}^{(i-1)}(\*\theta).
\end{align*}
Hence, \eqref{eq:log-ti-2} becomes:
\begin{align*}
T(\*\theta)^\intercal \*\eta^{(i)}_{k}
&= T(\*\theta)^\intercal \frac{\mathrm{d}}{\mathrm{d} \*\mu_{q^{(i-1)}}} \int q^{(i-1)}(\*\theta) \log p(\mathbf{y}_{k} | \*\theta) \mathrm{d}\*\theta + \frac{F(\alpha)}{\alpha},
\end{align*}
which is equivalent to:
\begin{align}
\label{eq:pep-update}
T(\*\theta)^\intercal \*\eta^{(i)}_{k}
&= T(\*\theta)^\intercal \frac{\mathrm{d}}{\mathrm{d} \*\mu_q} \int q(\*\theta) \log p(\mathbf{y}_{k} | \*\theta) \mathrm{d}\*\theta + \frac{F(\alpha)}{\alpha}.
\end{align}
Let $\bar{q}(\*\theta)$ be the normalised distribution of $q(\*\theta)$, i.e. $q(\*\theta) = Z_q \bar{q}(\*\theta)$. The fixed point update for local VI is:
\begin{align}
\*\eta^{(i)}_{k}  &
= \frac{\mathrm{d}}{\mathrm{d}\*\mu_{\bar{q}}} \int \bar{q}(\*\theta) \log p(\bfy_{k}|\*\theta) \mathrm{d}\*\theta, \nonumber
\end{align}
where $\*\mu_{\bar{q}} = \int T(\*\theta)^\intercal \bar{q}(\*\theta) \mathrm{d}\*\theta = \*\mu_q/Z_q$.
We have:
\begin{align*}
\frac{\mathrm{d}}{\mathrm{d}\*\mu_{\bar{q}}} \int \bar{q}(\*\theta) \log p(\bfy_{k}|\*\theta) \mathrm{d}\*\theta
&= \frac{\mathrm{d}\*\mu_q}{\mathrm{d}\*\mu_{\bar{q}}} \frac{\mathrm{d}}{\mathrm{d}\*\mu_q} \int \frac{1}{Z_q} q(\*\theta) \log p(\bfy_{k}|\*\theta) \mathrm{d}\*\theta \\
&= (\cancel{Z_q} \mathbb{I}) \frac{1}{\cancel{Z_q}} \frac{\mathrm{d}}{\mathrm{d}\*\mu_q} \int q(\*\theta) \log p(\bfy_{k}|\*\theta) \mathrm{d}\*\theta \\
&= \frac{\mathrm{d}}{\mathrm{d}\*\mu_q} \int q(\*\theta) \log p(\bfy_{k}|\*\theta) \mathrm{d}\*\theta.
\end{align*}
From this equation and the fact that $\frac{F(\alpha)}{\alpha} \rightarrow 0$ when $\alpha \rightarrow 0$, the fixed point update for PVI satisfies the Power-EP update in \eqref{eq:pep-update} when $\alpha \rightarrow 0$.

\section{Gradients of the Free-Energy With Respect to Hyperparameters}
\label{sec:hyper-gradients}
In this section, we derive the gradients of the global free energy w.r.t.\ the hyperparameters for general $q(\*\theta)$.

\subsection{General Derivation}
The global free-energy depends on the model hyperparameters through the joint distribution $p(\bfy, \*\theta | \*\epsilon)$,
\begin{equation}
    \mcF(\*\epsilon, q(\*\theta)) = \int q(\*\theta) \log \frac{p(\bfy, \*\theta | \*\epsilon)}{q(\*\theta)} \mathrm{d}\*\theta. \nonumber
\end{equation}
Taking derivatives w.r.t.\ $\*\epsilon$ gives
\begin{equation}
\begin{aligned}
    \frac{\mathrm{d}}{\mathrm{d}\*\epsilon} \mcF(\*\epsilon, q(\*\theta)) &= \sum_{m=1}^M \int q(\*\theta) \frac{\mathrm{d}}{\mathrm{d}\*\epsilon} \log p(\bfy_m | \*\theta) \mathrm{d}\*\theta + \int q(\*\theta) \frac{\mathrm{d}}{\mathrm{d}\*\epsilon} \log p(\*\theta | \*\epsilon) \mathrm{d}\*\theta \\
    &= \sum_{m=1}^M \Exp{q(\*\theta)}{\frac{\mathrm{d}}{\mathrm{d}\*\epsilon} \log p(\bfy_m | \*\theta)} + \Exp{q(\*\theta)}{\frac{\mathrm{d}}{\mathrm{d}\*\epsilon} \log p(\*\theta | \*\epsilon)}
\end{aligned}
\nonumber
\end{equation}

\subsection{Derivation for Collapsed Bound}
In this section, we consider the setting in which $q(\*\theta)$ is the optimal variational approximation for a fixed setting of the hyperparameters, so depends implicitly on $\*\epsilon$:
\begin{equation}
    q(\*\theta; \*\epsilon) = \argmax q(\*\theta) \int q(\*\theta) \log \frac{p(\bfy, \*\theta | \*\epsilon)}{q(\*\theta)} \mathrm{d}\*\theta. \nonumber
\end{equation}
Note that here we have been careful to represent the two distinct ways that the free-energy depends on the hyperparameters: 1) through the log-joint's dependence; and 2) through the optimal approximate posterior's implicit dependence. Taking derivatives w.r.t.\ $\*\epsilon$ gives
\begin{equation}
    \frac{\mathrm{d}}{\mathrm{d}\*\epsilon} \mcF(\*\epsilon, q(\*\theta; \*\epsilon)) = \left.\frac{\mathrm{d}}{\mathrm{d}\*\epsilon} \mcF(\*\epsilon, q(\*\theta; \*\epsilon')) \right|_{\*\epsilon' = \*\epsilon} + \left.\frac{\mathrm{d}}{\mathrm{d}\*\epsilon'} \mcF(\*\epsilon, q(\*\theta; \*\epsilon')) \right|_{\*\epsilon' = \*\epsilon}. \nonumber
\end{equation}
The first captures the dependence through the log-joint distribution and is given by
\begin{equation}
    \left.\frac{\mathrm{d}}{\mathrm{d}\*\epsilon} \mcF(\*\epsilon, q(\*\theta; \*\epsilon')) \right|_{\*\epsilon' = \*\epsilon} = \sum_{m=1}^M \Exp{q(\*\theta; \*\epsilon)}{\frac{\mathrm{d}}{\mathrm{d}\*\epsilon} \log p(\bfy_m | \*\theta)} + \Exp{q(\*\theta; \*\epsilon)}{\frac{\mathrm{d}}{\mathrm{d}\*\epsilon} \log p(\*\theta | \*\epsilon)}. \nonumber
\end{equation}
The second term captures the dependence through the optimal approximate posterior's implicit dependence on $\*\epsilon$ and is given by
\begin{equation}
    \left.\frac{\mathrm{d}}{\mathrm{d}\*\epsilon'} \mcF(\*\epsilon, q(\*\theta; \*\epsilon')) \right|_{\*\epsilon' = \*\epsilon} = \left.\frac{\mathrm{d q(\*\theta; \*\epsilon)}}{\mathrm{d}\*\epsilon'} \frac{\mathrm{d }}{\mathrm{d}q(\*\theta)} \mcF(\*\epsilon, q(\*\theta)) \right|_{\*\epsilon' = \*\epsilon, q(\*\theta)=q(\*\theta; \*\epsilon)} = \*0. \nonumber
\end{equation}
Here, we have used the fact that we are at the collapsed bound and so the functional derivative w.r.t.\ $q(\*\theta)$ is zero. Thus, for both the general $q(\*\theta)$ and optimal $q(\*\theta; \*\epsilon)$, we obtain the same expression for gradients of the global free-energy w.r.t.\ the hyperparameters.

\begin{figure}[!ht]
\begin{center}
\includegraphics[width=8cm]{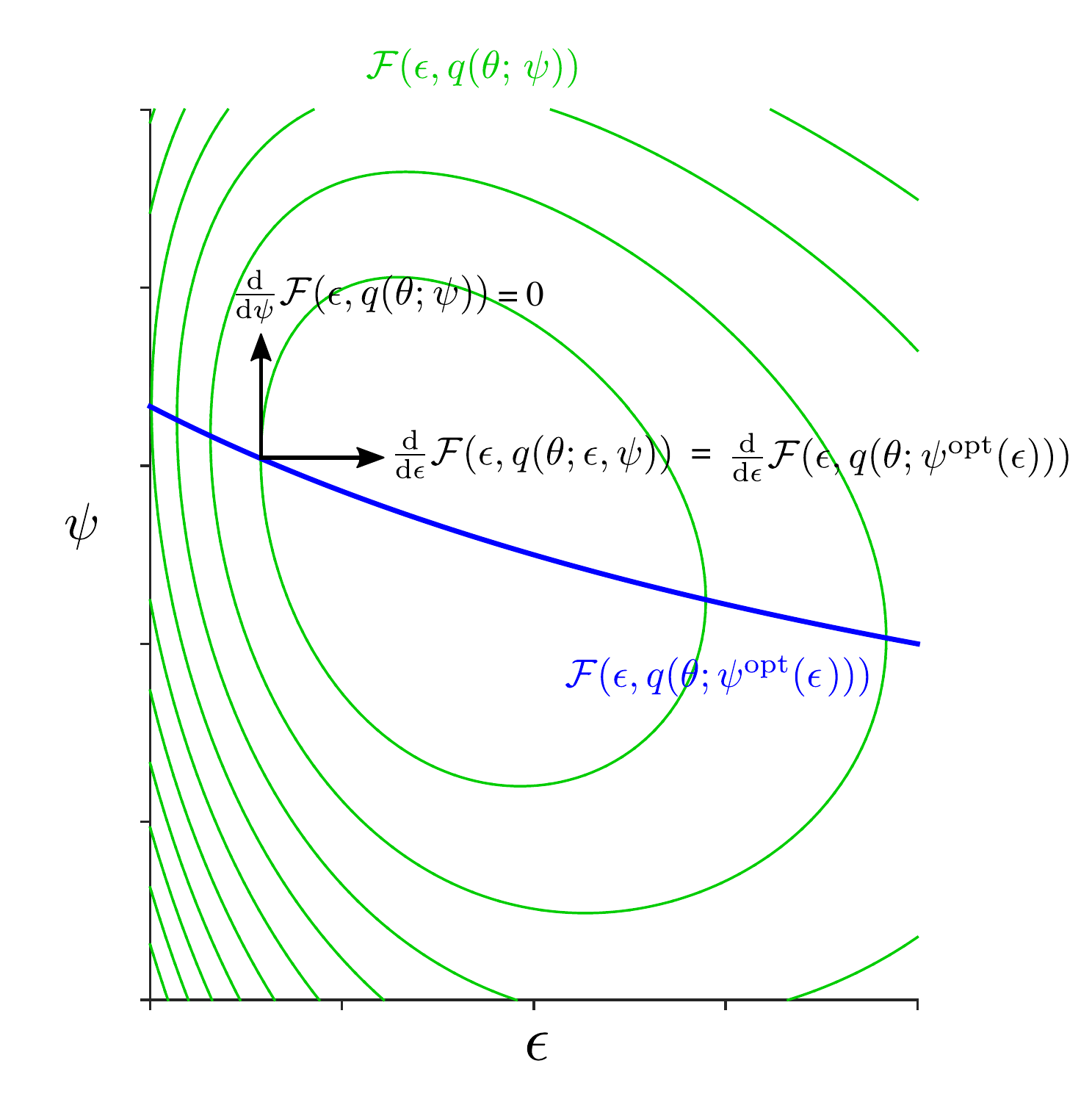}
\end{center}
\caption{Contours of the free-energy $\mathcal{F}( \*\epsilon, q(\*\theta;\*\psi))$ are shown in green as a function of the hyperparameters $\*\epsilon$ and the variational parameters of the approximate posterior $\phi$. The collapsed bound $\mathcal{F}(\*\epsilon, q(\*\theta; \*\psi^{\text{opt}}(\*\epsilon)))$  is shown in blue. The gradients of the free-energy with respect to the variational parameters are zero along the collapsed bound $\frac{\mathrm{d}}{ \mathrm{d} \*\psi} \mathcal{F}( \*\epsilon, q(\*\theta; \*\psi)) \vert_{\*\psi = \*\psi^{\text{opt}}} = 0$, by definition. This means that the gradients of the collapsed free-energy as a function of the hyperparameters are equal to those of the free-energy itself, $\frac{\mathrm{d}}{ \mathrm{d} \*\epsilon} \mathcal{F}(\*\epsilon,q(\*\theta;  \*\psi)) = \frac{\mathrm{d}}{ \mathrm{d} \*\epsilon} \mathcal{F}(\*\epsilon,q(\*\theta;  \*\psi^{\text{opt}}(\*\epsilon)))$.
 }\label{fig:free-energy-schem}
\end{figure}

\Cref{fig:free-energy-schem} provides some intuition for these results. Note that in the case where the approximating family includes the true posterior distribution, the collapsed bound is equal to the log-likelihood of the hyperparameters. So, the result shows that the gradient of the log-likelihood w.r.t. the hyperparameters is equal to the gradient of the free-energy w.r.t. the hyperparameters, treating $q$ as fixed. Often this is computed in the M-step of variational EM, but it is used in coordinate ascent, which can be slow to converge. Instead, this gradient can be passed to an optimiser to perform direct gradient-based optimisation of the log-likelihood.

\section{Forms of Inhomogeneity}
\label{sec:appen:inhomogeneity}
Throughout this paper we have used the blanket term `inhomogeneous' to describe a variety of different forms of inhomogeneity across data partitions. This section sheds light on the different forms of inhomogeneity and describes the suitability of PVI, and more generally federated learning, in the presence of each. Each client's dataset is assumed to be sampled according to a local dataset distribution $\mcP_m(\bfx, y)$, and we consider the task of learning a discriminative model $p(y | \bfx)$\footnote{If instead we consider the task of learning a generative model $p(\bfx, y)$, then it would be advantageous to learn personalised models in the presence of covariate shift.}. We consider three distinct sources of inhomogeneity:
\begin{enumerate}
    \item the local dataset distributions differ between clients;
    \item the size of local datasets differ between clients;
    \item the available compute differs between clients.
\end{enumerate}
If the local dataset distributions are the same between clients, then learning a shared model through federated learning is advantageous. However, there are a number of ways in which local dataset distributions can differ. The first is covariate shift, in which the marginal distributions $\mcP_m(\bfx)$ and $\mcP_m(y)$ differ between clients whilst $\mcP_m(y | \bfx)$ are shared. Since we are concerned with modelling the discriminative distribution $p(y | \bfx)$, the use federated learning of a single, shared model in this setting seems advantageous as data from each client provides us with information about the relationship between inputs and outputs. However, as we tend to use parametric models of fixed capacity it may be preferable to learn personalised models which more accurately model the discriminative distribution for inputs with high probability under $\mcP_m(\bfx)$. When the conditional distributions $\mcP_m(y | \bfx)$ differ between clients, it is clearly desirable to learn a personalised discriminative model for each client which models the local conditional distribution $\mcP_m(y | \bfx)$. However, although local dataset distributions may differ, they may be similar to the extent that it is still useful to share information between clients without assuming the same model for each---this describes the personalised federated learning setting.

\section{Experimental Details}
\label{sec:appen:experiments}
\subsection{UCI Classification}
\label{sec:appen:experiments:uci}
We follow the dataset distribution scheme of \cite{sharma2019differentially} to split each dataset into $M=10$ partitions. The degree of inhomogeneity is controlled by parameters $\kappa$ and $\beta$, which control label and size imbalance, respectively. Half the clients are labelled as `small' and the other half as `large'. The amount of data assigned to clients in the same group is governed by $\beta \in (0, 1)$: 
\begin{equation}
    N_{\text{small}} = \left \lfloor{\frac{N}{M} (1 - \beta)}\right \rfloor \qquad N_{\text{large}} = \left \lfloor{\frac{N}{M} (1 + \beta)}\right \rfloor. \nonumber
\end{equation}
where $N$ is the size of the dataset. As $\beta \rightarrow 1$ the class size imbalance increases. The fraction of the positive labels in small clients depends on $\kappa \in [-\frac{\eta}{1 - \eta}, 1]$:
\begin{equation}
    \eta_{\text{small}} = \eta + (1 - \eta) \kappa \nonumber
\end{equation}
where $\eta$ denotes the proportion of positive labels in the entire dataset. Setting $\kappa = -\frac{\eta}{1 - \eta}$ implies the small clients contain only positive labels, whereas setting $\kappa = 1$ implies the small clients contain only negative labels. For all three datasets, we consider two splits generated from two distinct tuples $(\beta, \kappa)$\footnote{A different $(\beta, \kappa)$ for split B was used for the credit dataset to ensure $\kappa$ is in range.}. The resultant data distributions are described in \Cref{tab:uci_splits}.

\begin{table}[!h]
    \centering
    \caption{The two different splits for each of the three UCI binary classification tasks. $\beta$ controls for class size imbalance and $\kappa$ controls for the fraction of positive labels in the small class. $\eta$ indicates the proportion of positive labels.}
    \small
    \begin{tabular}{rccccccccc}
    \toprule
         Dataset & Split & $\beta$ & $\kappa$ && $N_{\text{small}}$ & $\eta_{\text{small}}$ && $N_{\text{large}}$ & $\eta_{\text{large}}$  \\
         \midrule
         \multirow{2}{*}{adult} & A & 0 & 0 && 3907 & 0.240 && 3907 & 0.240 \\
         & B & 0.7 & -3 && 1172 & 0.959 && 6642 & 0.111 \\
         \midrule
         \multirow{2}{*}{bank} & A & 0 & 0 && 3616 & 0.117 && 3616 & 0.117 \\
         & B & 0.7 & -3 && 1085 & 0.466 && 6148 & 0.054 \\
         \midrule 
         \multirow{2}{*}{credit} & A & 0 & 0 && 52 & 0.550 && 52 & 0.550 \\
         & B & 0.3 & -0.7 && 36 & 0.944 && 67 & 0.337 \\
         \bottomrule
    \end{tabular}
    \label{tab:uci_splits}
\end{table}

The variational objectives were optimised using the Adam optimiser \citep{kingma2014adam} with a batch size of 128 and a learning rate of $1\times 10^{-2}$.

\subsection{Mortality Prediction}
\label{sec:appen:experiments:mortality}
The five splits used in the mortality prediction experiment are detailed in \Cref{tab:mimic3_splits}. These were obtained by performing k-means clustering on the input features, and can thus be interpreted as having different dataset distributions.

\begin{table}[!h]
    \centering
    \small
    \caption{The $M=5$ partitions of the MIMIC-III mortality prediction dataset. $\eta$ denotes the proportion of positive labels.}
    \begin{tabular}{cccccccccccccccccccc}
        \toprule
         \multicolumn{2}{c}{Client 1} && \multicolumn{2}{c}{Client 2} && \multicolumn{2}{c}{Client 3} && \multicolumn{2}{c}{Client 4} && \multicolumn{2}{c}{Client 5} \\
         $N_1$ & $\eta_1$ && $N_2$ & $\eta_2$ && $N_3$ & $\eta_3$ && $N_4$ & $\eta_4$ && $N_5$ & $\eta_5$ \\
         \midrule
         3252 & 0.315 && 3366 & 0.180 && 895 & 0.244 && 3702 & 0.023 && 6688 & 0.073 \\
         \bottomrule
    \end{tabular}
    \label{tab:mimic3_splits}
\end{table}

The variational objectives were optimised using the Adam optimiser \citep{kingma2014adam} with a batch size of 128 and a learning rate of $1\times 10^{-3}$ for both the logistic regression and BNN experiments.

\subsection{Federated MNIST}
\label{sec:appen:experiments:fmnist}
The variational objectives were optimised using the Adam optimiser \citep{kingma2014adam} with a batch size of 512 and a learning rate of $2\times 10^{-3}$.

\begin{figure}
    \centering
    \includegraphics[width=\textwidth]{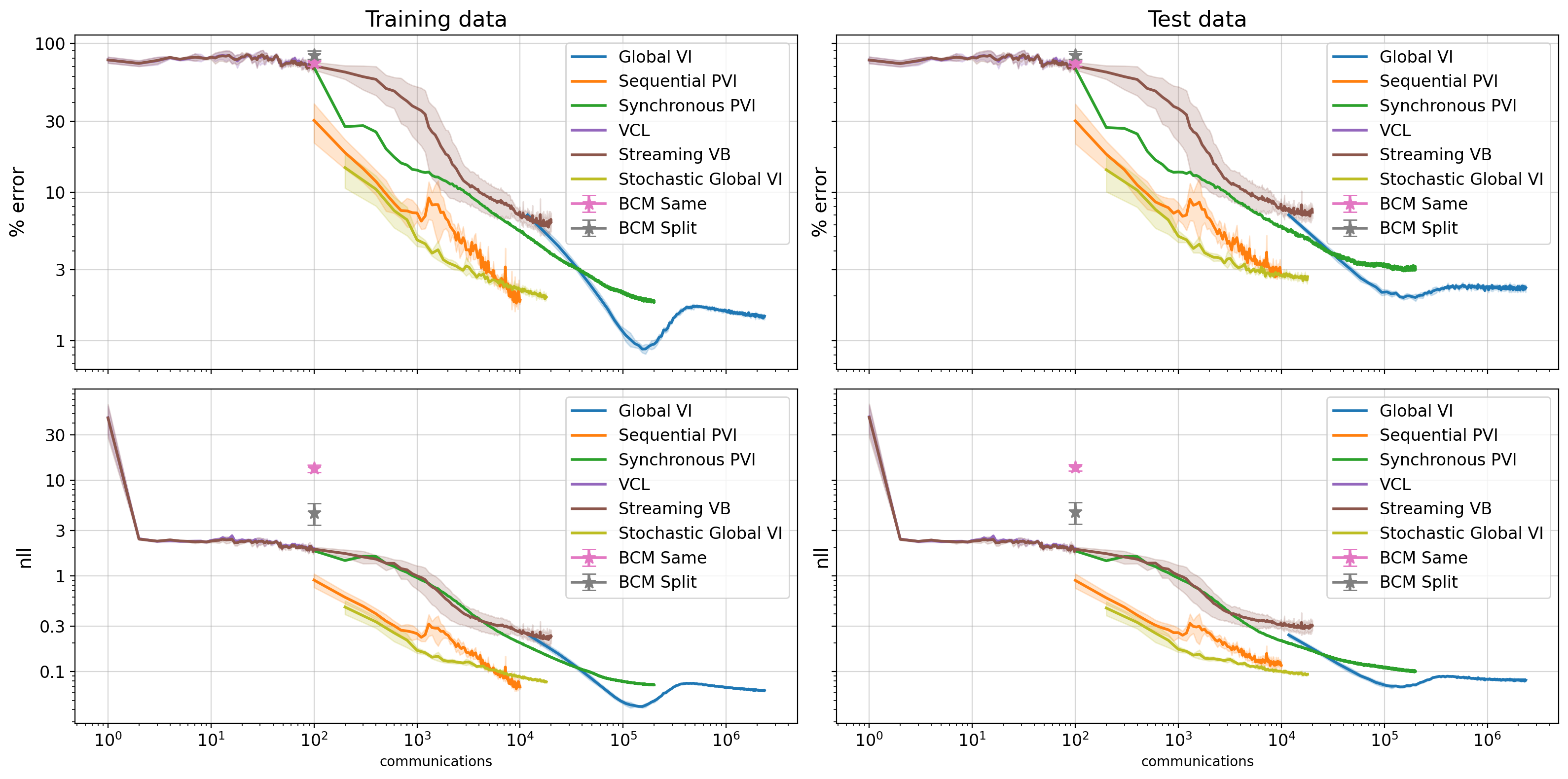}
    \caption{Plots of the predictive performance of the federated learning schemes on the inhomogeneous split of the MNIST classification task. The split is constructed by partitioning the 60,000 training images into $M=100$ partitions, each with at most two digits. The mean $\pm$ standard deviation computed using five random initialisations is shown.}
    \label{fig:fmnist_inhomo_pvi_sgvi}
\end{figure}

In \Cref{fig:fmnist_inhomo_pvi_sgvi}, we compare the performance of the stochastic implementation of global VI in which we visit a single client at a time performing an update of the approximate posterior using a biased estimate of the ELBO. This achieves faster initial convergence than sequential PVI, which can be attributed to the difficulty of learning approximate factors in the presence of severe inhomogeneity. 

\begin{figure}
    \centering
    \includegraphics[width=\textwidth]{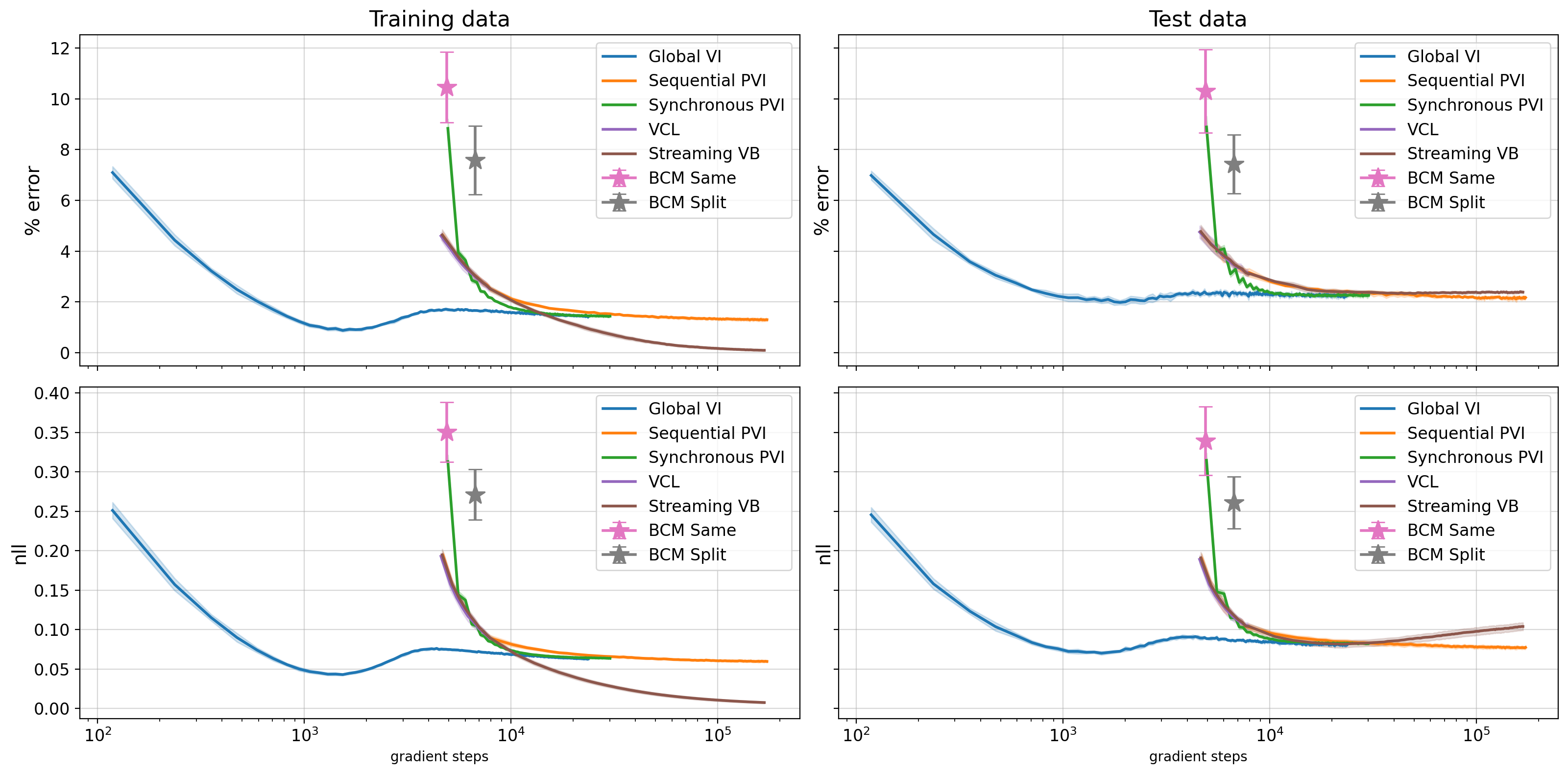}
    \caption[Predictive performance on the homogeneous split of the MNIST classification task.]{Plots of the predictive performance against number of gradient steps for the federated learning schemes on the homogeneous split of the MNIST classification task. The mean $\pm$ standard deviation computed using five random initialisations is shown.}
    \label{fig:fmnist_homo_pvi_steps}
\end{figure}

In \Cref{fig:fmnist_homo_pvi_steps}, we compare the number of synchronous gradient steps performed by each of the federated learning algorithms. If the time taken to communicate between the server and clients is ignored, then this serves as a proxy for the run time of each method. Sequential methods are the slowest as they do not update clients in parallel.

\begin{figure}
    \centering
    \includegraphics[width=\textwidth]{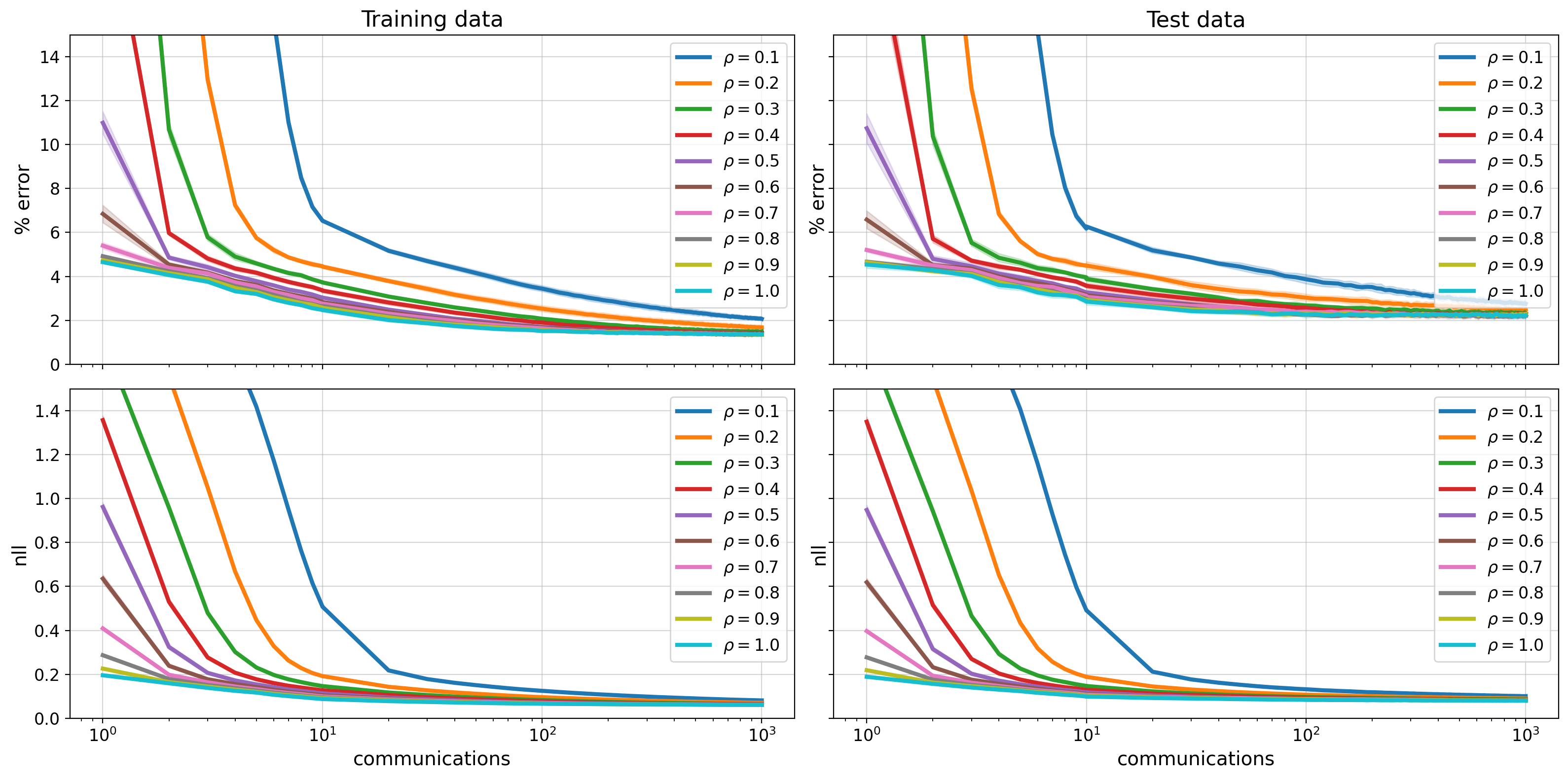}
    \caption[Predictive performance of sequential PVI on the homogeneous split of the MNIST classification task.]{Plots of the predictive performance against number of communications for sequential PVI with different damping factors on the homogeneous split of the MNIST classification task. The mean $\pm$ standard deviation computed using five random initialisations is shown.}
    \label{fig:fmnist_homo_sequential}
\end{figure}

\begin{figure}
    \centering
    \includegraphics[width=\textwidth]{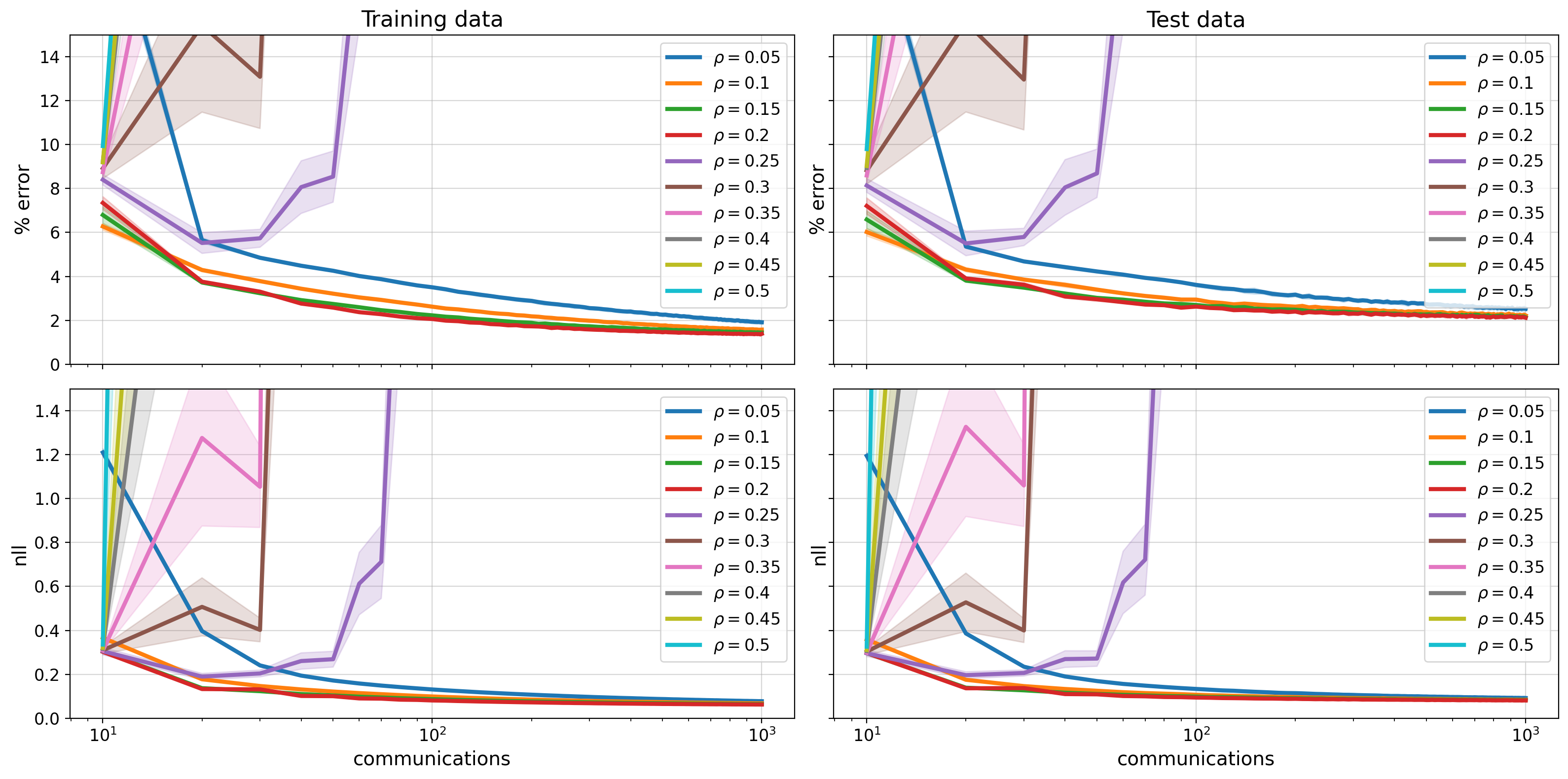}
    \caption[Predictive performance of synchronous PVI on the homogeneous split of the MNIST classification task.]{Plots of the predictive performance against number of communications for synchronous PVI with different damping factors on the homogeneous split of the MNIST classification task. The mean $\pm$ standard deviation computed using five random initialisations is shown.}
    \label{fig:fmnist_homo_synchronous}
\end{figure}

\Cref{fig:fmnist_homo_sequential,fig:fmnist_homo_synchronous} compare the effect the damping factor $\rho$ has on the convergence of sequential and synchronous PVI for the homogeneous MNIST classification task. For sequential PVI, less damping results in faster convergence. The same is not true for synchronous PVI: using less damping results in faster convergence for $\rho < 0.25$. For $\rho \geq 0.25$, the aggregation of approximate factors does not result in a normalisable distribution and the performance diverges.

\end{document}